\documentclass[a4paper,conference]{IEEEtran}
\ifCLASSINFOpdf
\else
\fi

\usepackage{graphicx}
\usepackage{amsmath}
\usepackage{amssymb}
\usepackage{enumitem}
\usepackage {xcolor}
\usepackage{multirow}
\graphicspath{ {./Pics/} }
\usepackage{float}
\usepackage[export]{adjustbox}
\usepackage{tensor}
\usepackage{hyperref}

\usepackage{graphicx}
\usepackage{amsmath}
\usepackage{amssymb}
\usepackage{enumitem}
\usepackage{multirow}
\graphicspath{ {./Pics/} }
\usepackage{float}
\usepackage[export]{adjustbox}
\usepackage{tensor}
\usepackage{caption}
\newcommand{\norm}[1]{\left\lVert#1\right\rVert}
\newcommand{\normmod}[1]{\biggr\lVert#1\biggr\rVert}
\renewcommand{\thefootnote}{\fnsymbol{footnote}}
\usepackage{array}
\usepackage[caption=false,font=footnotesize]{subfig}
\begin{document}
%
\def\BibTeX{{\rm B\kern-.05em{\sc i\kern-.025em b}\kern-.08em
    T\kern-.1667em\lower.7ex\hbox{E}\kern-.125emX}}
    
\title{DMD-Net: Deep Mesh Denoising Network}
\let\oldtwocolumn\twocolumn
\renewcommand\twocolumn[1][]{%
    \oldtwocolumn[{#1}{
    \begin{center}
        \vspace{-1cm}
          \includegraphics[width=\textwidth]{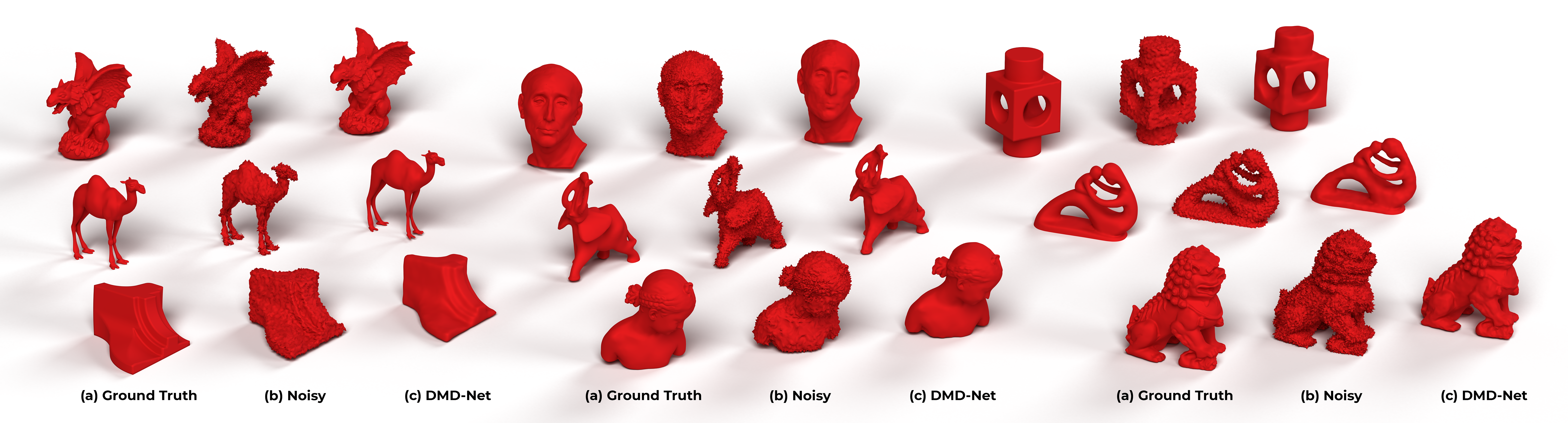}
          \captionof{figure}{Mesh denoising results obtained using DMD-Net on several meshes. The objects on the left (a) denote the original ground truth meshes. The objects in the middle (b) denote the noisy meshes obtained by adding noise to the ground truth meshes. The objects on the right (c) denote the denoised mesh results of DMD-Net acting on the noisy objects.}
          \label{fig:teaser}
        \end{center}
    }]
}

\author{\IEEEauthorblockN{Aalok Gangopadhyay$^*$}
\IEEEauthorblockA{CVIG Lab, IIT Gandhinagar\\
aalok@iitgn.ac.in}
\and
\IEEEauthorblockN{Shashikant Verma$^*$}
\IEEEauthorblockA{CVIG Lab, IIT Gandhinagar\\
shashikant.verma@iitgn.ac.in}
\and
\IEEEauthorblockN{Shanmuganathan Raman}
\IEEEauthorblockA{CVIG Lab, IIT Gandhinagar\\
shanmuga@iitgn.ac.in}}


%


\maketitle
\def\thefootnote{*}\footnotetext{These authors contributed equally to this work.}

\begin{abstract}
We present Deep Mesh Denoising Network (DMD-Net), an end-to-end deep learning framework, for solving the mesh denoising problem. DMD-Net consists of a Graph Convolutional Neural Network in which aggregation is performed in both the primal as well as the dual graph. This is realized in the form of an asymmetric two-stream network, which contains a primal-dual fusion block that enables communication between the primal-stream and the dual-stream. We develop a Feature Guided Transformer (FGT) paradigm, which consists of a feature extractor, a transformer, and a denoiser. The feature extractor estimates the local features, that guide the transformer to compute a transformation, which is applied to the noisy input mesh to obtain a useful intermediate representation. This is further processed by the denoiser to obtain the denoised mesh. Our network is trained on a large scale dataset of 3D objects. We perform exhaustive ablation studies to demonstrate that each component in our network is essential for obtaining the best performance. We show that our method obtains competitive or better results when compared with the state-of-the-art mesh denoising algorithms. We demonstrate that our method is robust to various kinds of noise. We observe that even in the presence of extremely high noise, our method achieves excellent performance.
\end{abstract}


%
\IEEEpeerreviewmaketitle

\begin{figure*}[t]
  \includegraphics[width=\linewidth]{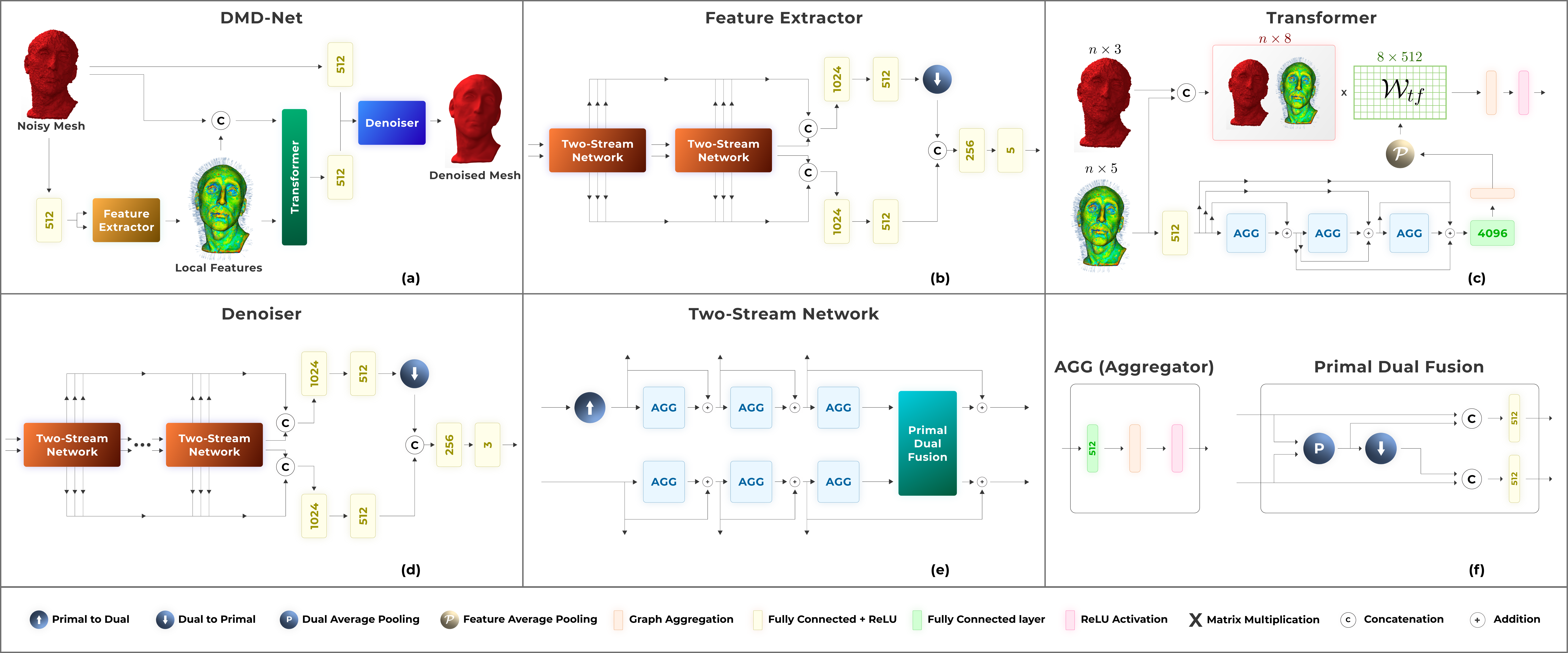}
  \caption{The proposed DMD-Net architecture for mesh denoising: (a) Overall architecture depicting the feature guided transformer (FGT) paradigm. (b) The Feature Extractor, (c) The Transformer, (d) The Denoiser,  (e) The Two-Stream Network, and (f) The Aggregator (AGG) and the Primal Dual Fusion (PDF) block.}
  \vspace{-0.5cm}
  \label{fig:DMD-Net}
\end{figure*}
\section{Introduction}
There has been a tremendous rise in the use of 3D acquisition technology for obtaining high fidelity digital models of real world 3D objects. 3D models obtained using CAD softwares are designed by humans and appear perfect in shape. On the contrary, 3D models obtained using 3D scanners contain noise owing to the measurement error of the scanning devices. In the case of 3D objects obtained using photogrammetry, considerable noise is introduced because of the reconstruction error. 
The task of eliminating these various types of noise to restore the object to its original shape is known as mesh denoising and is a fundamental problem in the field of 3D shape analysis.

\textit{Challenges}. 
Given a noisy mesh, there are many possible candidates for the original noise-free mesh. This makes mesh denoising a highly ill-posed inverse problem. There is a natural trade-off involved in mesh denoising between that of eliminating the noise and preserving the high frequency details. 
Striking the perfect balance in this trade-off is a crucial aspect for any good denoising algorithm. 
Existing works in the literature have addressed these challenges and have devised innovative strategies for solving mesh denoising. 
While this problem has been approached by a variety of classical approaches, there are not many learning based approaches proposed that have addressed this problem.


\textit{Contributions}. In this work, we design a novel GNN architecture to solve the mesh denoising problem. We use a large scale dataset to train a mesh denoising architecture in the deep learning framework. Unlike many existing learning-based methods, our method is end-to-end trainable and takes entire mesh as input instead of the patch based approach. Our contributions are three-fold:
\begin{enumerate}[noitemsep,nolistsep]
    
    \item We propose a novel Graph Convolutional Neural Network architecture, which employs primal-dual graph aggregation and a Feature Guided Transformer (FGT) paradigm. FGT serves the purpose of guiding the transformation from a noisy input mesh to its denoised representation.
     
    \item Noise of varying intensity and types can be robustly denoised by our method. Moreover, the proposed method successfully preserves high frequency features and obtains the best result in terms of minimizing deviation in facet normals.

    \item The architecture and loss functions proposed by us also work on meshes with non-manifold topology.
\end{enumerate}



\section{Related Works}
\label{sec:RelatedWorks}

\noindent \textit{Isotropic and Anisotropic Methods}. Initial works on mesh denoising such as \cite{taubin1995signal,vollmer1999improved,desbrun1999implicit, meyer2003discrete} applied mesh filters to remove noise. These methods were isotropic in nature as they did not depend on the geometry of the surface. As a result, these methods along with removing noise also blurred the high frequency details present in the mesh. In order to overcome these problems, later works have adopted anisotropic frameworks. Earlier works in this direction such as \cite{bajaj2003anisotropic,clarenz2000anisotropic,el2008global,ohtake2000polyhedral,desbrun2000anisotropic,tsuchie2012surface,huang2008fast,huang2008surface} were based on anisotropic geometric diffusion, a method inspired from \cite{perona1990scale}.

\noindent \textit{Bilateral Filters}. Another direction of work is based on the use of bilateral filters in meshes either in the vertex domain  \cite{fleishman2003bilateral,jones2003non} or in the facet normal domain \cite{zheng2010bilateral,lee2005feature,solomon2014general}. In \cite{fleishman2003bilateral}, the vertices of the mesh are filtered in the normal direction by making use of the local neighbourhoods. In \cite{jones2003non}, a robust estimator is used to update the position of each vertex by aggregating the predictions of its spatial neighbourhood.

\noindent \textit{Multi-Step Schemes and Variants}. The later anisotropic based works such as \cite{taubin2001linear,ohtake2001mesh,yagou2002mesh,yagou2003mesh,shen2004fuzzy,sun2007fast,sun2008random,lee2005feature,zheng2010bilateral,yadav2018robust} adopted a multi-step scheme, in which, for each face of the mesh, the normals are updated by averaging over the neighbouring normals and then the vertices are updated to be consistent with the new normals. Further, methods such as \cite{fan2009robust,wei2014bi,zhu2013coarse,wang2012cascaded,lu2015robust,centin2017mesh,wei2016tensor,zhang2015guided,li2019feature} have augmented the multi-step scheme with additional steps like adding feature detection and sub-neighbourhood searching. In \cite{zhang2015guided}, a two-stage scheme is used where the first step of normal filtering is guided by a normal field.

\noindent \textit{Sparse Optimization}. These methods are based on the idea that the occurrence of sharp features in a mesh are sparse. Methods such as \cite{wang2014decoupling} and \cite{wu2015mesh} perform ${L_1}$ optimization, \cite{he2013mesh} uses ${L_0}$ norm for surface curvature minimization, and \cite{zhang2015variational} performs variational denoising using total variation regularizer along with piecewise constant function spaces. 

\noindent \textit{Surface Reconstruction}. The objective of methods such as \cite{fleishman2005robust,sheung2009robust,huang2013edge,remil2017surface} is to perform surface reconstruction which is a task closely related to mesh denoising. These methods employ robust statistics to predict the surface normals.

\noindent \textit{Non-Local Similarity and Low Rank Matrix Recovery}. Methods based on non-local similarity \cite{zheng2010non,rosman2013patch,yoshizawa2006smoothing,dong2008level} and low rank matrix recovery \cite{xie2017implicit,xiao2015robust,wei2018mesh,wang2019data} use the idea that there is redundancy in the patches of a mesh with similar looking patches being present in different regions. These methods further construct a patch matrix and recover its low rank. 

\noindent \textit{Learning based}. These include methods such as \cite{8972221,diebel2006bayesian,wang2016mesh,zhao2019normalnet,weimesh,arvanitis2019feature, li2020dnf, armando2020mesh, li2020dnf}. In \cite{wang2016mesh}, the authors have proposed the filtered facet normal descriptor (FND) and have used multiple iterations of neural network which maps the FND to the facet normals of the denoised mesh. 

\begin{figure*}
\begin{center}
  \includegraphics[width=0.9\linewidth]{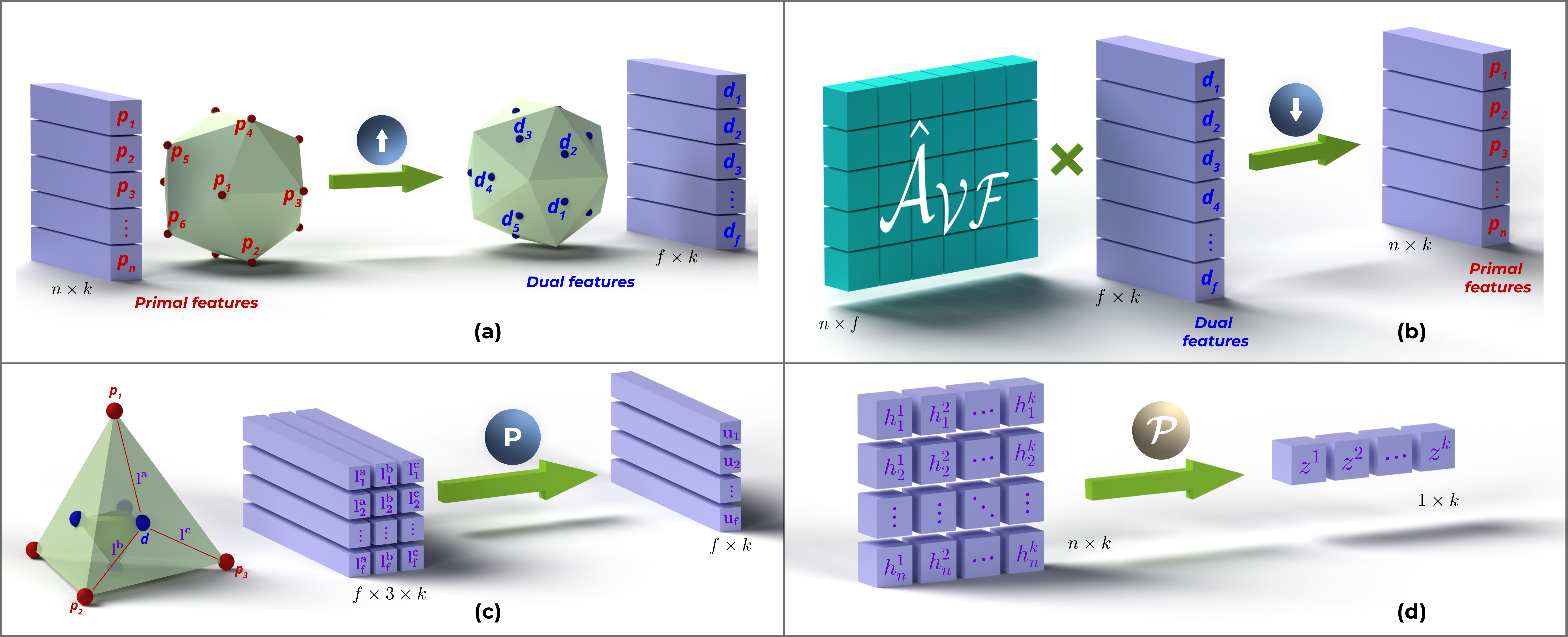}
  \caption{The proposed DMD-Net architecture for mesh denoising: (a) Primal to Dual (P2D) layer, (b) Dual to Primal (D2P) layer, (c) Dual Average Pooling (DAP) layer, and  (d) Feature Average Pooling (FAP) layer.}
  \vspace{-0.5cm}
  \label{fig:DMD-Net3D}
\end{center}
\end{figure*}

\section{DMD-Net: Deep Mesh Denoising Network}
In this section, first we introduce the mesh denoising problem. Then, we describe the structure of DMD-Net and the motivation behind it. Lastly, we mention the loss functions used for training DMD-Net.
\subsection{Problem Statement}
Consider a graph $\mathcal{G} = (\mathcal{V},\mathcal{E},\mathcal{F},\mathcal{P})$, in which $\mathcal{V}$ is the set of vertices, $\mathcal{E} \subseteq \mathcal{V}^2$ is the set of edges, $\mathcal{F} \subseteq \mathcal{V}^3$ is the set of faces, and $\mathcal{P}$ is the vertex-wise feature matrix. Let $|\mathcal{V}|=n$, $|\mathcal{E}|=m$, $|\mathcal{F}|=f$, and $\mathcal{P}$ is a matrix of size $n \times 3$, corresponding to the 3D spatial coordinates of the vertices. Now, suppose that due to a process of signal corruption, some noise gets introduced into the feature matrix of the graph. Let the graph obtained as a result of noise corruption be denoted as $\mathcal{G'} = (\mathcal{V},\mathcal{E},\mathcal{F},\mathcal{P'})$, where $\mathcal{P'}=\mathcal{P} + \mathcal{N}$, in which $\mathcal{N}$ is the noise matrix coming from some noise distribution. Note that both $\mathcal{G}$ and $\mathcal{G'}$ share the same edge and face connectivity. Recovering $\mathcal{G}$ from $\mathcal{G'}$ is the major goal of graph denoising. Since, the graphs under consideration are 3D triangulated meshes, graph denoising in this case will be referred to as mesh denoising.
\subsection{Network Architecture}

 Given a graph $\mathcal{G} = (\mathcal{V},\mathcal{E},\mathcal{F},\mathcal{P})$, let $\mathcal{A_V}$, $\mathcal{A_F}$, and $\mathcal{A_{VF}}$ denote the vertex adjacency matrix, the face adjacency matrix, and the vertex-face adjacency matrix, respectively.
DMD-Net is a Graph-CNN that is designed to solve the problem of mesh denoising. It performs graph aggregation in both the primal graph $\mathcal{G_V} = (\mathcal{V},\mathcal{A_V},\mathcal{X_V})$ as well as the dual graph $\mathcal{G_F} = (\mathcal{F},\mathcal{A_F},\mathcal{X_F})$, where $\mathcal{X_V}$ and $\mathcal{X_F}$ are the primal and the dual features, respectively. Here,  $\mathcal{G_V}$ and $\mathcal{G_F}$ are defined as triplets for the sake of convenience. A brief audio-visual explanation of DMD-Net is included in the supplementary video.

DMD-Net\footnote[1]{A detailed description of network architecture (Feature extractor, Transformer, Denoiser), FGT paradigm, and proposed layers (P2D, D2P, DAP, FAP) is included in the supplementary material. We also present exhaustive ablation studies on the importance of each block and the final choice of DMD-Net architecture in the supplementary material.} is based on the Feature Guided Transformer (FGT) paradigm and consists of three main components: the feature extractor, the transformer, and the denoiser as shown in Figure \ref{fig:DMD-Net}(a). We train the feature extractor to estimate normal vector, mean curvature, and Gaussian curvature for each vertex of the original mesh from the given noisy mesh. These estimated local features serve as guidance for the transformer to compute a transformation matrix $\mathcal{W}_{tf}$. 

The feature extractor (Figure \ref{fig:DMD-Net}(b)) internally contains a pair of two-stream networks which have two parallel streams, the upper one called the dual stream and the lower one called the primal stream. 
The transformer network consists of different layers as shown in Figure \ref{fig:DMD-Net}(c). We pool the final features using feature average-pooling layer (FAP) (Figure \ref{fig:DMD-Net3D}(d)) to learn the transformation $\mathcal{W}_{tf}$. The noisy input mesh is concatenated with the local features extracted by feature extractor network, on which the transformation $\mathcal{W}_{tf}$ is applied. These transformed features guide the denoiser network to accurately estimate a denoised representation of the input noisy mesh. The denoiser has a structure identical to that of the feature extractor as depicted in Figure \ref{fig:DMD-Net}(d).

The two-stream network is an asymmetric module consisting of two parallel streams, the lower one for performing aggregation in the primal graph and the upper one for performing aggregation in the dual graph. It consists of the primal-to-dual layer, a cascade of aggregator layers and a primal dual fusion layer. The primal-to-dual layer (Figure \ref{fig:DMD-Net3D}(a)) converts the primal graph features $\mathcal{X_V}$ into the dual graph features $\mathcal{X_F}$. In dual graph representation, we represent the feature of each face as the centroid of the features of the vertices constituting that face. The Aggregator (AGG) (Figure \ref{fig:DMD-Net}(f)) performs graph aggregation by pooling in the features of the neighbouring nodes. We use the graph aggregation formulation as described in \cite{kipf2016semi}. 
The input to AGG is $\mathcal{X}$ (feature matrix) and $\mathcal{A}$ (adjacency matrix). The input graph to AGG can be in both forms, primal as well as dual. The output of AGG is given by
$g(\mathcal{X},\mathcal{A}) = \sigma(\hat{\mathcal{D}}^{-\frac{1}{2}}\hat{\mathcal{A}}\hat{\mathcal{D}}^{-\frac{1}{2}}\mathcal{X}\mathcal{W})$.
Here, $\sigma$ is the ReLU activation function, $\mathcal{W}$ is the learnable weight matrix, $\hat{\mathcal{A}} = \mathcal{A} + \mathcal{I}$ ($\mathcal{I}$ being the identity matrix), and $\hat{\mathcal{D}}$ is the diagonal node degree matrix of $\hat{\mathcal{A}}$.
The primal dual fusion (PDF) block (Figure \ref{fig:DMD-Net}(f)) fuses the aggregated features from both the streams at the facet level. This fusion serves as a point of communication between the primal-dual streams, allowing flow of information from one stream to the other. PDF employs a dual average pooling layer that intermixes the primal and dual stream features as shown in (Figure \ref{fig:DMD-Net3D}(c)). The fused feature representation after dual-average pooling is in dual form $\mathcal{X_F}$, and is converted to primal form $\mathcal{X_V}$ using dual-to-primal layer (Figure \ref{fig:DMD-Net3D}(b)). We obtain $\mathcal{X_V}$ by pre-multiplying $\mathcal{X_F}$ with the degree normalized vertex-face adjacency matrix $\mathcal{\hat{A}_{VF}}$.

\begin{figure*}[htb]
    \begin{center}
    \captionsetup[subfigure]{labelformat=empty}
    \subfloat[]{\includegraphics[width=0.9\linewidth ]{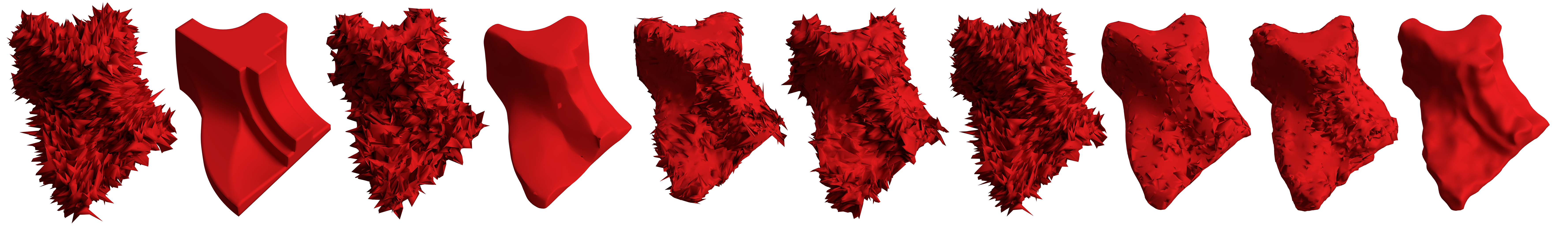}}\\
    \vspace{-0.8cm}
    \subfloat[]{\includegraphics[width=0.9\linewidth ]{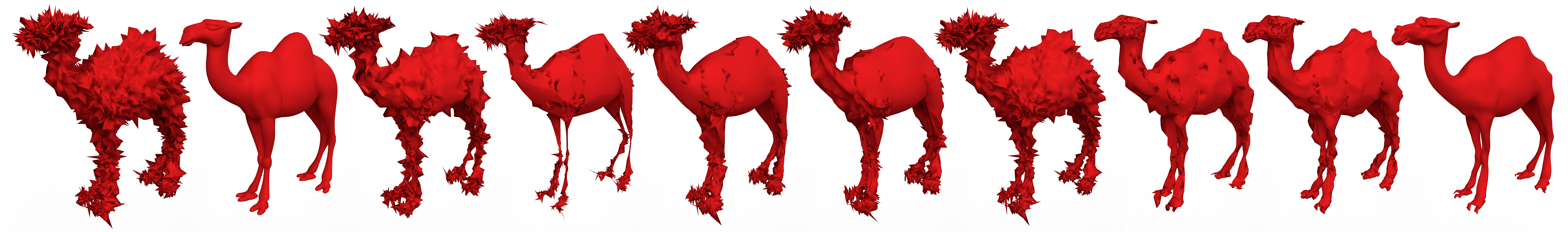}}\\
    \vspace{-0.8cm}
    \subfloat[Noisy]{\includegraphics[width=0.09\linewidth ]{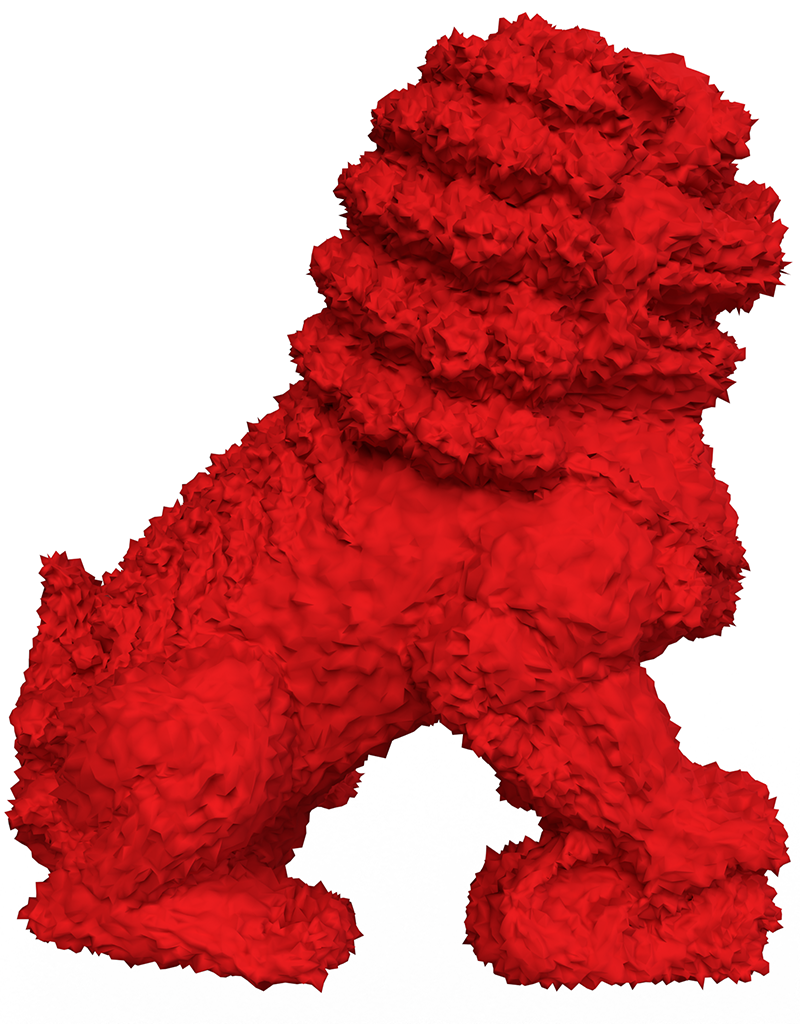}}
    \subfloat[Original]{\includegraphics[width=0.09\linewidth ]{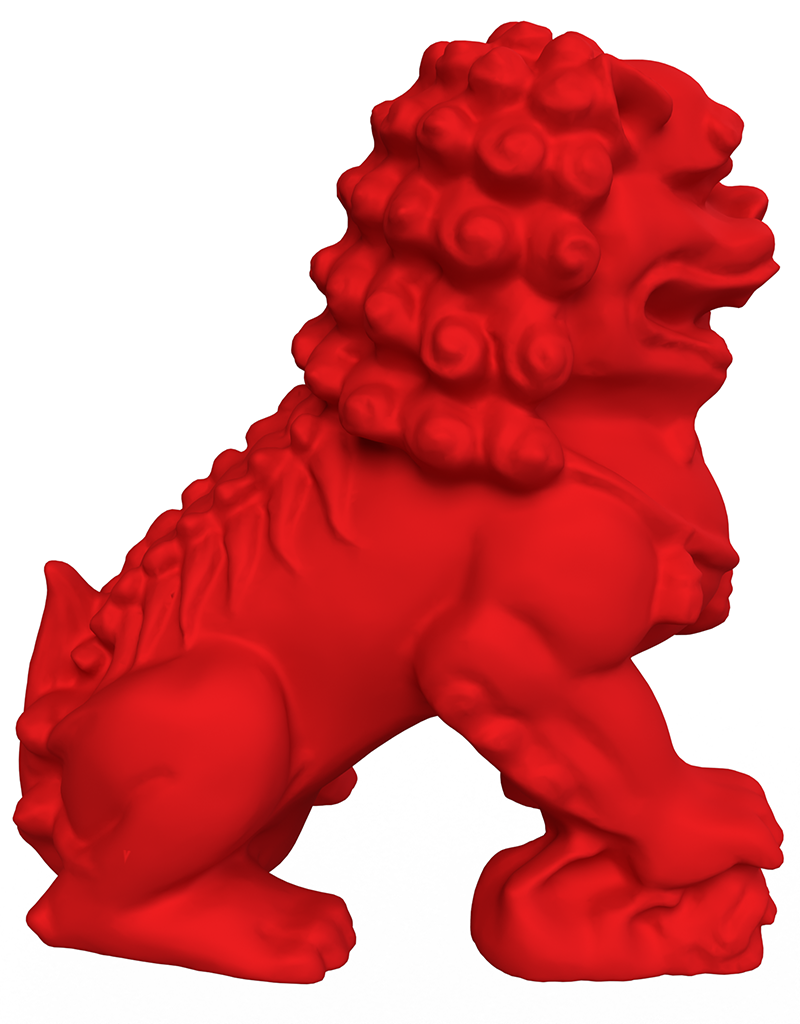}}
    \subfloat[BMD\cite{fleishman2003bilateral}]{\includegraphics[width=0.09\linewidth ]{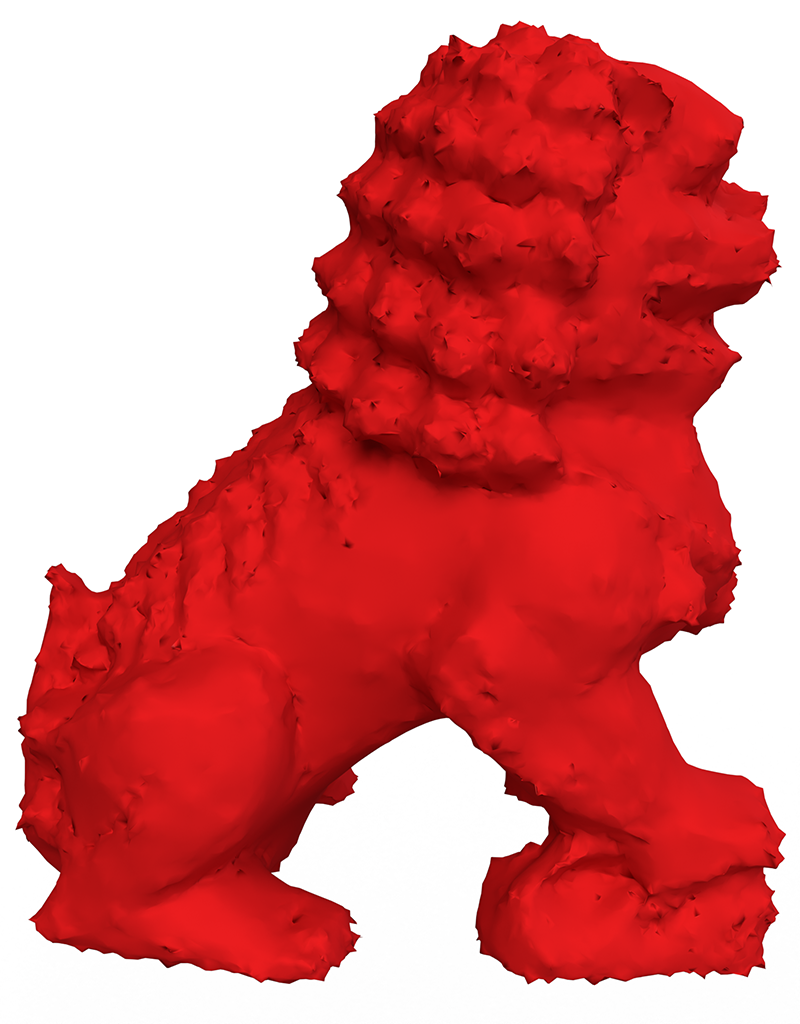}}
    \subfloat[L0M\cite{he2013mesh}]{\includegraphics[width=0.09\linewidth ]{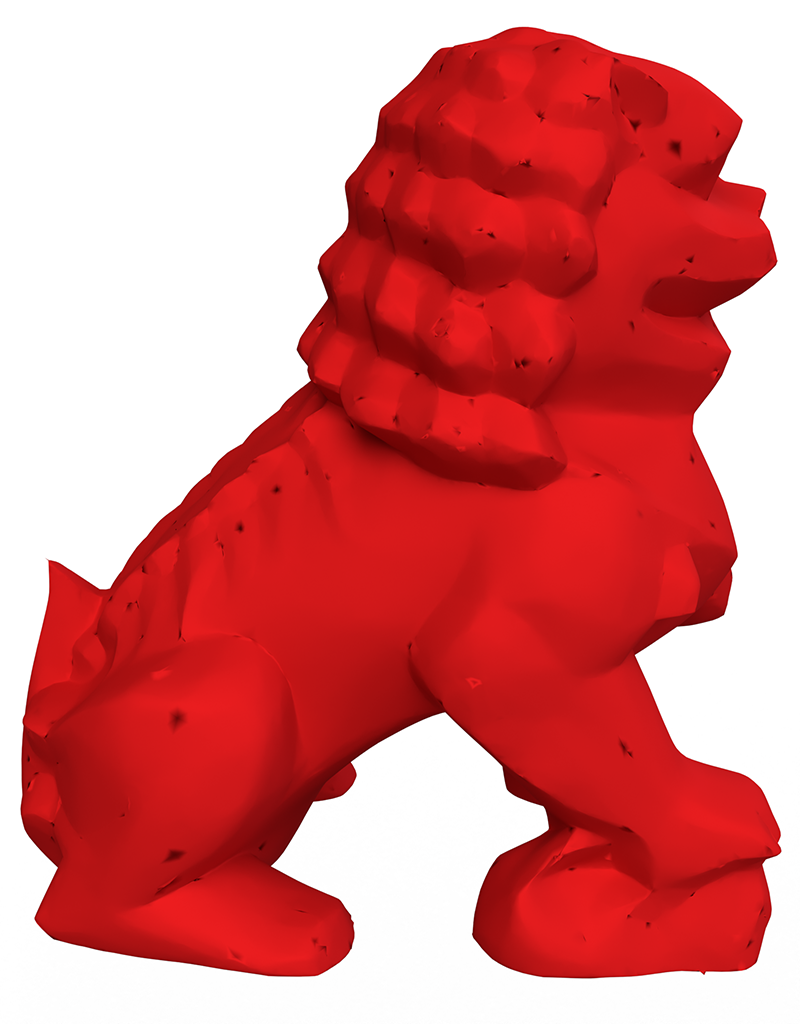}}
    \subfloat[BNF\cite{zheng2010bilateral}]{\includegraphics[width=0.09\linewidth ]{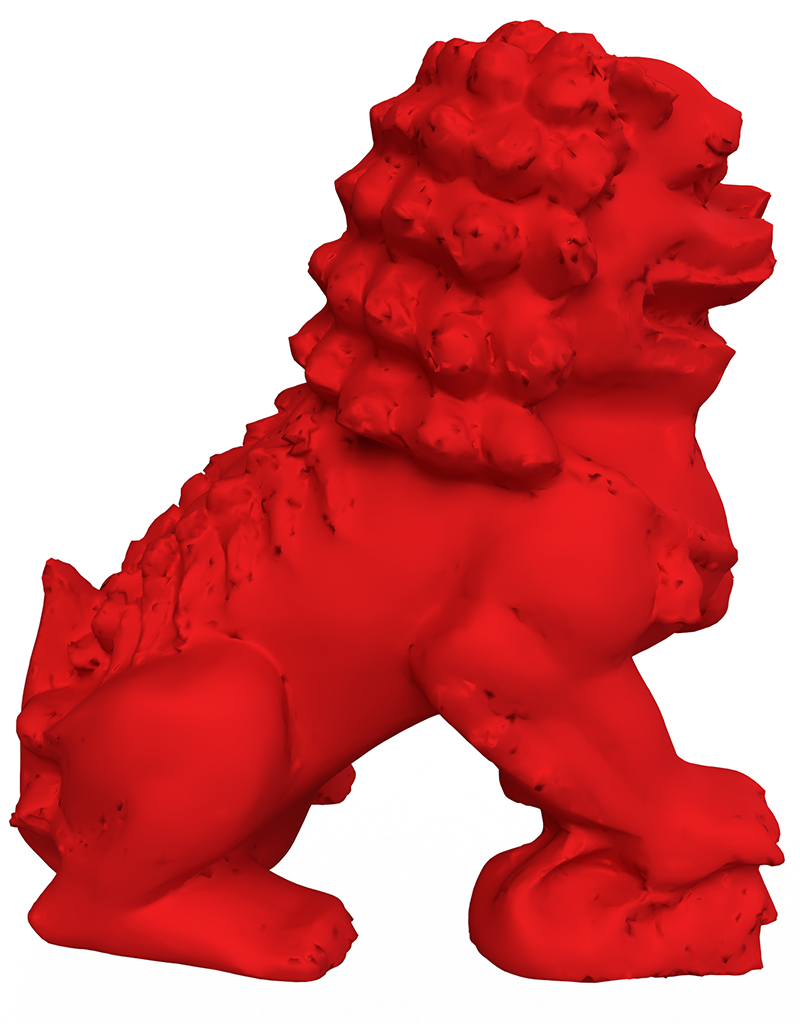}}
    \subfloat[FEFP\cite{sun2007fast}]{\includegraphics[width=0.09\linewidth ]{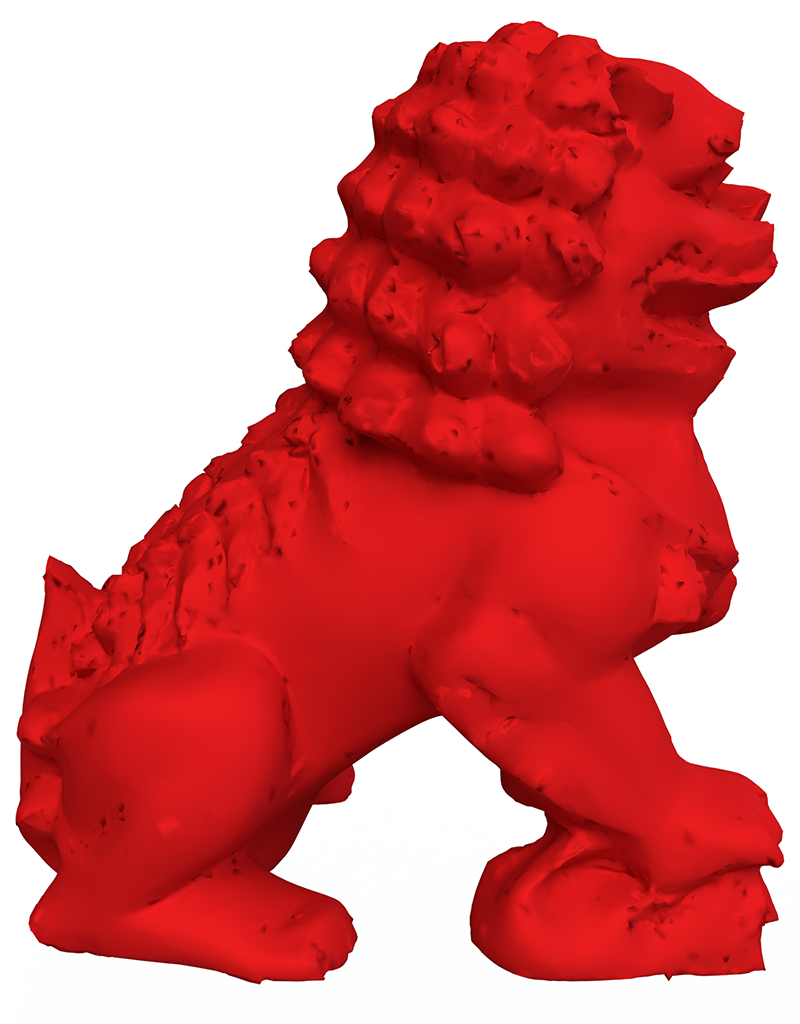}}
    \subfloat[NIFP\cite{jones2003non}]{\includegraphics[width=0.09\linewidth ]{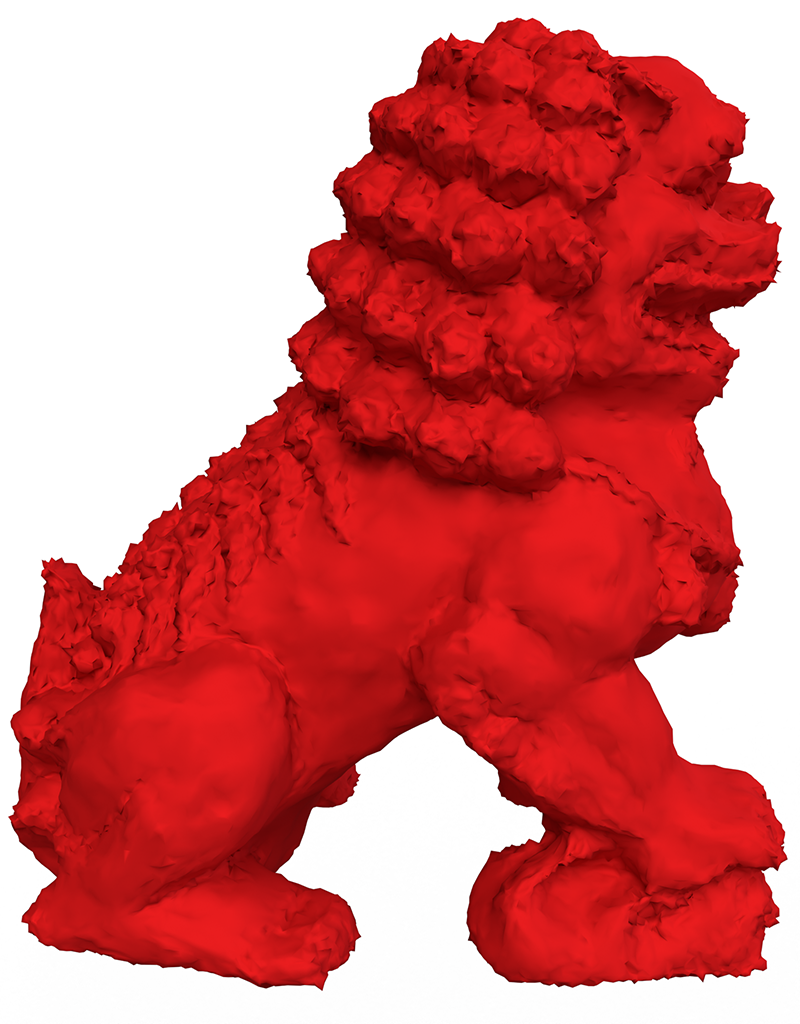}}
    \subfloat[GNF\cite{zhang2015guided}]{\includegraphics[width=0.09\linewidth ]{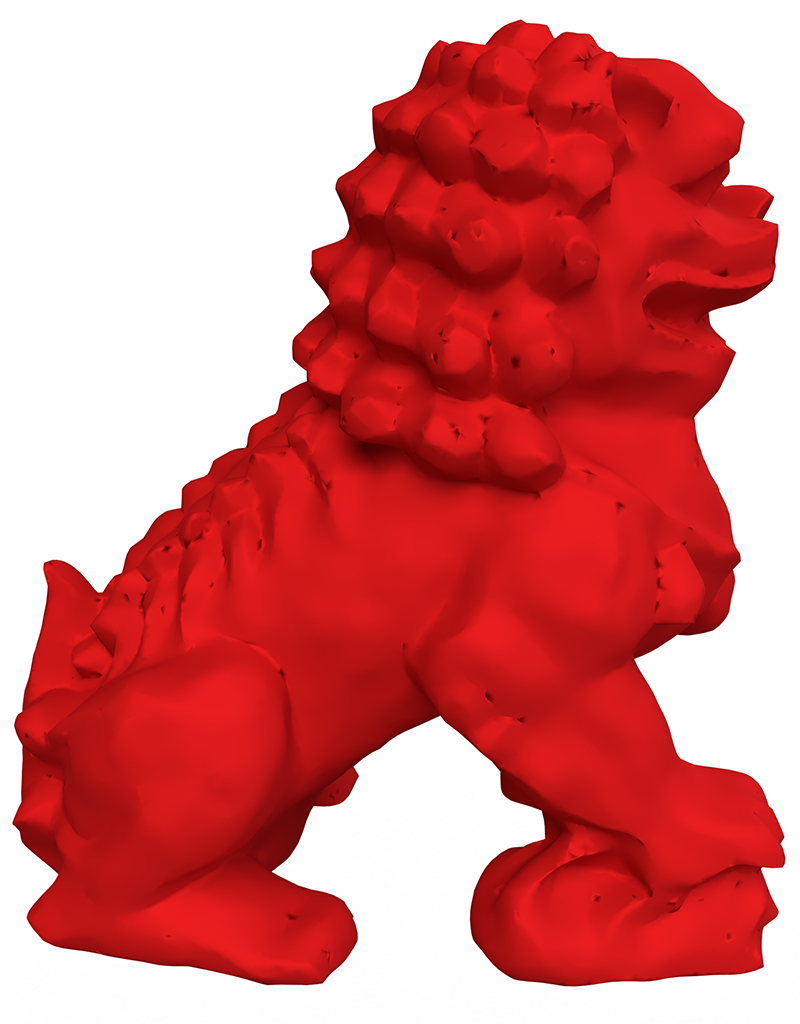}}
    \subfloat[CNR\cite{wang2016mesh}]{\includegraphics[width=0.09\linewidth ]{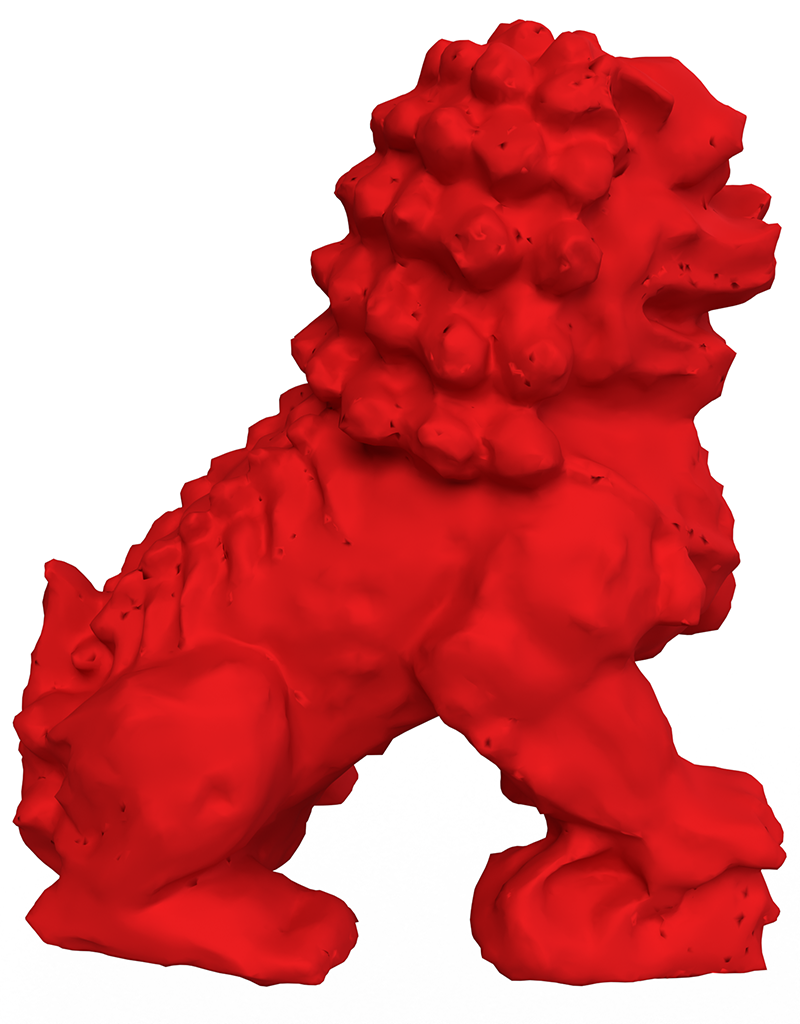}}
    \subfloat[Ours]{\includegraphics[width=0.09\linewidth ]{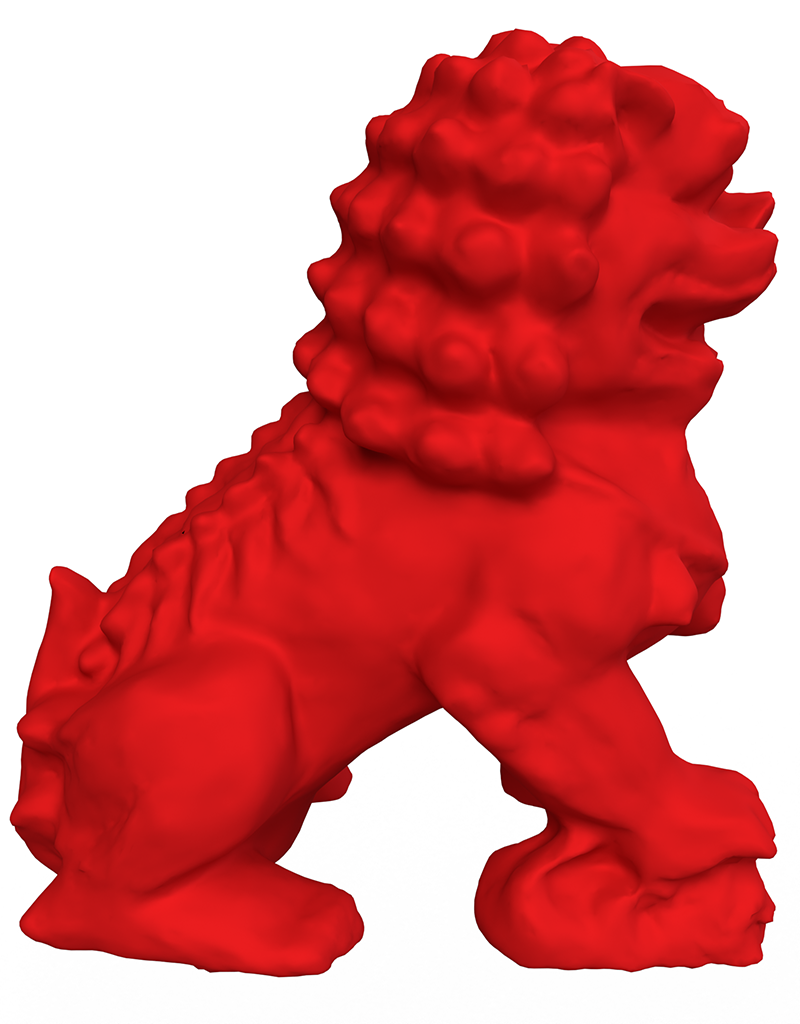}}
    \caption{Comparison of different mesh denoising methods on Fandisk (Noise: Gaussian $\mathcal{N}(0,0.05)$), Camel (Noise: Gaussian $\mathcal{N}(0,0.025)$) and Chineese Lion (Noise: Gaussian $\mathcal{N}(0,0.005)$).}
    \vspace{-0.3cm}
    \label{fig:comparison_SOTA1}
    \end{center}
\end{figure*}

\begin{table*}[t]
\begin{center}
\renewcommand{\arraystretch}{1.3}
\caption{Metric comparison of our method with state-of-the-art mesh denoising algorithms for six different meshes}
\label{table:SOTA}
\footnotesize
\begin{adjustbox}{width=\linewidth,center}
\begin{tabular}{|c|c|c|l|l|c|c|c|c|c|c|c|c|}
\hline
\textbf{Object}                   & \textbf{Noise}              & \multicolumn{3}{c|}{\textbf{Metric}}                         & \textbf{BMD}\cite{fleishman2003bilateral} & \textbf{L0M}\cite{he2013mesh} & \textbf{BNF}\cite{zheng2010bilateral} & \textbf{FEFP}\cite{sun2007fast} & \textbf{NIFP}\cite{jones2003non} & \textbf{GNF}\cite{zhang2015guided} & \textbf{CNR}\cite{wang2016mesh} & \textbf{Ours} \\ \hline
\multirow{3}{*}{Block } & \multirow{3}{*}{$\mathcal{N}(0,0.05)$} & \multicolumn{3}{c|}{Normal}      & 73.01 & 20.786 & 55.516 & 66.434 & 74.962 & 49.81 & 47.361 & \textbf{20.315}       \\ \cline{3-13} 
                         &                    & \multicolumn{3}{c|}{Vertex $(\times 10^{-3})$}      & 2.173 & 0.231 & 0.713 & 1.172 & 2.215 & 0.393 & 0.402 & \textbf{0.155}       \\ \cline{3-13} 
                         &                    & \multicolumn{3}{c|}{Chamfer}                        & 0.49 & 0.185 & 0.183 & 0.331 & 0.521 & 0.212 & 0.206 & \textbf{0.135}       \\ \hline
\multirow{3}{*}{Bunny  } & \multirow{3}{*}{$0.015 \cdot \Gamma(2,2)$} & \multicolumn{3}{c|}{Normal}     & 20.094 & 8.679 & 24.487 & 33.516 & 28.954 & 18.928 & 11.747 & \textbf{7.769}       \\ \cline{3-13} 
                         &                    & \multicolumn{3}{c|}{Vertex $(\times 10^{-3})$}      & 0.236 & \textbf{0.218} & 0.263 & 0.27 & 0.315 & 0.255 & 0.24 & 0.29       \\ \cline{3-13} 
                         &                    & \multicolumn{3}{c|}{Chamfer}                       & \textbf{0.193} & 0.216 & 0.207 & 0.195 & 0.252 & 0.226 & 0.228 & 0.258       \\ \hline
            
            
\multirow{3}{*}{Fertility } & \multirow{3}{*}{$0.035 \cdot \mathcal{U}(-1,1)$} & \multicolumn{3}{c|}{Normal}  & 60.853 & 49.574 & 30.182 & 37.627 & 65.037 & 33.676 & 24.268 & \textbf{11.285}       \\ \cline{3-13} 
                         &                    & \multicolumn{3}{c|}{Vertex $(\times 10^{-3})$}     & 0.294 & 0.209 & 0.07 & 0.135 & 0.259 & 0.085 & \textbf{0.054} & 0.063       \\ \cline{3-13} 
                         &                    & \multicolumn{3}{c|}{Chamfer}                       & 0.142 & 0.093 & 0.045 & 0.064 & 0.121 & 0.061 & \textbf{0.043} & 0.048       \\ \hline
            
\multirow{3}{*}{Chinese Lion} &\multirow{3}{*}{$\mathcal{N}(0,0.005)$} & \multicolumn{3}{c|}{Normal} & 20.505 & 16.864 & 16.19 & 17.581 & 29.482 & 14.547 & 15.068 & \textbf{12.881}      \\ \cline{3-13} 
                         &                    & \multicolumn{3}{c|}{Vertex $(\times 10^{-3})$}      & 0.014 & 0.02 & 0.007 & 0.008 & 0.009 & 0.007 & \textbf{0.006} & 0.039      \\ \cline{3-13} 
                         &                    & \multicolumn{3}{c|}{Chamfer}                        & 0.012 & 0.019 & 0.005 & 0.007 & 0.008 & 0.006 & \textbf{0.005} & 0.023      \\ \hline
            
\multirow{3}{*}{Camel  } & \multirow{3}{*}{$\mathcal{N}(0,0.025)$} & \multicolumn{3}{c|}{Normal}     & 71.597 & 65.904 & 60.208 & 64.855 & 73.529 & 49.176 & 51.623 & \textbf{24.726}       \\ \cline{3-13} 
                         &                    & \multicolumn{3}{c|}{Vertex $(\times 10^{-3})$}      & 0.519 & 0.393 & 0.288 & 0.313 & 0.449 & 0.132 & 0.114 &\textbf{ 0.108}    \\ \cline{3-13} 
                         &                    & \multicolumn{3}{c|}{Chamfer}                       & 0.13 & 0.154 & 0.074 & 0.089 & 0.115 & 0.077 & \textbf{0.059} & 0.061      \\ \hline
            
\multirow{3}{*}{Fan Disk  } & \multirow{3}{*}{$\mathcal{N}(0,0.05)$} & \multicolumn{3}{c|}{Normal}  & 72.47 & \textbf{17.249} & 47.726 & 58.636 & 73.392 & 40.586 & 42.508 & 19.82      \\ \cline{3-13} 
                         &                    & \multicolumn{3}{c|}{Vertex $(\times 10^{-3})$}      & 2.166 & 0.213 & 0.656 & 0.954 & 2.064 & 0.398 & 0.414 & \textbf{0.176}      \\ \cline{3-13} 
                         &                    & \multicolumn{3}{c|}{Chamfer}                        & 0.538 & 0.181 & 0.195 & 0.298 & 0.529 & 0.233 & 0.207 &\textbf{ 0.146}      \\ \hline
            
\end{tabular}
\end{adjustbox}
\vspace{-0.5cm}
\end{center}

\end{table*}


\subsection{Loss Functions}
\label{sec:LossFunctions}
\renewcommand{\thefootnote}{\fnsymbol{footnote}}

To train our network, we use the following loss functions\footnote[2]{A detailed description and mathematical formulation of all the loss functions is included in supplementary material. We also present ablation studies on the importance and choice of training weights for each loss function in the supplementary material.}: (a) \textbf{Vertex loss} ($\mathcal{L}_{vertex}$) - which computes the mean Euclidean distance between the corresponding vertices, (b) \textbf{Normal loss} ($\mathcal{L}_{normal}$) - which computes the mean angular deviation between the normals of the corresponding faces, (c) \textbf{Curvature loss} ($\mathcal{L}_{curvature}$) - where we compute the mean curvature error and the Gaussian curvature error of the vertices \cite{meyer2003discrete}, (d) \textbf{Chamfer loss} \cite{barrow1977parametric} ($\mathcal{L}_{chamfer}$), and (e) \textbf{Feature Extractor loss} ($\mathcal{L}_{FE}$). 
During training, we use a linear combination of the loss functions described above. The weights used for vertex loss, normal loss, curvature loss, chamfer loss and the feature extractor loss are $\lambda_{V}=1$, $\lambda_{N}=0.2$, $\lambda_{\kappa}=0.01$, $\lambda_{C}=0.05$ and  $\lambda_{FE}=1$, respectively.


\begin{table}[t]
\begin{center}
\renewcommand{\arraystretch}{1.3}
\caption{Performance of DMD-Net on different noise types and levels}
\vspace{-0.2cm}
\label{table:Noise_Evaluation}
\footnotesize
\begin{adjustbox}{width=\linewidth,center}
{
\begin{tabular}{|c|l|c|c|c|c|}
\hline
\multicolumn{2}{|c|}{\multirow{3}{*}{\textbf{Noise}}} & \multicolumn{4}{c|}{\textbf{test-intra + test-inter}} \\ \cline{3-6} 
\multicolumn{2}{|c|}{} & \multicolumn{2}{c|}{\begin{tabular}[c]{@{}c@{}}Vertex $(\times 10^{-4})$\end{tabular}} & \multicolumn{2}{c|}{\begin{tabular}[c]{@{}c@{}}Normal (degrees)\end{tabular}} \\ \cline{3-6} 
\multicolumn{2}{|c|}{} & Loss & Reference & Loss & Reference \\ \hline
\multirow{3}{*}{Gamma noise} & $0.03 \cdot \Gamma(2,2)$                     & $3.995$ & $13.345$ & $33.239$ & $72.186$ 
  \\ \cline{2-6} 
 & $0.04 \cdot \Gamma(2,2)$                                             	& $4.699$ & $23.724$ & $37.331$ & $75.908$ 
  \\ \cline{2-6} 
 & $0.05 \cdot \Gamma(2,2)$                                             	& $5.587$ & $37.069$ & $40.88$ & $78.39$
  \\ \hline
\multirow{3}{*}{Gaussian Noise} & $ \mathcal{N}(0,0.025)$              	    & $3.285$ & $6.176$ & $26.692$ & $67.482$
  \\ \cline{2-6} 
 & $ \mathcal{N}(0,0.05)$                                               	& $3.952$ & $24.705$ & $34.372$ & $77.088$
  \\ \cline{2-6} 
 & $\mathcal{N}(0,0.1)$                                                 	& $6.478$ & $98.82$ & $45.334$ & $83.268$
  \\ \hline
\multirow{3}{*}{Impulse Noise} & $0.05 \cdot \delta(0.15,0.15)$         	& $3.333$ & $7.408$ & $27.424$ & $47.756$
  \\ \cline{2-6} 
 & $0.1 \cdot \delta(0.15,0.15)$                                        	& $4.118$ & $29.633$ & $35.377$ & $52.65$
  \\ \cline{2-6} 
 & $0.2 \cdot \delta(0.15,0.15)$                                        	& $7.074$ & $118.531$ & $46.306$ & $55.351$
  \\ \hline
\multirow{3}{*}{Uniform noise} & $0.05 \cdot \mathcal{U}(-1,1)$        	    & $3.366$ & $8.233$ & $28.158$ & $71.154$
  \\ \cline{2-6} 
 & $0.1 \cdot \mathcal{U}(-1,1)$                                       	    & $4.254$ & $32.932$ & $36.677$ & $79.704$
  \\ \cline{2-6} 
 & $0.2 \cdot \mathcal{U}(-1,1)$                                        	& $7.698$ & $131.729$ & $48.307$ & $84.799$
 \\ \hline
\end{tabular}
}
\end{adjustbox}
\end{center}
\vspace{-0.5cm}
\end{table}

\section{Results and Discussions}

\subsection{Dataset Generation}
We use ShapeNet \cite{chang2015shapenet}, a large scale repository of 3D models, as our dataset for training the network. ShapeNet has around 50k objects spanning 55 categories. The entire dataset is split into three parts: a) train, b) test-intra, and c) test-inter. For each of the 50 categories, 80\% of the 3D models are included in train and the rest 20\% is included in test-intra. All the objects in the remaining 5 held out classes are included in test-inter. The objective of test-inter dataset is to evaluate how well our network performs on categories it has not seen before. We include a detailed description of data preparation process and creation of noisy and its corresponding ground-truth counterpart in supplementary material. We further augment the data by rotating the mesh in random orientation in every epoch of training.

The final dataset contains $80357$ meshes to which rotation augmentation and noise is added. This dataset is split into train, test-intra and test-inter containing $61842$, $15377$, and $3138$ meshes respectively. During training our network, each epoch consists of $10^4$ meshes which are randomly selected from the training set.

\subsection{Results}


In Table \ref{table:SOTA}, we quantitatively compare our method with several existing mesh denoising algorithms on some of the popular meshes which are not part of the ShapeNet dataset. To each of these meshes we add a different type and level of noise. To compare the different methods we use the following metrics: Normal loss, Vertex loss and Chamfer loss. 
Note that in terms of angular divergence (Normal loss), our method shows significantly better performance compared to other methods.
The resulting denoised mesh obtained using various method are visually compared in Figure \ref{fig:comparison_SOTA1}.
Further, we show the visual results of DMD-Net on nine popular meshes in Figure \ref{fig:teaser}. A low noise of $\mathcal{N}(0,0.01)$ is used for all the nine meshes.


Since CNR\cite{wang2016mesh} uses a learning based framework, we perform the following two comparisons. First we train both DMD-Net and CNR network on the ShapeNet training dataset and compare their performance on test-intra and test-inter sets. We then train both DMD-Net and CNR network on the CNR train dataset and then compare their performance on the CNR test dataset. This comparison is shown in Table \ref{table:DMDvsCNR}.
We show the visual results of the four methods, as mentioned above, in Figure \ref{fig:CNRvsDMD}. We find the results of DMD-Net visually more plausible and close to the ground truth in the visual sense.
Further, in Figure \ref{fig:comparison_real_scans}, we present a visual comparison of several denoising methods on Kinect scans from the CNR dataset, and show that, DMD-Net achieves competitive performance.



\begin{figure}[t]
\vspace{-0.5cm}
\begin{center} 
\captionsetup[subfigure]{labelformat=empty}
\hspace*{\fill}
    \begin{minipage}[t]{.075\textwidth}
        \centering
        \vspace{1.2cm}
        \subfloat[Original]
            {\includegraphics[valign=c,width=1\textwidth]{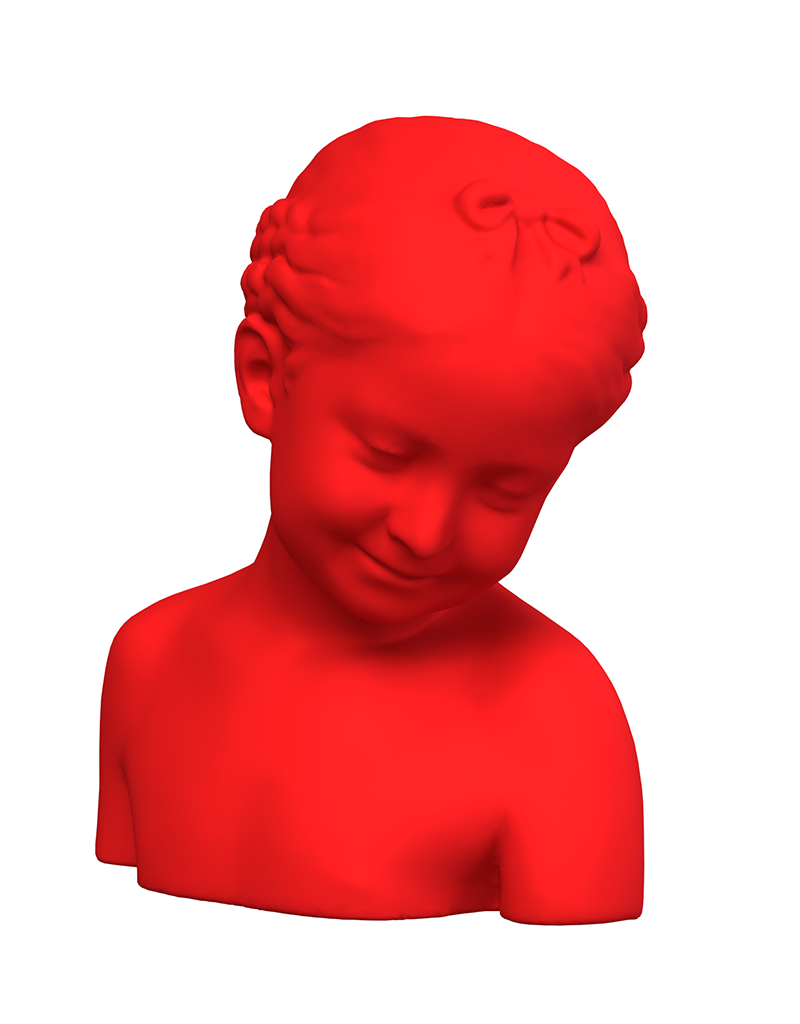}}
    \end{minipage}
    \begin{minipage}[t]{.375\textwidth}
        \centering
        \subfloat[$\mathcal{N}(0,0.005)$]
        {\includegraphics[valign=t,width=0.23\textwidth]{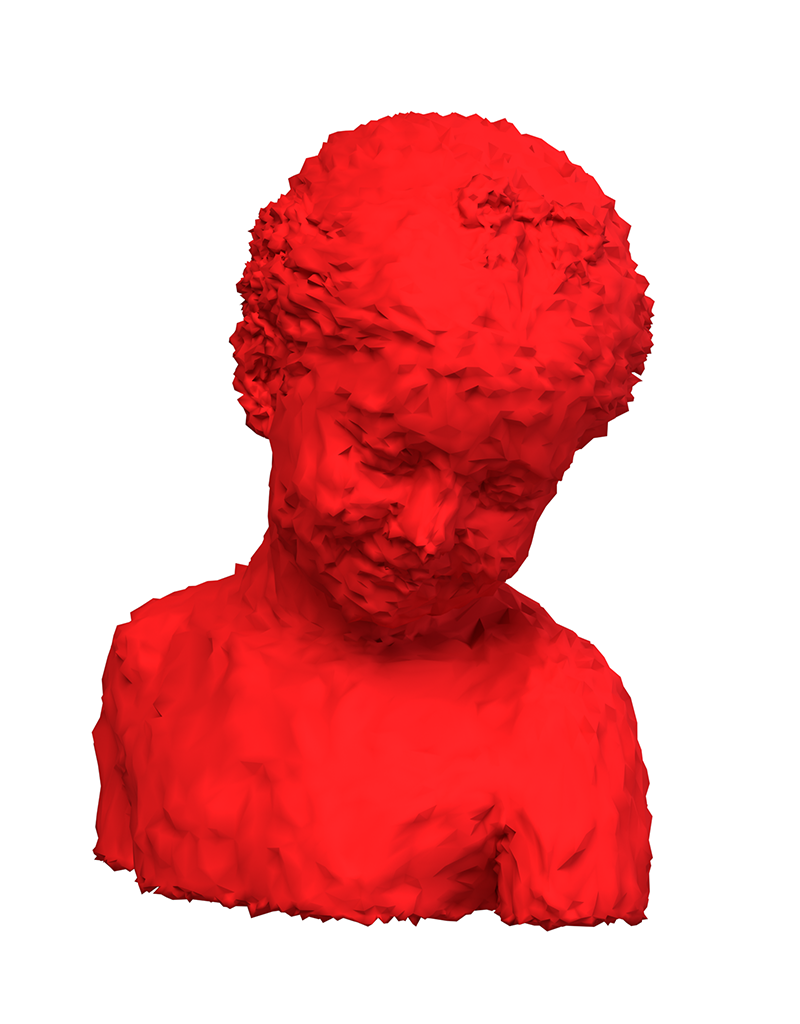}}
        \subfloat[$\mathcal{N}(0,0.01)$]
        {\includegraphics[valign=t,width=0.23\textwidth]{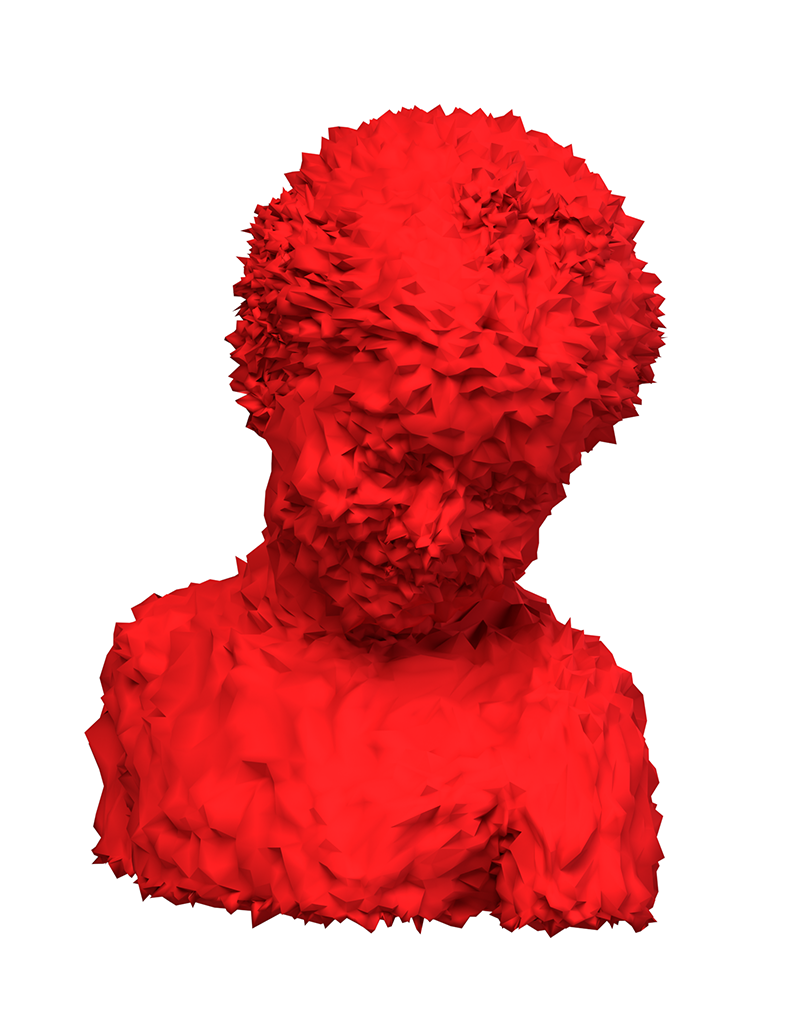}}
        \subfloat[$\mathcal{N}(0,0.03)$]
        {\includegraphics[valign=t,width=0.23\textwidth]{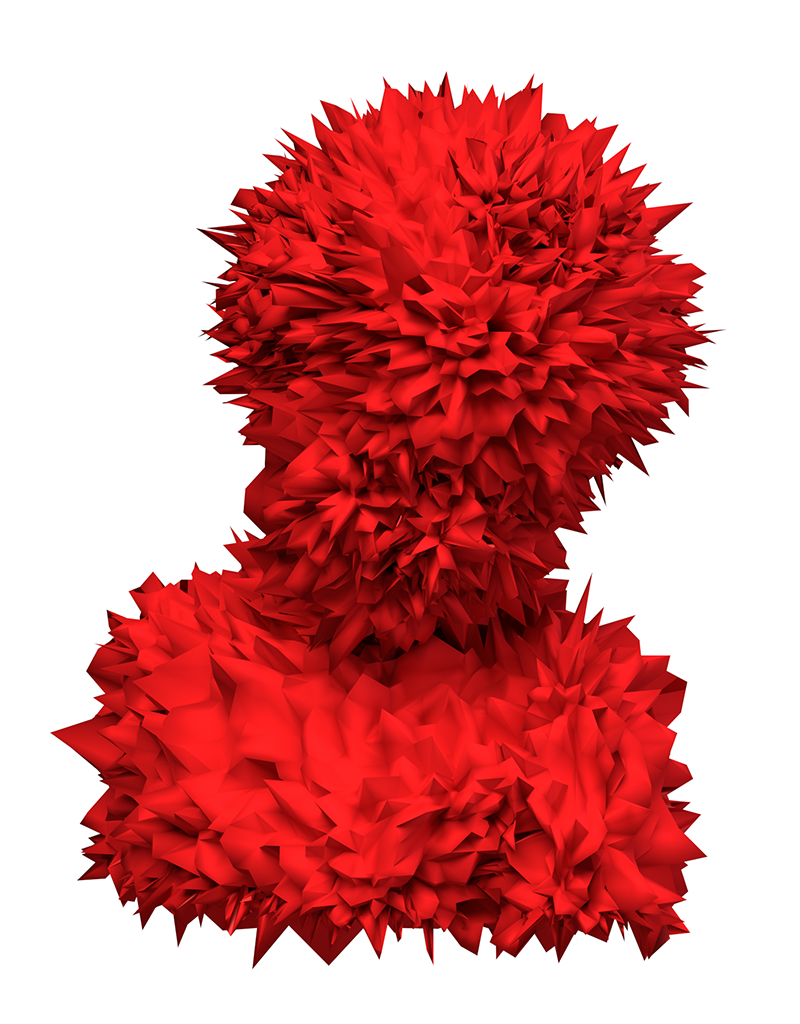}}
        \subfloat[$\mathcal{N}(0,0.05)$]
        {\includegraphics[valign=t,width=0.23\textwidth]{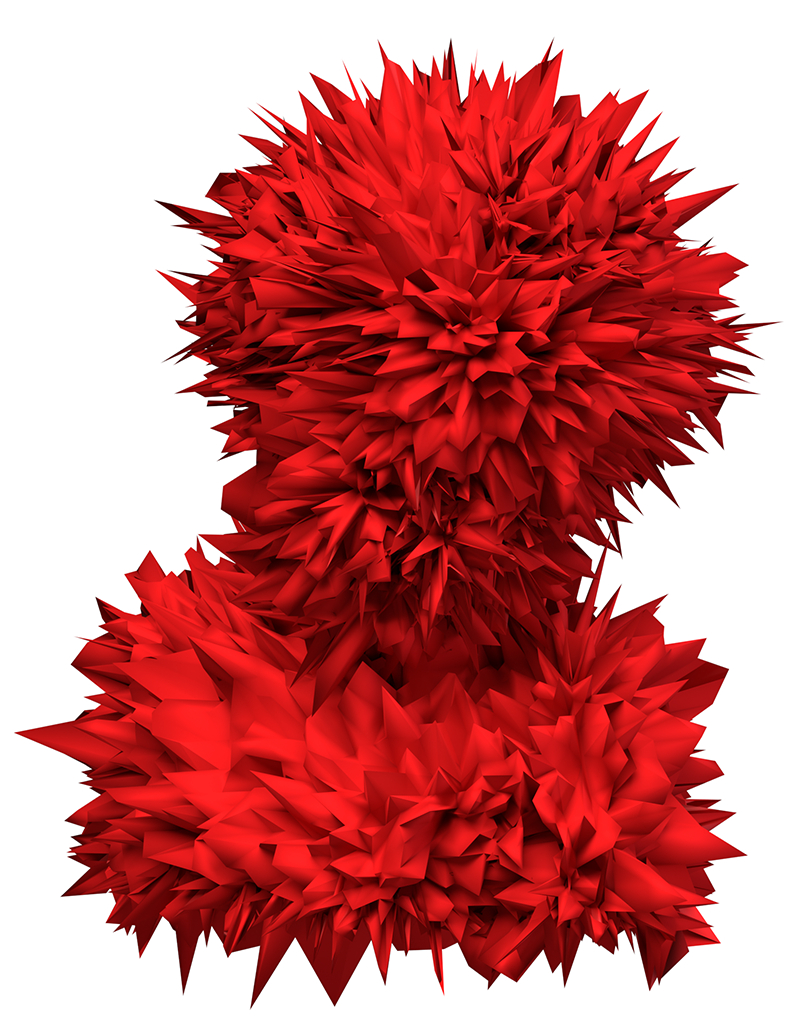}}\\
        \vspace{0cm}
        {\includegraphics[valign=t,width=0.23\textwidth]{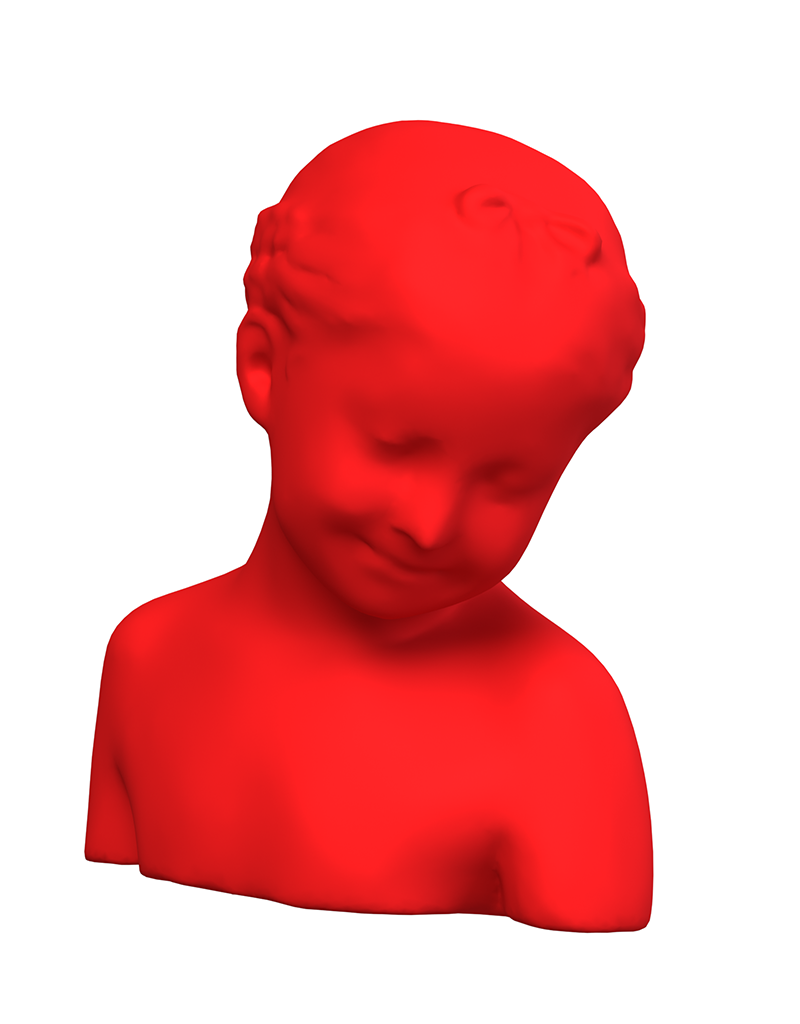}}
        {\includegraphics[valign=t,width=0.23\textwidth]{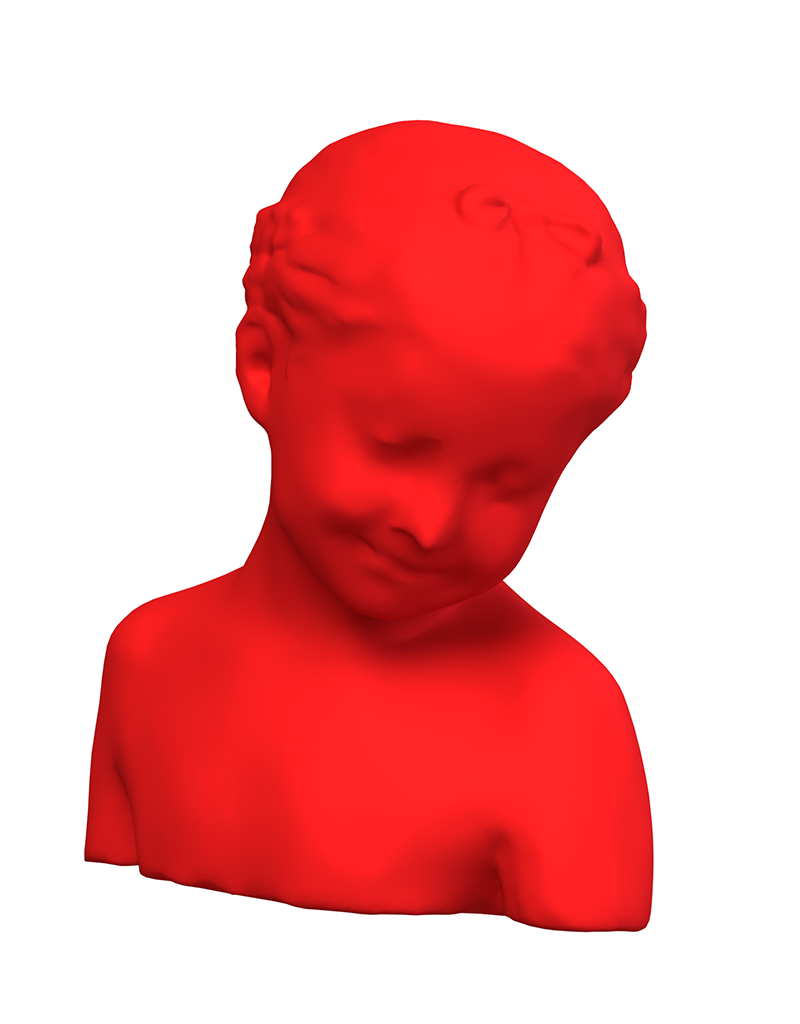}}
        {\includegraphics[valign=t,width=0.23\textwidth]{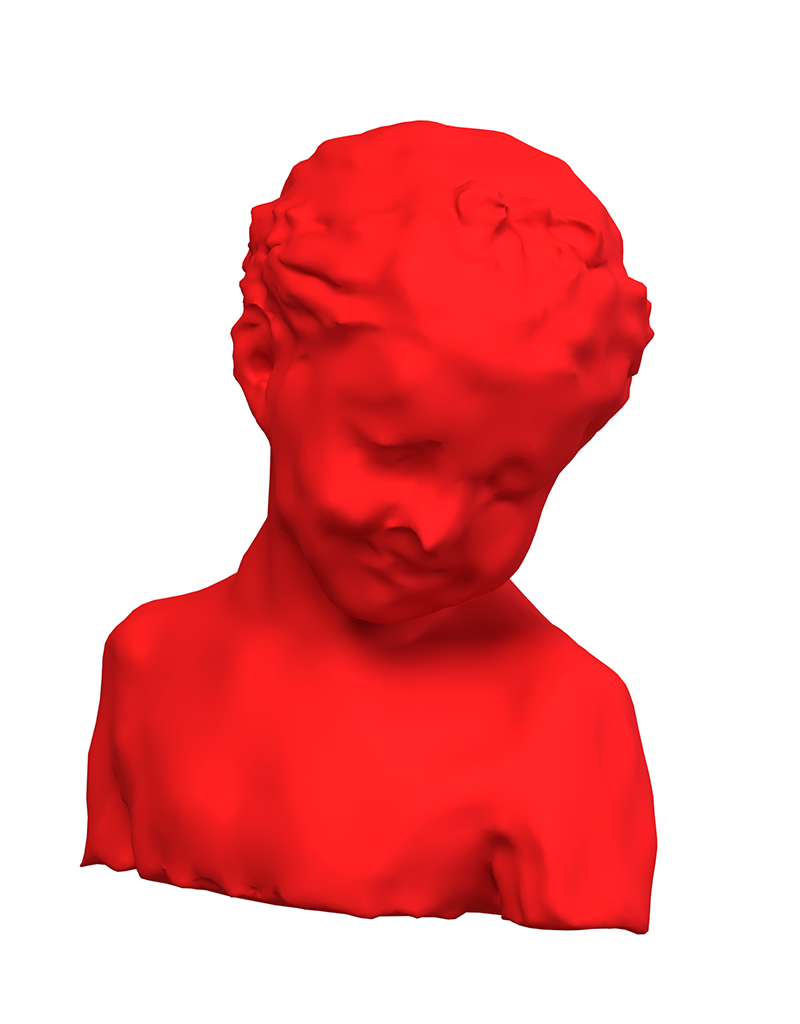}}
        {\includegraphics[valign=t,width=0.23\textwidth]{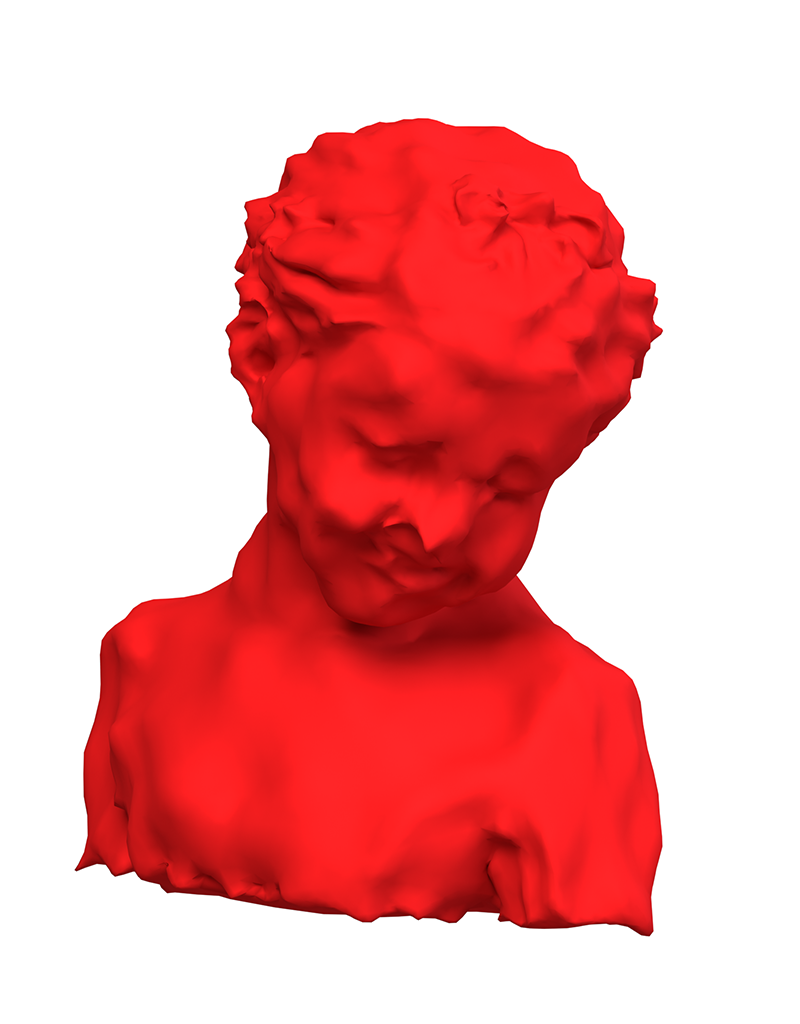}}
    \end{minipage}  
    \caption{Output of DMD-Net on the bust model under Gaussian noise of various levels. The top row denotes the noisy mesh and the bottom row denotes the denoised result of DMD-Net.}
     \label{fig:Mixed_Noise}
     \vspace{-0.5cm}
     \end{center}
\end{figure}


\begin{table*}[!htb]
\renewcommand{\arraystretch}{1.3}
\caption{Comparison of DMD-Net with CNR\cite{wang2016mesh} on both the ShapeNet as well as the CNR dataset}
\vspace{-0.3cm}
\label{table:DMDvsCNR}
\footnotesize
\begin{center}
\begin{tabular}{|c|c|c|c|c|c|c|c|c|c|}
\hline
\multirow{3}{*}{\textbf{Model}}        & \multicolumn{6}{c|}{\textbf{Trained On ShapeNet dataset}}                  & \multicolumn{3}{c|}{\textbf{Trained on CNR dataset}} \\ \cline{2-10} 
                              & \multicolumn{3}{c|}{test-intra ShapeNet} & \multicolumn{3}{c|}{test-inter ShapeNet} & \multicolumn{3}{c|}{test CNR Dataset}                   \\ \cline{2-10} 
 &
  \begin{tabular}[c]{@{}c@{}}Vertex\\ $(\times10^{-4})$\end{tabular} &
  \begin{tabular}[c]{@{}c@{}}Normal\\ (degrees)\end{tabular} &
  \begin{tabular}[c]{@{}c@{}}Chamfer\\ $(\times10^{-4})$\end{tabular} &
  \begin{tabular}[c]{@{}c@{}}Vertex\\ $(\times10^{-4})$\end{tabular} &
  \begin{tabular}[c]{@{}c@{}}Normal\\ (degrees)\end{tabular} &
  \begin{tabular}[c]{@{}c@{}}Chamfer\\ $(\times10^{-4})$\end{tabular} &
  \begin{tabular}[c]{@{}c@{}}Vertex\\ $(\times10^{-4})$\end{tabular} &
  \begin{tabular}[c]{@{}c@{}}Normal\\ (degrees)\end{tabular} &
  \begin{tabular}[c]{@{}c@{}}Chamfer\\ $(\times10^{-4})$\end{tabular} \\ \hline
\multicolumn{1}{|l|}{CNR\cite{wang2016mesh}}      & 4.522 & 54.221 & 1.789        & 4.661 & 53.824  & 1.947        &    \textbf{0.978}    &  $23.043$    &    \textbf{0.591}       \\ \hline
\multicolumn{1}{|l|}{DMD-Net}  & \textbf{3.301} &\textbf{ 25.37} &\textbf{ 1.786} & \textbf{3.271} & \textbf{25.053} & \textbf{1.815 }      &   $2.561$     & \textbf{14.369}     &    $1.372$  \\ \hline
\end{tabular}
\end{center}
\end{table*}

\begin{figure*}[!htb]
    \begin{center}
    \captionsetup[subfigure]{labelformat=empty}
    \subfloat[(a)]{\includegraphics[width=0.16\linewidth ]{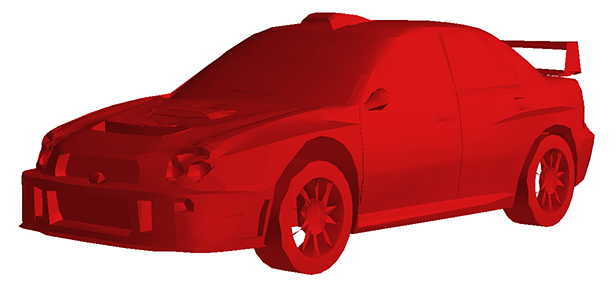}}
    \subfloat[(b)]{\includegraphics[width=0.16\linewidth ]{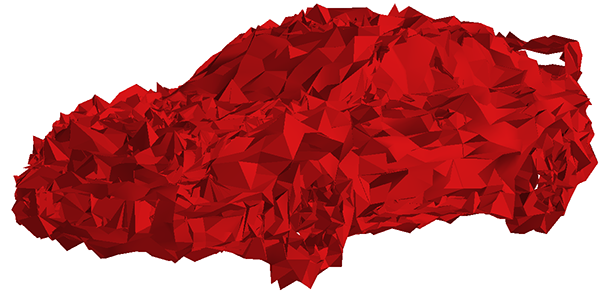}}
    \subfloat[(c)]{\includegraphics[width=0.16\linewidth ]{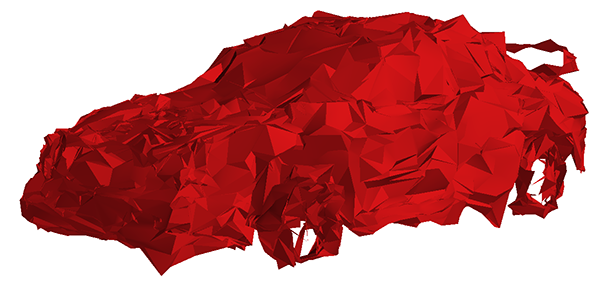}}
    \subfloat[(d)]{\includegraphics[width=0.16\linewidth ]{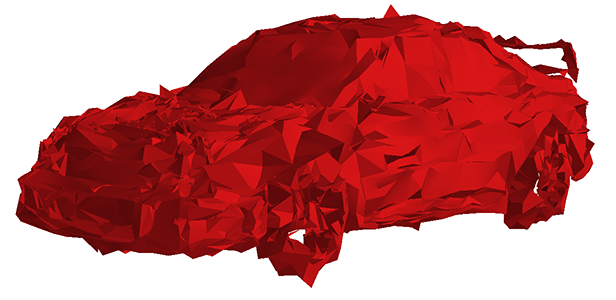}}
    \subfloat[(e)]{\includegraphics[width=0.16\linewidth ]{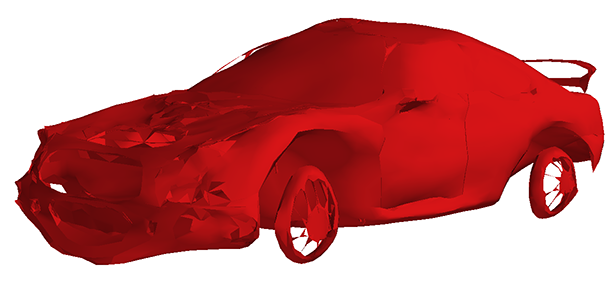}}
    \subfloat[(f)]{\includegraphics[width=0.16\linewidth ]{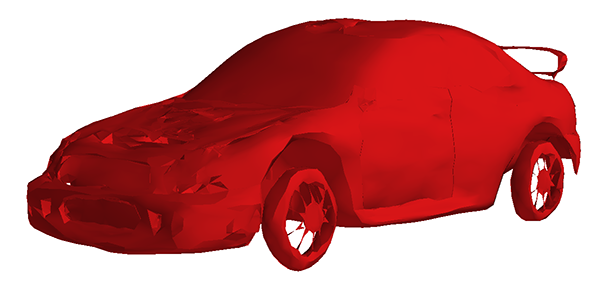}}
    \end{center}
    \vspace{-0.3cm}
    \caption{Comparison of DMD-Net and CNR\cite{wang2016mesh} trained on both ShapeNet and CNR dataset. (a) Ground truth, (b) Noisy Input, Output of (c) CNR network trained on CNR dataset, (d) CNR network trained on ShapeNet, (e) DMD-Net trained on CNR dataset, and (f) DMD-Net trained on ShapeNet.}
    \label{fig:CNRvsDMD}
\end{figure*}

\begin{figure*}[!htb]
    \begin{center}
    
    \vspace{-0.5cm}
    
    \captionsetup[subfigure]{labelformat=empty}
    
    \subfloat[]{\includegraphics[width=0.09\linewidth ]{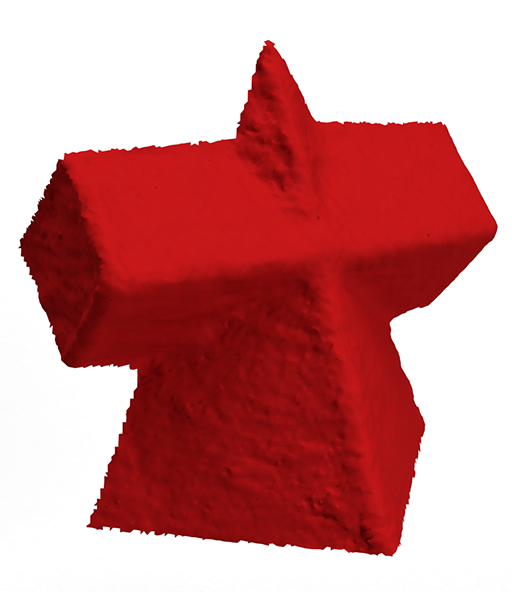}}
    \subfloat[]{\includegraphics[width=0.09\linewidth ]{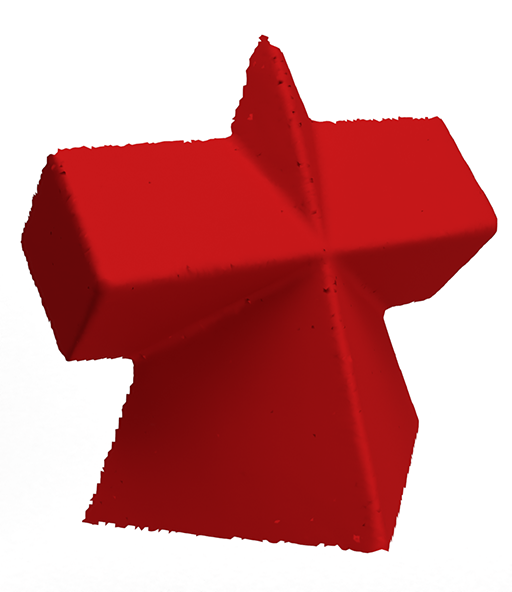}}
    \subfloat[]{\includegraphics[width=0.09\linewidth ]{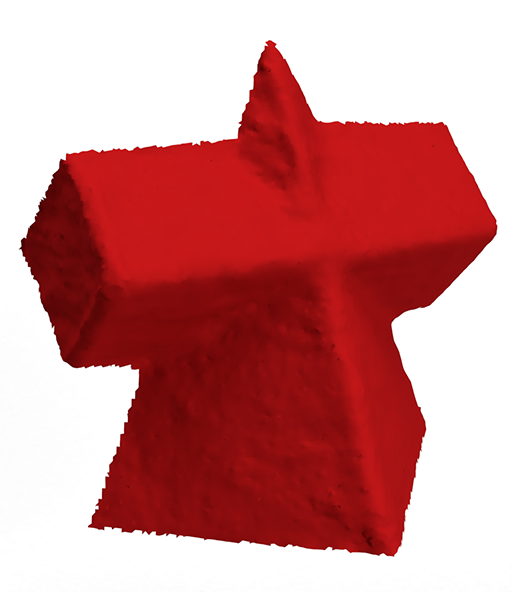}}
    \subfloat[]{\includegraphics[width=0.09\linewidth ]{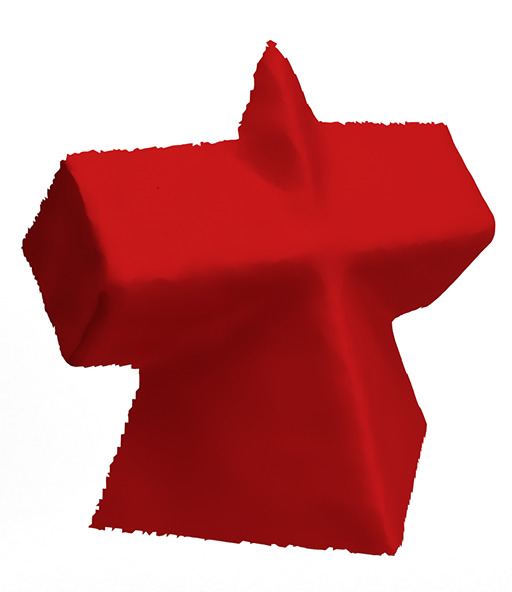}}
    \subfloat[]{\includegraphics[width=0.09\linewidth ]{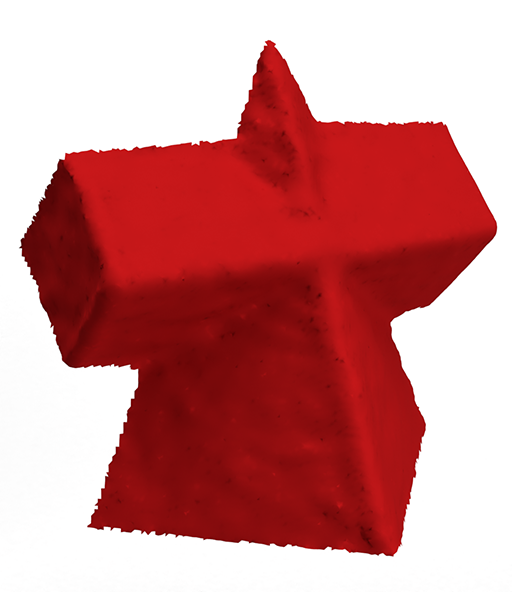}}
    \subfloat[]{\includegraphics[width=0.09\linewidth ]{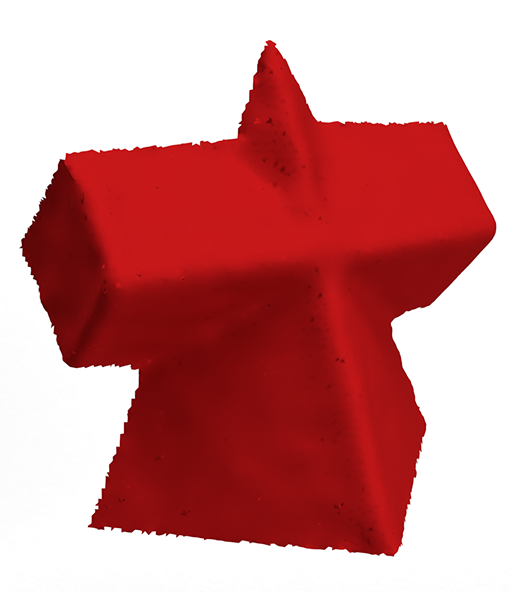}}
    \subfloat[]{\includegraphics[width=0.09\linewidth ]{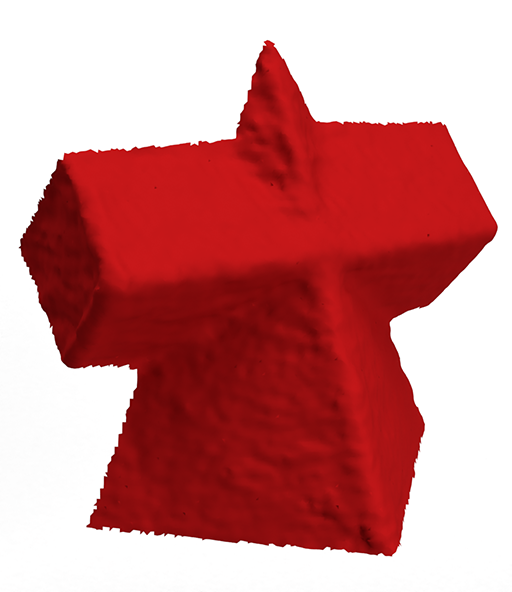}}
    \subfloat[]{\includegraphics[width=0.09\linewidth ]{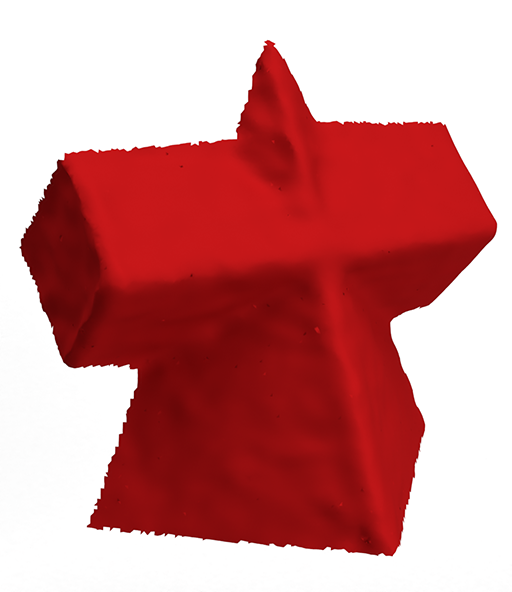}}
    \subfloat[]{\includegraphics[width=0.09\linewidth ]{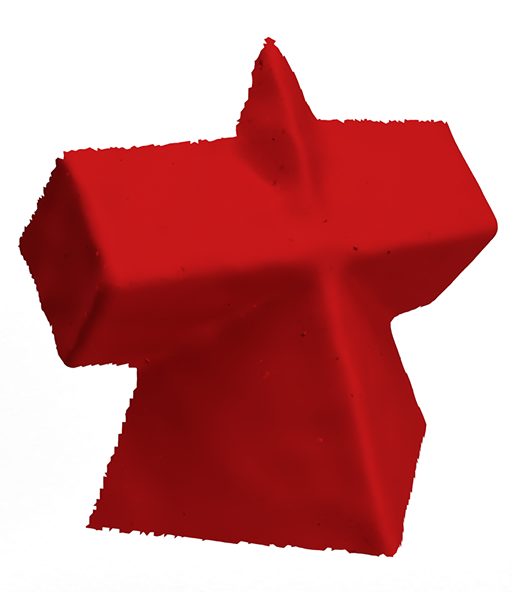}}
    \subfloat[]{\includegraphics[width=0.09\linewidth ]{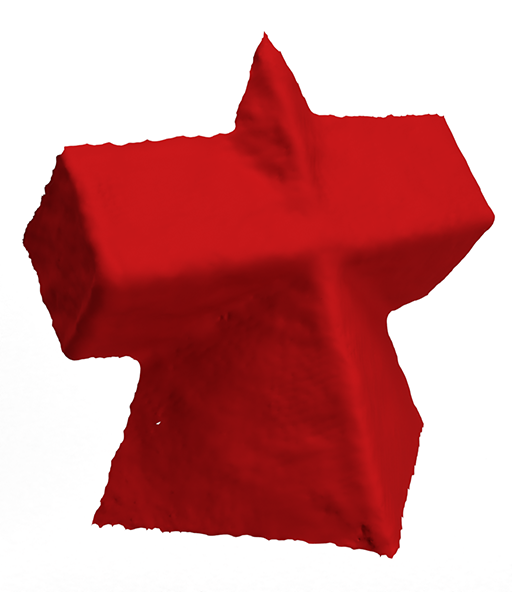}} \\
    \vspace{-0.8cm}
    \subfloat[Noisy]{\includegraphics[width=0.09\linewidth ]{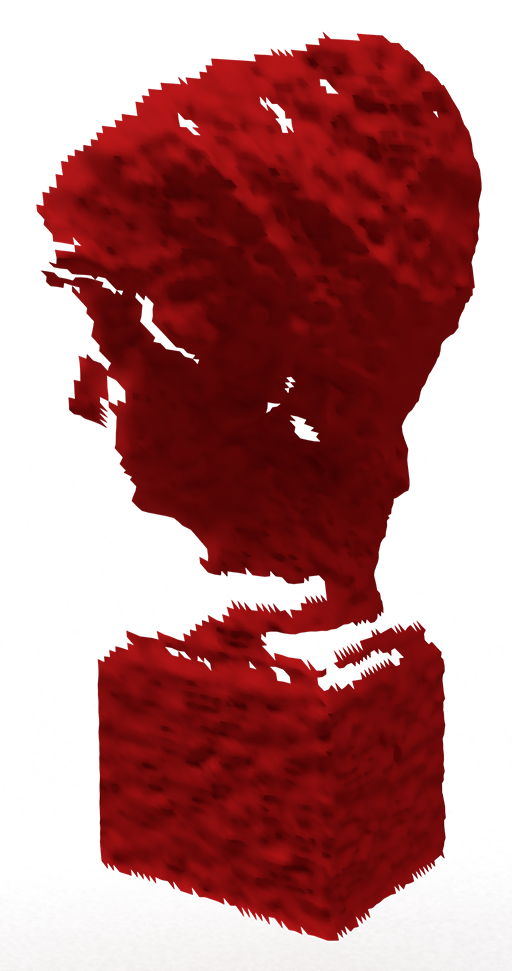}}
    \subfloat[Original]{\includegraphics[width=0.09\linewidth ]{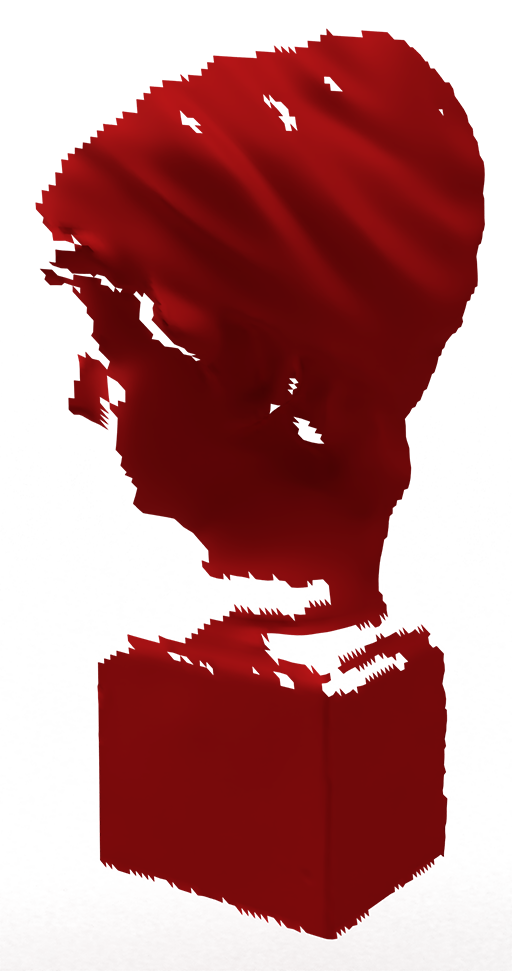}}
    \subfloat[BMD\cite{fleishman2003bilateral}]{\includegraphics[width=0.09\linewidth ]{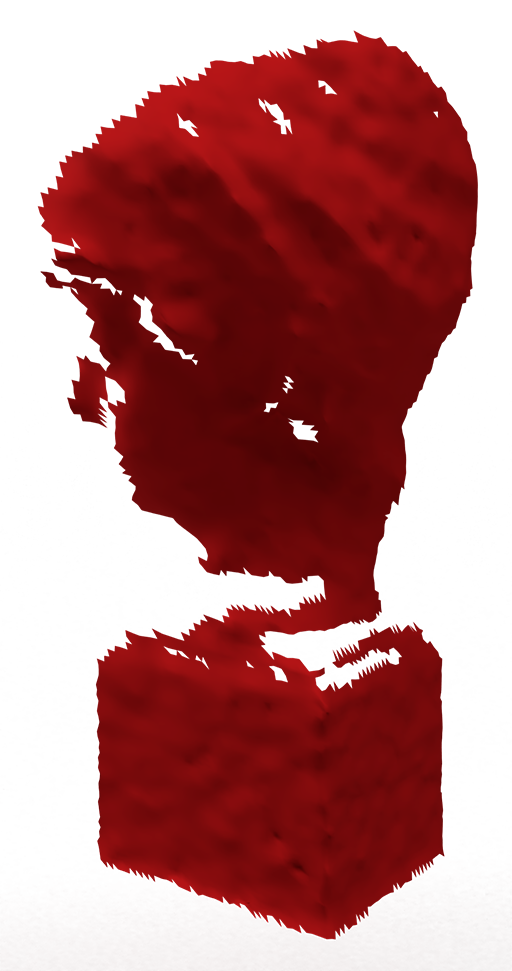}}
    \subfloat[L0M\cite{he2013mesh}]{\includegraphics[width=0.09\linewidth ]{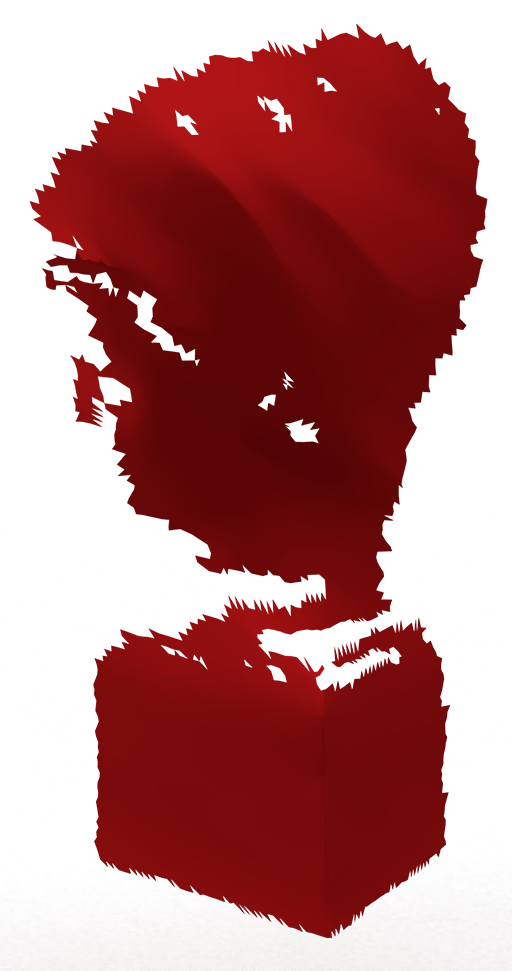}}
    \subfloat[BNF\cite{zheng2010bilateral}]{\includegraphics[width=0.09\linewidth ]{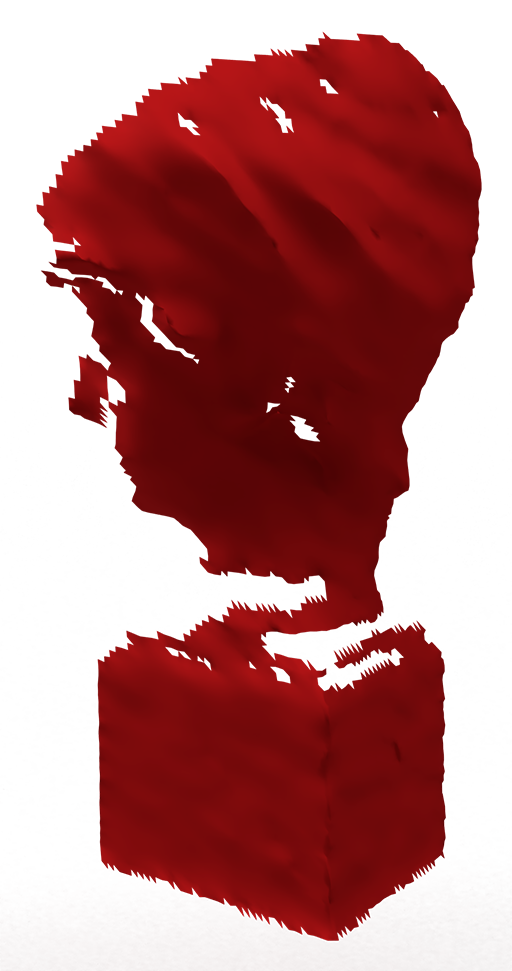}}
    \subfloat[FEFP\cite{sun2007fast}]{\includegraphics[width=0.09\linewidth ]{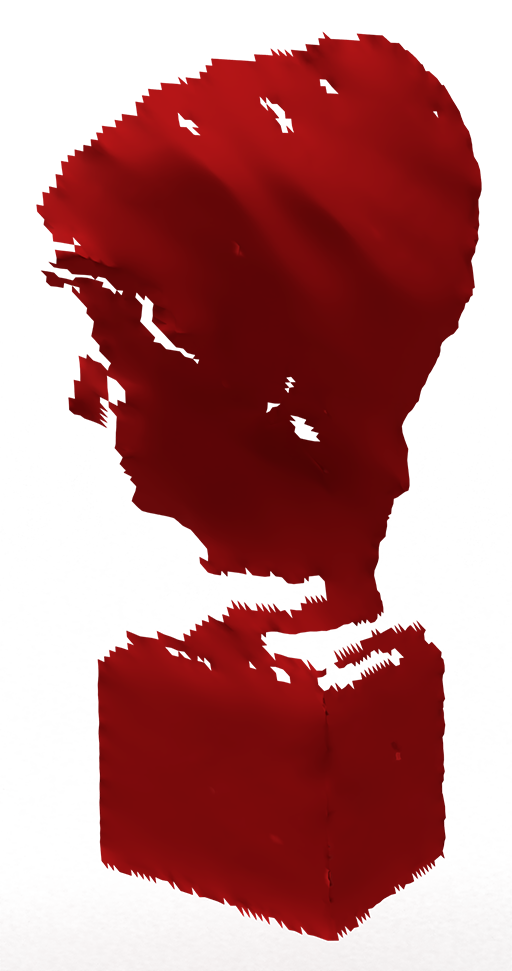}}
    \subfloat[NIFP\cite{jones2003non}]{\includegraphics[width=0.09\linewidth ]{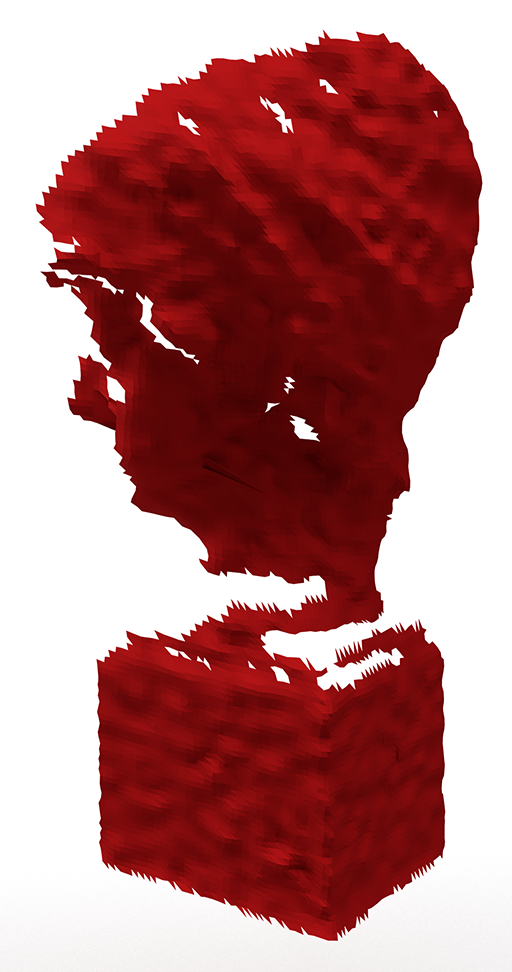}}
    \subfloat[GNF\cite{zhang2015guided}]{\includegraphics[width=0.09\linewidth ]{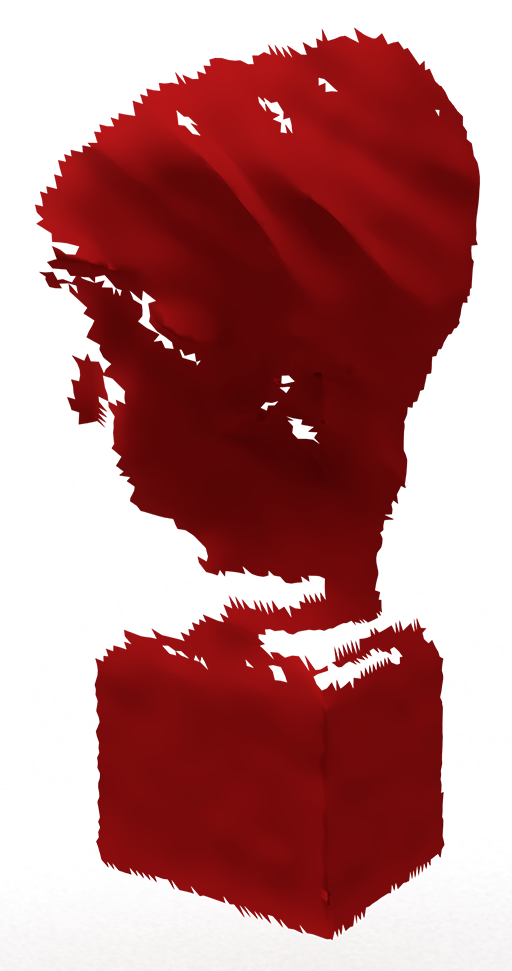}}
    \subfloat[CNR\cite{wang2016mesh}]{\includegraphics[width=0.09\linewidth ]{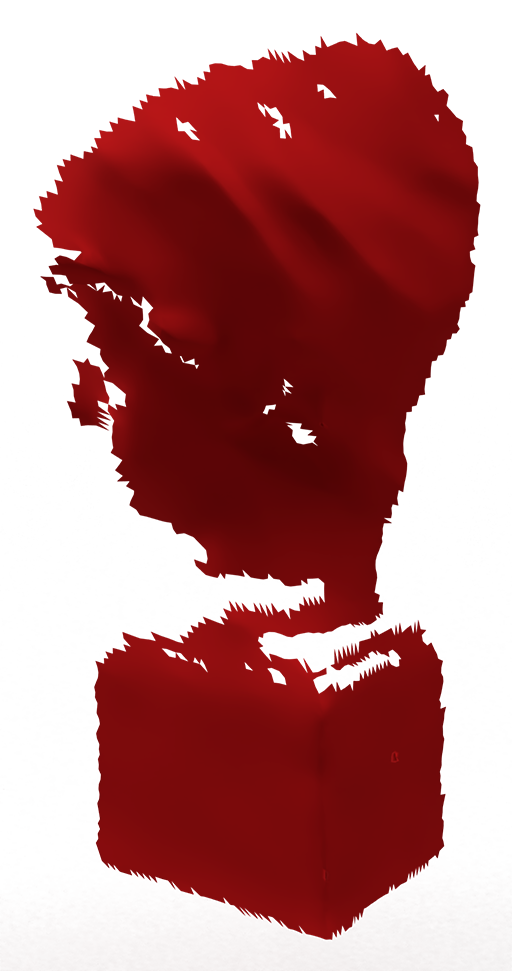}}
    \subfloat[Ours]{\includegraphics[width=0.09\linewidth ]{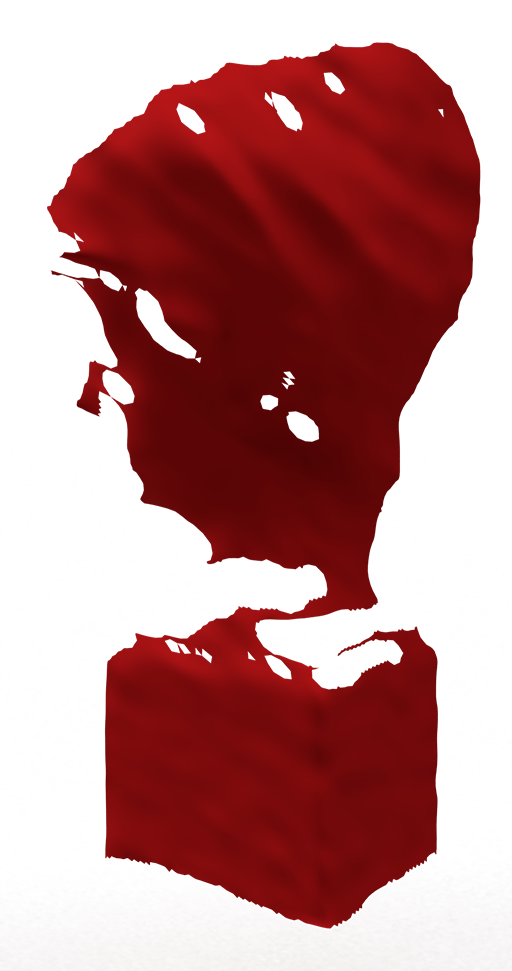}}
    \caption{Comparison of different mesh denoising methods on Kinect scans from the CNR dataset. }.
    \vspace{-1cm}
    \label{fig:comparison_real_scans}
    \end{center}
\end{figure*}

During training, we use four different noise types each with five different noise levels.  We show qualitative results under different noise levels in Figure \ref{fig:Mixed_Noise}. We also evaluate the performance of DMD-Net in case of noise levels that were not included during training in Table \ref{table:Noise_Evaluation}. Here, reference value refers to the metric distance between the noisy and the original mesh. As can be seen in Figure \ref{fig:Mixed_Noise}, our method performs very well even in the presence of high noise.


\subsection{Hyperparameters} \label{sec:hyperparameters}

DMD-Net contains around 30 million learnable parameters. We use ADAM \cite{kingma2014adam} optimizer with the following exponential decay rates: $\beta_1=0.9$ and $\beta_2=0.999$. We set our initial learning rate as $lr = 10^{-4}$. We train our final network for $200$ epochs. One epoch takes about an hour on NVIDIA Quadro RTX 5000.

\subsection{Ablation Studies} 

We conduct several ablation studies\footnote[3]{Ablation study Tables and their detailed explanations are presented in the supplementary material.}, where, we devise several variants of the proposed approach and show that the proposed approach outperforms all the variants. These include Training Scheme Ablation,  Dropout Rate Ablation, DMD-Net Structure Ablation, Loss Function Ablation, and Depth Ablation studies. In all of our ablation studies, we train the networks on mixed noise, that is, we randomly choose the noise type and the noise level in each iteration. For the depth ablation, we train all the variants for 200 epochs. For the rest of the ablation studies, we train all the variants for 60 epochs. Except the loss ablation study and the training scheme ablation study, all other ablation studies use a linear combination of loss functions with the following weights: $\lambda_{V} = \lambda_{N} = \lambda_{\kappa} = \lambda_{FE} = \lambda_{C} = 1$.


\subsection{Transformation Equivariance}
Let $\mathcal{T}$ be a transformation operator, let $\mathcal{D}$ be a mesh denoising algorithm and let $\mathcal{G}$ denote a mesh. We say that $\mathcal{D}$ has $\mathcal{T}$-equivariance, if for any given mesh, $\mathcal{G}$ we have  $\mathcal{T}(\mathcal{D}(\mathcal{G})) = \mathcal{D}(\mathcal{T}(\mathcal{G}))$. That is,  $\mathcal{D}$ is $\mathcal{T}$-equivariant, if  $\mathcal{D}$ and $\mathcal{T}$ commute. 
We now discuss whether DMD-Net is equivariant with respect to the following mentioned transformations.

Given a mesh, we first normalize it to fit inside a unit cube and shift the mesh to the origin. We then denoise it using DMD-Net by un-normalizing it back to it's original scale and shifting it back to its original location. Thus, DMD-Net is scale and translation equivariant as it first converts the mesh into a cannonical representation before denoising. DMD-Net is not rotation equivariant in the theoretical sense. However, the network learns to preserve equivariance to a high level. The study on rotation equivariance is presented in supplementary material along with both quantitative and qualitative results.

\subsection{Computation Time}
\begin{figure}[t]
\vspace{-0.4cm}
\centering
        \subfloat[]
        {\includegraphics[valign=t,width=0.5\linewidth]{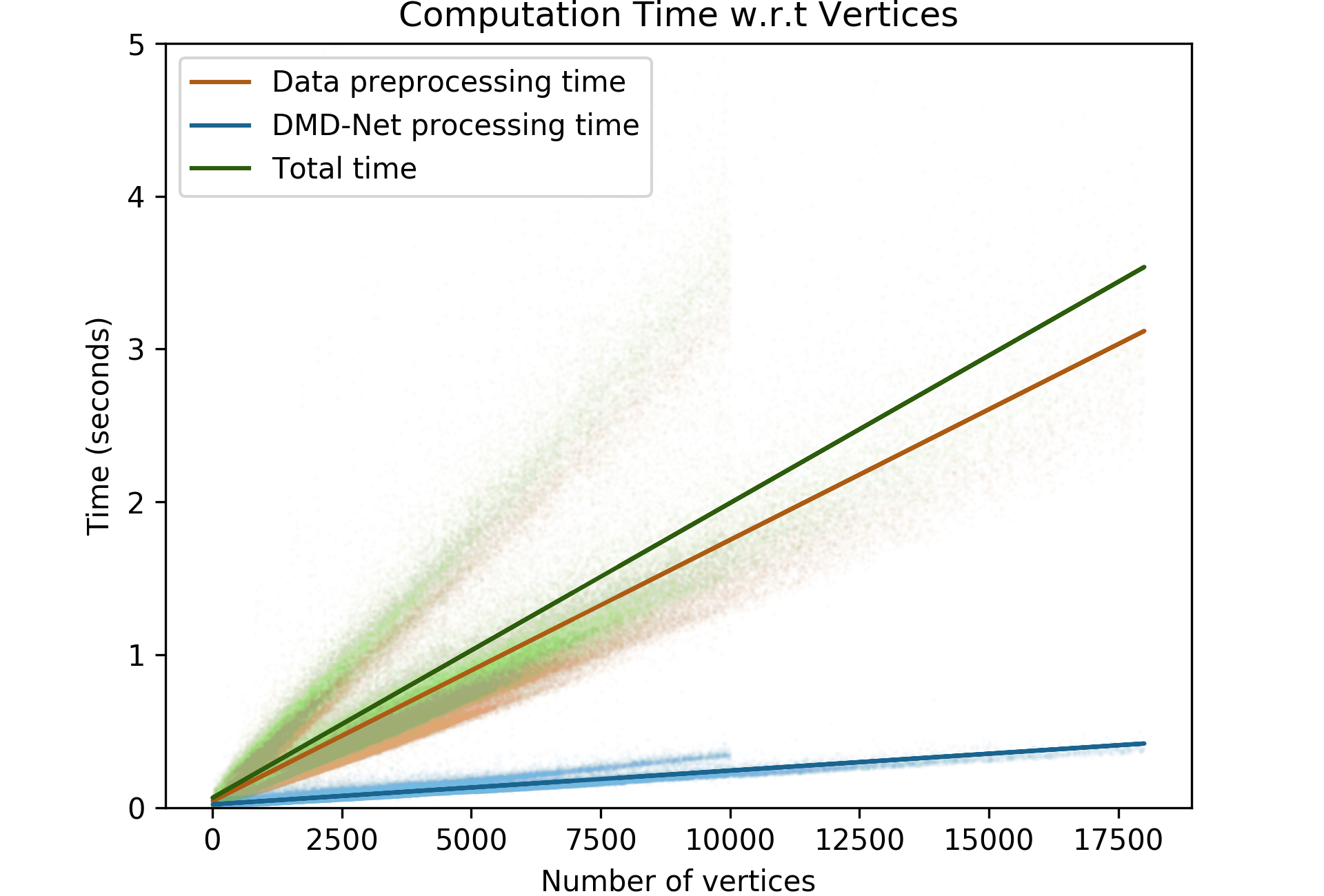}}
        \subfloat[]
        {\includegraphics[valign=t,width=0.5\linewidth]{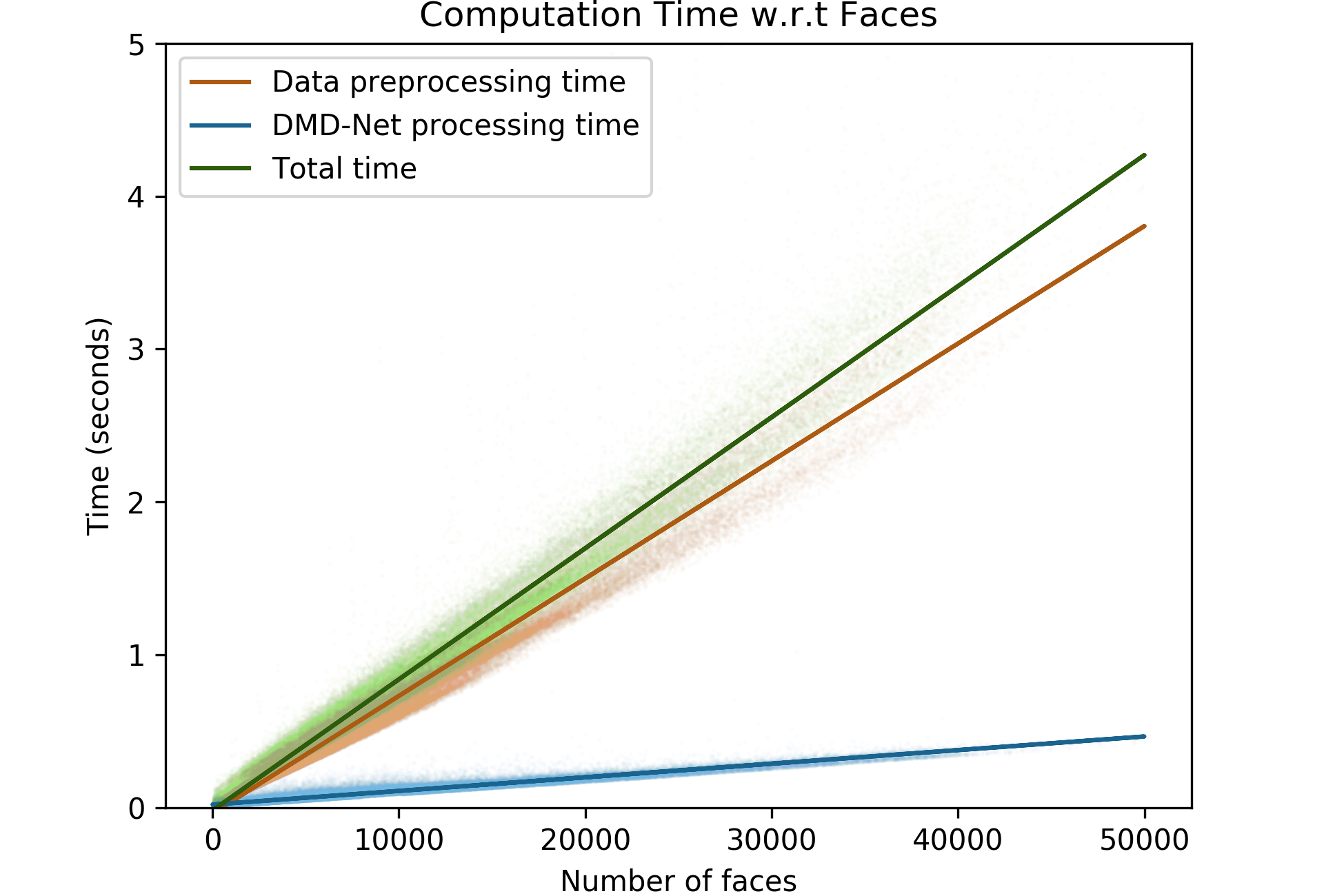}}
\caption{Growth of computation time with increase in (a) number of vertices and (b) number of faces.}
\vspace{-0.5cm}
  \label{fig:computation_time}
\end{figure}

The visual results shown in Figure \ref{fig:computation_time} contains a scatter-plot of data points along with the trendlines. As can be seen in the figures, our algorithm possesses a linear time growth which is an advantage of our method. When executed on a GPU, there is a significant speedup in the inference time of DMD-Net. The high speed and accuracy of DMD-Net makes it a suitable candidate for real-time mesh denoising.

\section{Conclusion and Future Work}
We have proposed a deep graph learning based framework to solve the mesh denoising problem. Through various experiments, we show that our method outperforms all proposed variants in the ablation studies. We also show that our method strikes a good balance between eliminating noise and avoiding over-smoothing. In this work, we make the assumption that the noise introduced during degradation is additive in nature. It would be interesting to explore how our methods performs on different types of noise models. DMD-Net is highly efficient in terms of computation time. In future, we propose to make it memory efficient by using model compression techniques, that would enable us to efficiently implement our method on a mobile phone or to directly integrate it into a portable 3D scanner device.

\small
\bibliographystyle{IEEEtran}
\bibliography{main}

\clearpage
\section*{\centering Appendix}

\setcounter{section}{0}
\section{DMD-Net Architecture}

In this section, we provide a detailed explanation about the architecture of DMD-Net .

\subsection{Feature Guided Transformer (FGT)} \label{sec:FGT}
DMD-Net is based on the Feature Guided Transformer paradigm and consists of three main components: the feature extractor, the transformer, and the denoiser. The main idea behind FGT is as follows. Consider the scenario where an oracle gives us access to the local features of the original ground-truth mesh. With such an access, we could use these local features to guide the denoising process. But, in reality, we do not have such an access to the original mesh features, so we instead try to construct an artificial oracle, which we call the feature extractor network. We train the feature extractor to estimate normal vector, mean curvature, and Gaussian curvature for each vertex of the original mesh from the given noisy mesh. 

These estimated local features serve as guidance for the transformer to compute a transformation matrix $\mathcal{W}_{tf}$. The noisy mesh is combined with the estimated local features through concatenation. The transformation  $\mathcal{W}_{tf}$ is applied on this combination to obtain an intermediate representation. The noisy mesh and its transformed intermediate representation are both passed to the denoiser which generates the final denoised mesh. The detailed inner workings are described in the subsequent subsections.

\subsection{Feature Extractor} \label{sec:FE}
The feature extractor internally contains a pair of two-stream networks which have two parallel streams, the upper one called the dual stream and the lower one called the primal stream.

\subsection{Transformer}
  The objective of the transformer is to convert the noisy input mesh into an intermediate representation, which is easier to perform denoising on. For this, we first use the feature extractor to estimate the local features for each vertex, such as the normal vectors, the mean curvature and the Gaussian curvature which are of size $n\times 3$, $n\times 1$ and $n\times 1$, respectively. Thus, the combined size of the local features is $n\times 5$. 
  
  These estimated local features are then used by the transformer to compute the transformation $\mathcal{W}_{tf}$. In order to compute this transformation, the local features are passed through a composition of several different layers, such as fully connected layer with ReLU activation, a series of densely connected aggregators (AGG), a graph aggregation layer, and a feature average pooling layer. Note that the second fully connected layer is not accompanied by a ReLU activation. This is because the distribution of scalar entries in matrix $\mathcal{W}_{tf}$ should not be skewed towards positive values. 
  
  The input to the feature average pooling layer is a matrix of size $n\times 4096$. The output obtained after pooling is a $4096$-dimensional vector which is reshaped to form the transformation matrix $\mathcal{W}_{tf}$ of size $8 \times 512$. On the other hand, the noisy input mesh is concatenated with the local features to obtain a combined input of size $n \times 8$. The transformation $\mathcal{W}_{tf}$ is then applied on this combined input to obtain an $n\times 512$ matrix, which is further passed through a graph aggregation layer along with ReLU activation. This is then processed by the denoiser to obtain the output denoised mesh.

\subsection{Denoiser} \label{sec:Denoiser}
The denoiser has a structure identical to that of the feature extractor. The only difference is the number of two-stream networks used and the number of hidden units in the last fully connected layer.

\subsection{Two-Stream Network} \label{sec:2SN}
The two-stream network is an asymmetric module consisting of two parallel streams, the lower one for performing aggregation in the primal graph and the upper one for performing aggregation in the dual graph. It consists of the primal-to-dual layer, a cascade of aggregator layers and a primal dual fusion layer.

\subsection{Aggregator (AGG)} \label{sec:AGG}
The Aggregator (AGG) performs graph aggregation by pooling in the features of the neighbouring nodes. The input to AGG is $\mathcal{X}$ (feature matrix) and $\mathcal{A}$ (adjacency matrix). The input graph to AGG can be in both forms, primal as well as dual. The output of AGG is given by
$g(\mathcal{X},\mathcal{A}) = \sigma(\hat{\mathcal{D}}^{-\frac{1}{2}}\hat{\mathcal{A}}\hat{\mathcal{D}}^{-\frac{1}{2}}\mathcal{X}\mathcal{W})$.
Here, $\sigma$ is the ReLU activation function, $\mathcal{W}$ is the learnable weight matrix, $\hat{\mathcal{A}} = \mathcal{A} + \mathcal{I}$ ($\mathcal{I}$ being the identity matrix), and $\hat{\mathcal{D}}$ is the diagonal node degree matrix of $\hat{\mathcal{A}}$.
In the two-stream network, the three AGG blocks are connected via residual skip-connections to avoid node feature collision (refer section \ref{sec:nfc}).

\subsection{Primal Dual Fusion (PDF)} \label{sec:PDF}
In Primal Dual Fusion, the input from both the streams are passed into the dual average pooling (DAP) layer, which intermixes the features from both the streams at the facet level. 
The PDF serves as a point of communication, allowing flow of information from one stream to the other. 

\subsection{Primal to Dual (P2D)} 
\label{sec:P2D}
The primal-to-dual layer appears in the dual stream of the two-stream network, where its objective is to convert the primal graph features $\mathcal{X_V}$ into the dual graph features $\mathcal{X_F}$.
The feature of each face is represented as the centroid of the features of the vertices constituting that face. Thus, we have $\mathcal{X_F} = \mathcal{D}_\mathcal{FV}^{-1} \mathcal{A_{FV}} \mathcal{X_V}$, where, $\mathcal{A_{FV}} = \mathcal{A}^T_\mathcal{VF}$ and $\mathcal{D}_\mathcal{FV}$ is the degree matrix denoting the number of vertices belonging to each face. Since our mesh is triangulated, each face is a triangle and each diagonal entry in $\mathcal{D}_\mathcal{FV}$ is 3. Therefore, we have the simplified expression $\mathcal{X_F} = \frac{1}{3} \mathcal{A_{FV}} \mathcal{X_V}$.

\subsection{Dual to Primal (D2P)} \label{sec:D2P}
The dual-to-primal layer is mainly used to convert the dual features $\mathcal{X_F}$ into a primal form $\mathcal{X_V}$.
We obtain $\mathcal{X_V}$ by pre-multiplying $\mathcal{X_F}$ with the degree normalized vertex-face adjacency matrix $\mathcal{\hat{A}_{VF}}$. Here, $\mathcal{\hat{A}_{VF}} = \mathcal{D}_\mathcal{VF}^{-1} \mathcal{A_{VF}}$, in which $\mathcal{A_{VF}}$ is the vertex-face adjacency matrix and $\mathcal{D}_\mathcal{VF}$ is the degree matrix denoting the number of faces in which a particular vertex lies.

\subsection{Dual Average Pooling (DAP)} \label{sec:DAP}
Let $\mathcal{X_V}$ and $\mathcal{X_F}$ be the input to the Dual Average Pooling layer arriving via the primal stream and the dual stream, respectively.
The sizes of $\mathcal{X_V}$ and $\mathcal{X_F}$ are $n \times k$ and $f \times k$, respectively. Let $\mathcal{F}$ denote the collection of faces of the primal graph. For each face in $\mathcal{F}$, we perform the following: gather the features of the three vertices belonging to that face from $\mathcal{X_V}$ and gather the feature of that face from $\mathcal{X_F}$. Let $p_1$, $p_2$, and $p_3$ denote the features of the three vertices and $d$ denote the feature of the face. Note that, in general, $d$ need not be the centroid of $p_1$, $p_2$, and $p_3$. We measure the distance of $d$ from the three vertices in each of the $k$ dimensions by evaluating the following three quantities: $l^a = |p_1 - d|$, $l^b = |p_2 - d|$, and $l^c = |p_3 - d|$, where $|\cdot|$ denotes the element-wise absolute value. $l^a$, $l^b$, and $l^c$ are all k-dimensional vectors. We thus obtain $l_i = ( l^a_i, l^b_i, l^c_i )$, the fused feature descriptor for $i^{th}$ face. Collecting the fused descriptor for all the faces gives us $L = ( l_1, l_2,\cdots,l_i,\cdots,l_f )$, a tensor with three axes having size $f \times 3 \times k$. 
In order to keep our network permutation-invariant, we devise the following strategy to pool the fused features.
We pool by using average pooling across the second axis, which gives an output $u$ of size $f \times k$. Let $L^a = (l_1^a, l_2^a, \cdots , l_f^a)$, $L^b = (l_1^b, l_2^b, \cdots , l_f^b)$, and  $L^c = (l_1^c, l_2^c, \cdots , l_f^c)$, each of size $f \times k$. Then, $u = \frac{1}{3} \sum_{i\in\{a,b,c\}}L^i$ is the output of DAP. Figure 3(c) in main paper illustrates the pooling mechanism of DAP. In the figure, the outer tetrahedron with red nodes is the primal mesh, whereas the inner tetrahedron with blue nodes is the dual mesh. These meshes are outputs of the aggregation block, which separately processes the primal and the dual meshes. Hence, the blue nodes are not centroids of the red nodes, in general.

\subsection{Feature Average Pooling (FAP)} \label{sec:FAP}
The Feature Average Pooling layer is used inside the transformer to pool the features across all the vertices and output the transformation. Given $H = \{ h_1, h_2, \cdots, h_i, \cdots, h_n \}$ as the input to the FAP layer, it outputs $z=\frac{1}{n} \sum_{i=1}^{n}h_i$.

\section{Node Feature Collision}
\label{sec:nfc}
We explain in detail the concept of \textit{node feature collision}, an undesirable artifact of the vanilla Graph-CNN network. We also discuss the strategy adopted by us to eliminate this problem. 

Let $g(\mathcal{X},\mathcal{A}) = \sigma(\hat{\mathcal{D}}^{-\frac{1}{2}}\hat{\mathcal{A}}\hat{\mathcal{D}}^{-\frac{1}{2}}\mathcal{X}\mathcal{W})$ denote the output of the aggregator layer, for the inputs $\mathcal{X}$ (node feature matrix) and $\mathcal{A}$ (adjacency matrix). In the transformer as well as the two-stream network, there are three AGG blocks used in series along with residual skip connections. We now justify the use of residual skip connections. Assume the case where we do not use these skip connections. Then the output would be $\mathcal{X'} = g( g( g(\mathcal{X},\mathcal{A}),\mathcal{A}),\mathcal{A})$. The series of AGG blocks in this form encounters a peculiar issue in some cases. If there exist two adjacent vertices in the graph whose neighbourhood is exactly the same, or in other words, if two rows of $\hat{\mathcal{A}}$ are identical, then the two vertices would get mapped on to the same output point in the feature space by $g$. For instance, if the input is a tetrahedron, then in the normalized adjacency matrix, all the four vertices are adjacent to each other and all four have exactly the same neighbourhood (note that $\hat{\mathcal{A}}$ contains self loop for each vertex). Thus, all four points of the tetrahedron get mapped to the same output after aggregation. This effect is highly undesirable and we refer to it as \textit{node feature collision}. We solve this problem by making use of residual skip connections from the input to the output. Thus, the modification now becomes: $h(\mathcal{X},\mathcal{A}) = g(\mathcal{X},\mathcal{A}) + \mathcal{X}$. With the introduction of the skip connections we thus have the output $\mathcal{X'} = h( h( h(\mathcal{X},\mathcal{A}),\mathcal{A}),\mathcal{A})$.


\section{Loss Functions}

Let $\mathcal{G}_{gt} = (\mathcal{V},\mathcal{E},\mathcal{F},\mathcal{P}_{gt})$, $\mathcal{G}_{noisy} = (\mathcal{V},\mathcal{E},\mathcal{F},\mathcal{P}_{noisy})$, and $\mathcal{G}_{out} = (\mathcal{V},\mathcal{E},\mathcal{F},\mathcal{P}_{out})$ denote the original ground truth mesh, the noisy mesh, and the denoised mesh obtained using DMD-Net, respectively. Let $\theta_{gt}$,  $\theta_{noisy}$, and  $\theta_{out}$ denote the interior angles of the vertices on their respective faces. If vertex $v$ is contained in face $u$, then $\theta(v,u)$ denotes the interior angle of $v$ on triangular face $u$. We define several loss functions whose objective is to make $\mathcal{G}_{out}$ as close to $\mathcal{G}_{gt}$ as possible.

\subsection{Vertex Loss} 
The vertex loss computes the mean Euclidean distance between the corresponding vertices. It is defined in Equation \ref{eq:vertex_loss}, where, $\mathcal{P}_{out}(v)$ and $\mathcal{P}_{gt}(v)$ denote the feature vectors of the vertex $v$.
\begin{equation}
\mathcal{L}_{vertex} = \frac{1}{|\mathcal{V}|}\sum_{v \in \mathcal{V}}\norm{\mathcal{P}_{out}(v) - \mathcal{P}_{gt}(v)}_2^2
\label{eq:vertex_loss}
\end{equation}

\subsection{Normal Loss} \label{sec:loss_normal}
The normal loss computes the mean angular deviation between the normals of the corresponding faces. 
Let $\widehat{\mathcal{N}}_{out}(s)$ and $\widehat{\mathcal{N}}_{gt}(s)$ denote the unit normal of the face $s$, where the normal direction of a face is computed by taking the cross product between two edges of the triangular face. The normal loss is defined in Equation \ref{eq:normal_loss}, where, $\langle \cdot,\cdot \rangle$ denotes inner-product.
\begin{equation}
\mathcal{L}_{normal} = \frac{1}{|\mathcal{F}|}\sum_{s \in \mathcal{F}}\cos^{-1}\bigl(\langle\widehat{\mathcal{N}}_{out}(s), \widehat{\mathcal{N}}_{gt}(s)\rangle\bigr)
\label{eq:normal_loss}
\end{equation}

\subsection{Curvature Loss} \label{sec:loss_curvature}
We use two types of curvature loss: the mean curvature loss and the Gaussian curvature loss. The mean curvature $\kappa^H$  is computed using the mean curvature normal operator and the Gaussian curvature $\kappa^G$ is computed using the Gaussian curvature operator. Let $\mathcal{N}(v)$ denote the neighbourhood of vertex $v$ in the graph and let $\mathcal{N_F}(v)$ denote the set of faces that contain the vertex $v$. Then the curvature values at $v$ for the two graphs $\mathcal{G}_{gt}$ and $\mathcal{G}_{out}$ are denoted by $\kappa^H_{out}(v)$, $\kappa^H_{gt}(v)$, $\kappa^G_{out}(v)$, and $\kappa^G_{gt}(v)$. The mean curvatures are defined in Equations \ref{eq:k_H_out} and \ref{eq:k_H_gt}.
\begin{equation}
\kappa^H_{out}(v) = \frac{1}{4\mathcal{A}_{Mixed}(v)}\normmod{\sum_{u \in \mathcal{N}(v)}w_{vu}(\mathcal{P}_{out}(v) - \mathcal{P}_{out}(u))}_2 \\
\label{eq:k_H_out}
\end{equation}
\begin{equation}
\kappa^H_{gt}(v) = \frac{1}{4\mathcal{A}_{Mixed}(v)}\normmod{\sum_{u \in \mathcal{N}(v)}w_{vu}(\mathcal{P}_{gt}(v) - \mathcal{P}_{gt}(u))}_2 \\
\label{eq:k_H_gt}
\end{equation}
Here, $w_{vu} = \sum_{f \in \mathcal{F}_{vu}}\cot\alpha_{vu}^{f}$ denotes the cotangent weights, in which, $\alpha_{vu}^{f}$ is the interior angle of the vertex opposite to edge $vu$ in face $f$ and $\mathcal{F}_{vu}$ is the collection of all faces that contain edge $vu$.
$\mathcal{A}_{Mixed}(v)$ denotes the augmented version of the Voronoi region area of a vertex $v$. This augmentation takes into account the case where the triangular faces of the mesh are obtuse. The Gaussian curvatures are defined in Equations \ref{eq:k_G_out} and \ref{eq:k_G_gt}.
\begin{equation}
\kappa^G_{out}(v) = \frac{1}{\mathcal{A}_{Mixed}(v)}(2\pi - \sum_{u \in \mathcal{N_F}(v)}\theta_{out}(v,u)) \\
\label{eq:k_G_out}
\end{equation}
\begin{equation}
\kappa^G_{gt}(v) = \frac{1}{\mathcal{A}_{Mixed}(v)}(2\pi - \sum_{u \in \mathcal{N_F}(v)}\theta_{gt}(v,u)) \\
\label{eq:k_G_gt}
\end{equation}
The two components of curvature loss are then defined in Equation \ref{eq:L_H} and \ref{eq:L_G}.
\begin{equation}
\mathcal{L}_{H} = \frac{1}{|\mathcal{V}|}\sum_{v \in \mathcal{V}}\norm{\kappa^H_{out}(v) - \kappa^H_{gt}(v)}_1 \biggl(\kappa^H_{gt}(v)\biggr)^2
\label{eq:L_H}
\end{equation}
\begin{equation}
\mathcal{L}_{G} = \frac{1}{|\mathcal{V}|}\sum_{v \in \mathcal{V}}\norm{\kappa^G_{out}(v) - \kappa^G_{gt}(v)}_1 \biggl(\kappa^G_{gt}(v)\biggr)^2
\label{eq:L_G}
\end{equation}
Here, $\norm{\cdot}_1$ is the absolute value operator. Finally we define the total curvature loss as the average of the two curvature losses in Equation \ref{eq:curvature_loss}.
\begin{equation}
  \mathcal{L}_{curvature} = (\gamma_H \mathcal{L}_{H} + \gamma_G \mathcal{L}_{G})
  \label{eq:curvature_loss}
\end{equation}
Here, $\gamma_H$ and $\gamma_G$ are weights used for combining the two curvatures. Based on distribution of objects in our dataset, we choose $\gamma_H = 10^{-6}$ and $\gamma_G = 1$.

\subsection{Chamfer Loss} \label{sec:loss_chamfer}
Let $\mathcal{V}_{out}$ and $\mathcal{V}_{gt}$ denote the vertex sets of the two graphs and let $\mathcal{P}_{out}(v)$ and $\mathcal{P}_{gt}(v)$ denote the features of the vertex $v$ in the two graphs. Then the Chamfer loss is defined in Equation \ref{eq:chamfer_loss}

\begin{multline}
  \mathcal{L}_{chamfer} = \frac{1}{|\mathcal{V}_{out}|}\sum_{u \in \mathcal{V}_{out}}\min_{v \in \mathcal{V}_{gt}}\norm{\mathcal{P}_{out}(u) - \mathcal{P}_{gt}(v)}_2^2  \\ + \frac{1}{|\mathcal{V}_{gt}|}\sum_{v \in \mathcal{V}_{gt}}\min_{u \in \mathcal{V}_{out}}\norm{\mathcal{P}_{out}(u) - \mathcal{P}_{gt}(v)}_2^2 
  \label{eq:chamfer_loss}
\end{multline}

\subsection{Feature Extractor Loss} \label{sec:loss_FE}
The output of the feature extractor block is a matrix of size $n \times 5$ which contains the estimate of normals and curvatures for each vertex. Let  $\widehat{\mathcal{N}}_{fe}(v)$, $\kappa^H_{fe}(v)$, and $\kappa^G_{fe}(v)$ be the normal, the mean curvature and the Gaussian curvature respectively estimated by the feature extractor for vertex $v$. Let  $\widehat{\mathcal{N}}_{gt}(v)$, $\kappa^H_{gt}(v)(v)$, and $\kappa^G_{gt}(v)$ be the normal, the mean curvature and the Gaussian curvature respectively of vertex $v$ in the original ground-truth mesh. The feature extractor loss is then given in Equation \ref{eq:feature_extractor_loss}

\begin{multline}
    \mathcal{L}_{FE} = \frac{1}{|\mathcal{V}|}\sum_{v \in \mathcal{V}}  \biggr(\norm{\widehat{\mathcal{N}}_{fe}(v) - \widehat{\mathcal{N}}_{gt}(v)}_2^2\\
    +  \norm{\kappa^H_{fe}(v) -\kappa^H_{gt}(v)}_2^2 
    +  \norm{\kappa^G_{fe}(v) -\kappa^G_{gt}(v)}_2^2 \biggr)
    \label{eq:feature_extractor_loss}
\end{multline}

\subsection{Loss function weights} \label{sec:loss_weights}
For training our network, we use a linear combination of the loss functions described above.The weights used for vertex loss, normal loss, curvature loss, chamfer loss and the feature extractor loss are $\lambda_{V}=1$, $\lambda_{N}=0.2$, $\lambda_{\kappa}=0.01$, $\lambda_{C}=0.05$ and  $\lambda_{FE}=1$, respectively.

We define several loss functions whose objective is to make $\mathcal{G}_{out}$ as close to $\mathcal{G}_{gt}$ as possible. In order to get $\mathcal{P}_{out}$ close to $\mathcal{P}_{gt}$, the natural loss function to use would be a metric that compares the distance between the corresponding pair of vertices in the two graphs. In cases where the correspondence between vertices is known, we can directly use the vertex-wise Euclidean distance and then average them over all the vertices. We use this loss and call it vertex loss. However, in cases where the correspondence between the vertices of the two graphs is not known, it is appropriate to use distance metric defined between two sets as the loss function, such as, Hausdorff distance, Chamfer distance, Earth mover's distance, etc. We include the Chamfer loss in our framework.

It might be argued that, since in DMD-Net the correspondence between vertices is preserved throughout, what is the need for using Chamfer distance? A response to this argument might be to note that while performing denoising using a deep learning framework, the network can contain components that may destroy the correspondence. For instance, one can conceive of an autoencoder which obtains a latent code for the entire graph in the encoding stage, thereby destroying all the structure, and while decoding, it might output a fixed number of vertices thereby changing the number of vertices in the two graphs. There could be many other such possibilities that destroy the correspondence. In such cases, vertex loss would fail. Hence, for the sake of making our framework more general and robust to these possibilities, we allow the use of both Chamfer loss and Vertex loss.

In the absence of original correspondence, we can still obtain a correspondence by solving the assignment problem through cost minimization. Once a correspondence is obtained, many other loss functions can further be used. For instance, comparing the length of edges, area of faces, location of centroids, direction of normal of faces, laplacian of vertices, curvature of vertices, etc. However, in spite of these several possibilities, we use only the loss functions that enforce the vertex locations, face normals, and curvatures of the output denoised mesh to be aligned with that of the ground-truth mesh.

\section{Ablation Studies} \label{sec:ablation}

We conduct several ablation studies, where, we devise several variants of the proposed approach and show that the proposed approach outperforms all the variants. This establishes the importance of each component in our network. In all of our ablation studies, we train the networks on mixed noise, that is, we randomly choose the noise type and the noise level in each iteration. For the depth ablation we train all the variants for 200 epochs. For the rest of the ablation studies, we train all the variants for 60 epochs. Except the loss ablation study and the training scheme ablation study, all other ablation studies use a linear combination of loss functions with the following weights: $\lambda_{V} = \lambda_{N} = \lambda_{\kappa} = \lambda_{FE} = \lambda_{C} = 1$

\subsection{Training Scheme Ablation}

The loss functions discussed in Section \ref{sec:LossFunctions} can be segregated into two categories. The first category consists of the feature extractor loss which provides feedback solely to the feature extractor. The second category consists of the rest of the loss functions which provide feedback to the entire network. Thus, this segregation, naturally opens the possibility of various different competing training schemes as below.

\noindent \textbullet\ \textit{Joint Optimization (Joint)}.
The final loss function is a linear combination of all the loss functions. All the components of the network are jointly trained using this final loss function. In this case $\lambda_{V} = \lambda_{N} = \lambda_{\kappa} = \lambda_{FE} = \lambda_{C} = 1$.

\noindent \textbullet\ \textit{Alternating Optimization (Alter)}.
The training alternates between training the feature extractor and training the rest of the network. In odd iterations, the feature extractor is trained using the feature extractor loss while the rest of the network is frozen. We thus have, $\lambda_{FE} = 1$ and $\lambda_{V} = \lambda_{N} = \lambda_{\kappa} = \lambda_{C} = 0$. In the even iterations, the feature extractor is frozen while the rest of the network is trained using the loss functions in the second category. We thus have $\lambda_{FE} = 0$ and $\lambda_{V} = \lambda_{N} = \lambda_{\kappa} = \lambda_{C} = 1$.

\begin{table}[t]
\begin{center}
\renewcommand{\arraystretch}{1.3}
\caption{Training Scheme - Ablation Study.}
\label{table:training_strategy}
\footnotesize
\begin{adjustbox}{width=\linewidth,center}
{
\begin{tabular}{|l|c|c|c|c|c|c|}
\hline
\multicolumn{1}{|c|}{\multirow{2}{*}{\textbf{Model}}} & \multicolumn{3}{c|}{\textbf{test-intra}} & \multicolumn{3}{c|}{\textbf{test-inter}} \\ \cline{2-7} 
\multicolumn{1}{|c|}{} &
  \begin{tabular}[c]{@{}c@{}}Vertex\\ $(\times 10^{-4})$\end{tabular} &
  \begin{tabular}[c]{@{}c@{}}Normal\\ (degrees)\end{tabular} &
  \begin{tabular}[c]{@{}c@{}}Chamfer\\ $(\times 10^{-4})$\end{tabular} &
  \begin{tabular}[c]{@{}c@{}}Vertex\\ $(\times 10^{-4})$\end{tabular} &
  \begin{tabular}[c]{@{}c@{}}Normal\\ (degrees)\end{tabular} &
  \begin{tabular}[c]{@{}c@{}}Chamfer\\ $(\times 10^{-4})$\end{tabular} \\ \hline
Joint                         & \textbf{3.917} & \textbf{25.458} & \textbf{2.028} & \textbf{3.717} & \textbf{25.184} & \textbf{2.035}          \\ \hline
Alter                    & $5.112$ & $26.458$ & $2.46$ & $4.882$ & $26.056$ & $2.459$          \\ \hline
2-Phase                & $3.987$ & $26.216$ & $2.097$ & $3.937$ & $25.722$ & $2.155$          \\ \hline
U-FE             & $4.768$ & $25.767$ & $2.399$ & $5.007$ & $25.387$ & $2.544$          \\ \hline
\end{tabular}
}
\end{adjustbox}
\end{center}
\end{table}

\begin{table}[t]
\begin{center}
\renewcommand{\arraystretch}{1.3}
\caption{Dropout Rate - Ablation Study.}
\label{table:dropout}
\footnotesize
\begin{adjustbox}{width=\linewidth,center}
{
\begin{tabular}{|l|c|c|c|c|c|c|}
\hline
\multicolumn{1}{|c|}{\multirow{2}{*}{\textbf{Model}}} & \multicolumn{3}{c|}{\textbf{test-intra}} & \multicolumn{3}{c|}{\textbf{test-inter}} \\ \cline{2-7} 
\multicolumn{1}{|c|}{} &
  \begin{tabular}[c]{@{}c@{}}Vertex\\ $(\times10^{-4})$\end{tabular} &
  \begin{tabular}[c]{@{}c@{}}Normal\\ (degrees)\end{tabular} &
  \begin{tabular}[c]{@{}c@{}}Chamfer\\ $(\times10^{-4})$\end{tabular} &
  \begin{tabular}[c]{@{}c@{}}Vertex\\ $(\times10^{-4})$\end{tabular} &
  \begin{tabular}[c]{@{}c@{}}Normal\\ (degrees)\end{tabular} &
  \begin{tabular}[c]{@{}c@{}}Chamfer\\ $(\times10^{-4})$\end{tabular} \\ \hline
$D_r=0$                             & \textbf{3.917} & \textbf{25.458} & \textbf{2.028} & \textbf{3.717} & \textbf{25.184} & \textbf{2.035}          \\ \hline
$D_r=0.2$                            & $101.777$ & $45.86$ & $17.912$ & $98.108$ & $46.778$ & $18.743$          \\ \hline
$D_r=0.35$                           & $94.623$ & $47.578$ & $19.767$ & $95.461$ & $47.836$ & $21.056$          \\ \hline
$D_r=0.5$                            & $220.129$ & $47.905$ & $33.388$ & $242.294$ & $49.462$ & $37.828$          \\ \hline
\end{tabular}
}
\end{adjustbox}
\end{center}
\end{table}

\noindent \textbullet\ \textit{Two Phase Optimization (2-Phase)}.
The training process is divided into two phases: the pre-training phase and the post-training phase. In the pre-training phase the feature extractor is trained for several epochs keeping the rest of the network frozen, with the following weights of loss functions $\lambda_{FE} = 1$ and $\lambda_{V} = \lambda_{N} = \lambda_{\kappa} = \lambda_{C} = 0$. In the post-training phase the rest of the network is trained for the remaining epochs keeping the feature extractor frozen, in which case, the loss function weights are $\lambda_{FE} = 0$ and $\lambda_{V} = \lambda_{N} = \lambda_{\kappa} = \lambda_{C} = 1$.

\noindent \textbullet\ \textit{Unsupervised Feature Extraction (U-FE)}.
The feature extractor loss is switched off in this case. The feature extractor receives feedback from rest of the loss functions. Since the feature extractor loss is disabled, the supervised training with local features does not occur. Hence, the feature extraction step becomes an unsupervised process. The weights of the loss functions are $\lambda_{FE} = 0$ and $\lambda_{V} = \lambda_{N} = \lambda_{\kappa} = \lambda_{C} = 1$.


In this ablation study, we use a pair of two-stream networks in both the feature extractor as well as the denoiser. As visible in Table \ref{table:training_strategy}, the joint optimization scheme outperforms the rest of the training schemes. From now on we use joint optimization for the remaining ablation studies.

\begin{table}[t]
\begin{center}
\renewcommand{\arraystretch}{1.3}
\caption{Graph Attention Networks - Ablation Study.}
\label{table:attention}
\footnotesize
\begin{adjustbox}{width=\linewidth,center}
{
\begin{tabular}{|l|c|c|c|c|c|c|}
\hline
\multicolumn{1}{|c|}{\multirow{2}{*}{\textbf{Model}}} & \multicolumn{3}{c|}{\textbf{test-intra}} & \multicolumn{3}{c|}{\textbf{test-inter}} \\ \cline{2-7} 
\multicolumn{1}{|c|}{} &
  \begin{tabular}[c]{@{}c@{}}Vertex\\ $(\times10^{-4})$\end{tabular} &
  \begin{tabular}[c]{@{}c@{}}Normal\\ (degrees)\end{tabular} &
  \begin{tabular}[c]{@{}c@{}}Chamfer\\ $(\times10^{-4})$\end{tabular} &
  \begin{tabular}[c]{@{}c@{}}Vertex\\ $(\times10^{-4})$\end{tabular} &
  \begin{tabular}[c]{@{}c@{}}Normal\\ (degrees)\end{tabular} &
  \begin{tabular}[c]{@{}c@{}}Chamfer\\ $(\times10^{-4})$\end{tabular} \\ \hline
With Attention                               & $27.881$ & $28.689$ & $6.845$ & $31.399$ & $28.493$ & $7.974$          \\ \hline
Without Attention                            & \textbf{3.917} & \textbf{25.458} & \textbf{2.028} & \textbf{3.717} & \textbf{25.184} & \textbf{2.035}          \\ \hline
\end{tabular}
}
\end{adjustbox}
\end{center}
\end{table}

\begin{table}[t]
\begin{center}
\renewcommand{\arraystretch}{1.3}
\caption{Network Structure - Ablation Study.}
\label{table:structure_ablation}
\footnotesize
\begin{adjustbox}{width=\linewidth,center}
{
\begin{tabular}{|l|l|l|l|l|l|l|}
\hline
\multicolumn{1}{|c|}{\multirow{2}{*}{\textbf{Model}}} & \multicolumn{3}{c|}{\textbf{test-intra}} & \multicolumn{3}{c|}{\textbf{test-inter}} \\ \cline{2-7} 
\multicolumn{1}{|c|}{} & \multicolumn{1}{c|}{\begin{tabular}[c]{@{}c@{}}Vertex\\ $(\times10^{-4})$\end{tabular}} & \multicolumn{1}{c|}{\begin{tabular}[c]{@{}c@{}}Normal\\ (degrees)\end{tabular}} & \multicolumn{1}{c|}{\begin{tabular}[c]{@{}c@{}}Chamfer\\ $(\times10^{-4})$\end{tabular}} & \multicolumn{1}{c|}{\begin{tabular}[c]{@{}c@{}}Vertex\\ $(\times10^{-4})$\end{tabular}} & \multicolumn{1}{c|}{\begin{tabular}[c]{@{}c@{}}Normal\\ (degrees)\end{tabular}} & \multicolumn{1}{c|}{\begin{tabular}[c]{@{}c@{}}Chamfer\\ $(\times10^{-4})$\end{tabular}} \\ \hline
FGT-8              & \textbf{3.917} & \textbf{25.458} & \textbf{2.028} & \textbf{3.717} & \textbf{25.184} & \textbf{2.035} \\ \hline
FGT-3              & $4.506$ & $26.396$ & $2.224$ & $4.306$ & $25.917$ & $2.14$ \\ \hline
W/O T           & $6.112$ & $26.328$ & $2.849$ & $5.588$ & $25.994$ & $2.696$  \\ \hline
MT            & $5.586$ & $26.366$ & $2.653$ & $5.521$ & $26.097$ & $2.643$  \\ \hline
DO-2S     & $5.398$ & $26.224$ & $2.649$ & $4.967$ & $25.878$ & $2.548$  \\ \hline
DO-1S   & $6.097$ & $27.356$ & $2.827$ & $6.02$ & $26.801$ & $2.888$  \\ \hline
FGT-1S            & $9.62$ & $28.368$ & $3.969$ & $9.101$ & $27.963$ & $3.654$  \\ \hline
\end{tabular}
}
\end{adjustbox}
\end{center}
\end{table}

\begin{figure*}[!htb]
    \begin{center}
    \captionsetup[subfigure]{labelformat=empty}
    \subfloat[(a)]{\includegraphics[width=0.11\linewidth ]{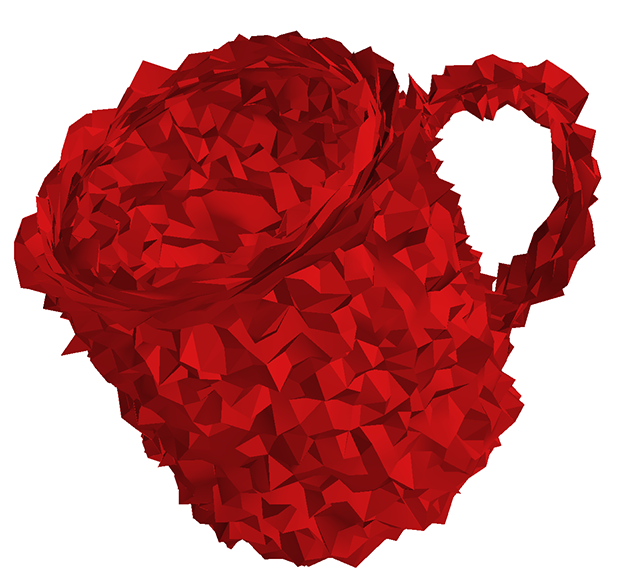}}
    \subfloat[(b)]{\includegraphics[width=0.11\linewidth ]{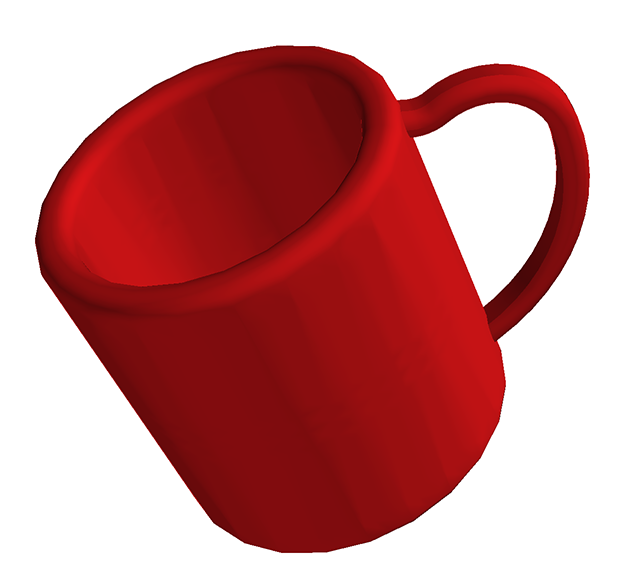}}
    \subfloat[(c)]{\includegraphics[width=0.11\linewidth ]{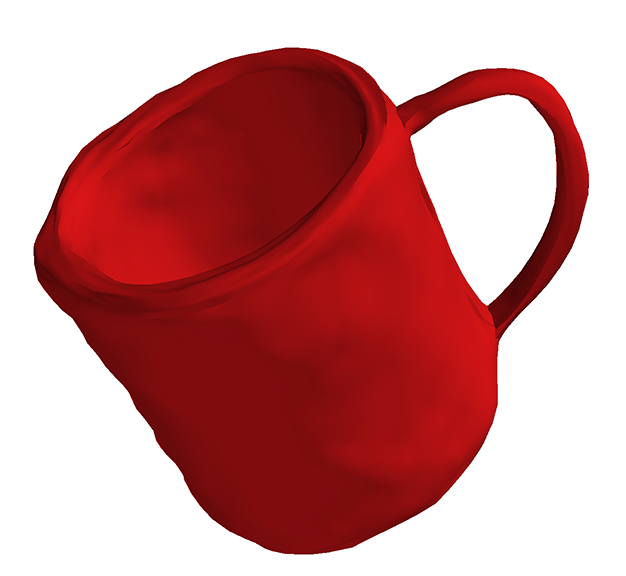}}
    \subfloat[(d)]{\includegraphics[width=0.11\linewidth ]{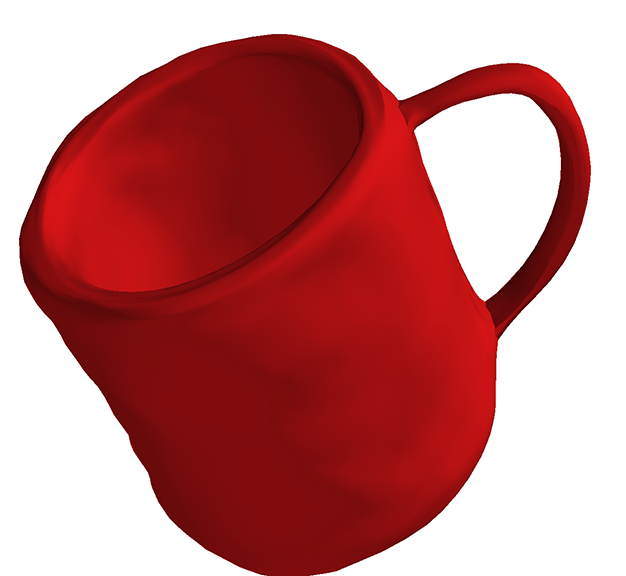}}
    \subfloat[(e)]{\includegraphics[width=0.11\linewidth ]{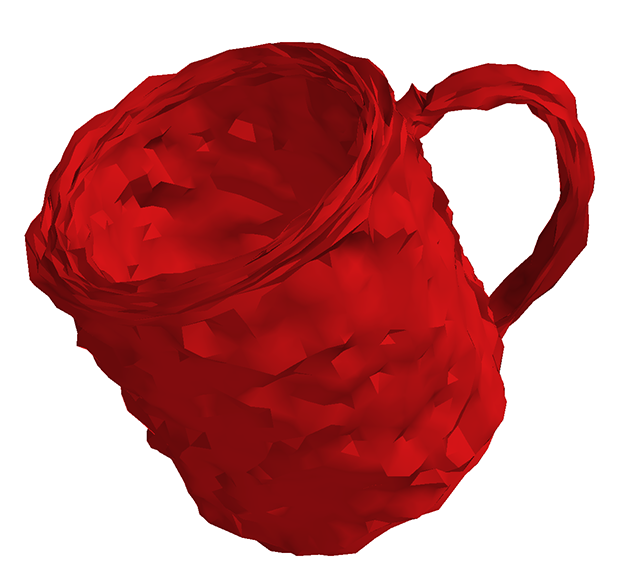}}
    \subfloat[(f)]{\includegraphics[width=0.11\linewidth ]{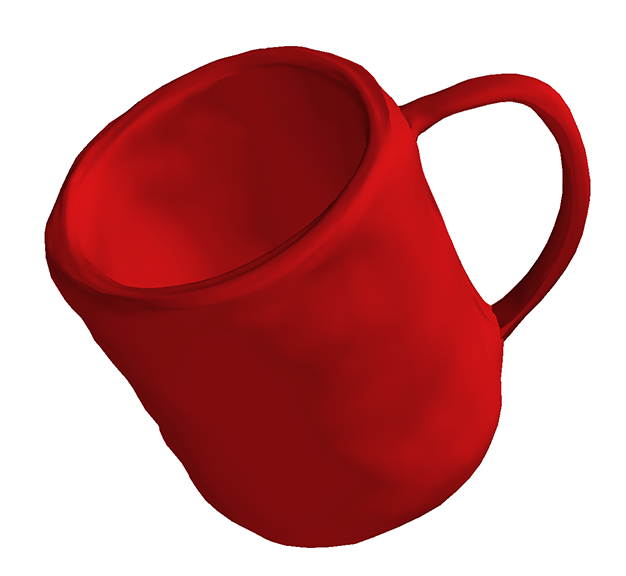}}
    \subfloat[(g)]{\includegraphics[width=0.11\linewidth ]{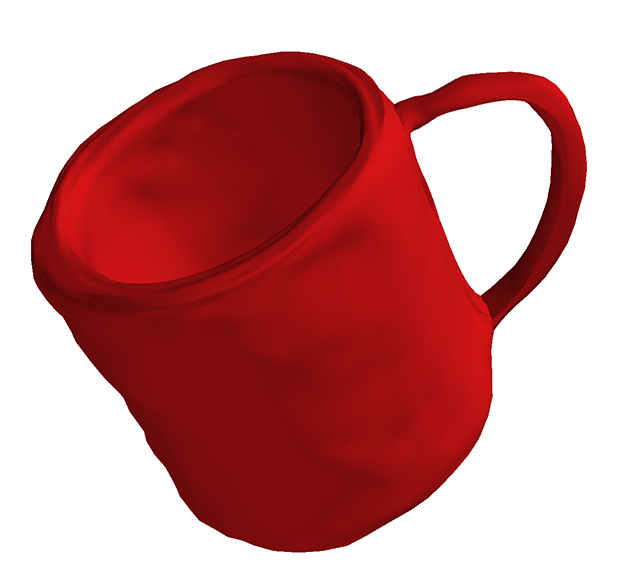}}
    \subfloat[(h1)]{\includegraphics[width=0.11\linewidth ]{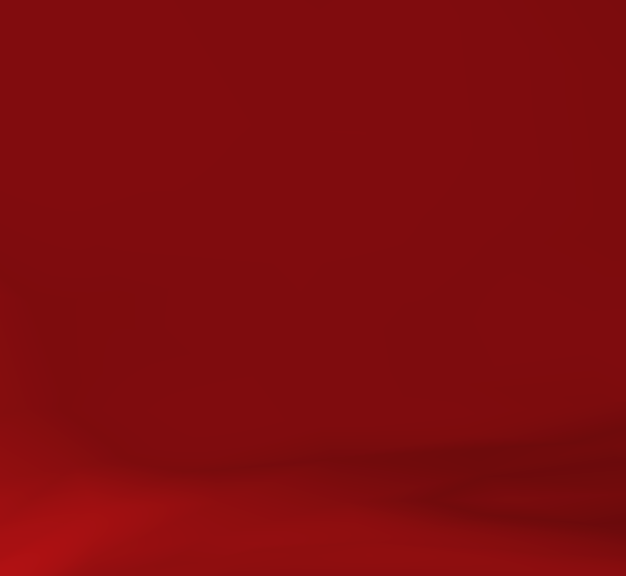}}
    \subfloat[(h2)]{\includegraphics[width=0.11\linewidth ]{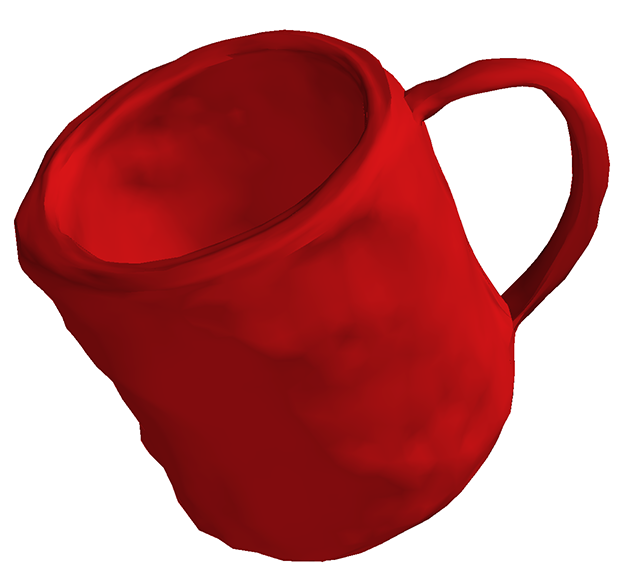}}
    \end{center}
    \caption{Output of loss ablation variants on the cup model. (a) Noisy, (b) Original, (c) With All, (d) Without Vertex, (e) Without Normal, (f) Without Curvature, (g) Without Chamfer, (h1) Without Vertex and Chamfer, and (h2) Without Vertex and Chamfer (zoomed-out). The figures from (a)-(h1) are all under the same camera setting. (h2) is the zoomed out version of (h1). }
    \label{fig:LossAblationFigs}
\end{figure*}

\begin{table*}[t]
\begin{center}
\renewcommand{\arraystretch}{1.3}
\caption{Loss Ablation - Study.}
\label{table:loss_ablation}
\footnotesize
\begin{tabular}{|l|c|c|c|c|c|c|c|c|c|c|}
\hline
\multicolumn{5}{|c|}{\textbf{Model}} & \multicolumn{3}{c|}{\textbf{test-intra}} & \multicolumn{3}{c|}{\textbf{test-inter}} \\ \hline
\multicolumn{1}{|c|}{Loss} & $\lambda_{V}$ & $\lambda_{N}$ & $\lambda_{\kappa}$ & $\lambda_{C}$ & \multicolumn{1}{c|}{\begin{tabular}[c]{@{}c@{}}Vertex\\ $(\times 10^{-4})$\end{tabular}} & \multicolumn{1}{c|}{\begin{tabular}[c]{@{}c@{}}Normal\\ (degrees)\end{tabular}} & \multicolumn{1}{c|}{\begin{tabular}[c]{@{}c@{}}Chamfer\\ $(\times 10^{-4})$\end{tabular}} & \multicolumn{1}{c|}{\begin{tabular}[c]{@{}c@{}}Vertex\\ $(\times 10^{-4})$\end{tabular}} & \multicolumn{1}{c|}{\begin{tabular}[c]{@{}c@{}}Normal\\ (degrees)\end{tabular}} & \multicolumn{1}{c|}{\begin{tabular}[c]{@{}c@{}}Chamfer\\ $(\times 10^{-4})$\end{tabular}} \\ \hline
With All & 1 & 1 & 1 & 1                         & $4.624$ & $25.98$ & $2.337$ & $4.353$ & $25.725$ & $2.264$  \\ \hline
Without Vertex & 0 & 1 & 1 & 1                   & $17.291$ & $26.214$ & $5.98$ & $14.282$ & $25.961$ & $5.188$ \\ \hline
Without Normal & 1 & 0 & 1 & 1                   & $\textbf{1.923}$ & $40.37$ & $\textbf{1.037}$ & $\textbf{1.907}$ & $39.985$ & $\textbf{1.054}$  \\ \hline
Without Curvature & 1 & 1 & 0 & 1                & $4.585$ & $26.493$ & $2.342$ & $4.551$ & $25.966$ & $2.379$  \\ \hline
Without Chamfer & 1 & 1 & 1 & 0                  & $4.601$ & $26.086$ & $2.359$ & $4.406$ & $25.844$ & $2.286$  \\ \hline
Without Vertex and Chamfer & 0 & 1 & 1 & 0       & $1.039 \times 10^{16}$ & $\textbf{25.302}$ & $2.938 \times 10^{15}$ & $9.449\times 10^{15} $ & $\textbf{24.979}$ & $3.113\times 10^{15}$  \\ \hline
\end{tabular}
\end{center}
\end{table*}

\begin{table}[t]
\begin{center}
\renewcommand{\arraystretch}{1.3}
\caption{Depth Ablation - Study.}
\label{table:depth_ablation}
\footnotesize
\begin{adjustbox}{width=\linewidth,center}
{
\begin{tabular}{|l|c|c|c|c|c|c|}
\hline
\multicolumn{1}{|c|}{\multirow{2}{*}{\textbf{Model}}} & \multicolumn{3}{c|}{\textbf{test-intra}} & \multicolumn{3}{c|}{\textbf{test-inter}} \\ \cline{2-7} 
\multicolumn{1}{|c|}{} &
  \begin{tabular}[c]{@{}c@{}}Vertex\\ $(\times10^{-4})$\end{tabular} &
  \begin{tabular}[c]{@{}c@{}}Normal\\ (degrees)\end{tabular} &
  \begin{tabular}[c]{@{}c@{}}Chamfer\\ $(\times10^{-4})$\end{tabular} &
  \begin{tabular}[c]{@{}c@{}}Vertex\\ $(\times10^{-4})$\end{tabular} &
  \begin{tabular}[c]{@{}c@{}}Normal\\ (degrees)\end{tabular} &
  \begin{tabular}[c]{@{}c@{}}Chamfer\\ $(\times10^{-4})$\end{tabular} \\ \hline
DMD-Net 1                                   & $5.498$ & $26.12$ & $2.487$ & $4.968$ & $25.8$ & $2.369$          \\ \hline
DMD-Net 2                                   & \textbf{3.301} & \textbf{25.37} & \textbf{1.786} & \textbf{3.271} & \textbf{25.053} &\textbf{ 1.815}          \\ \hline
DMD-Net 3                                   & $4.968$ & $26.004$ & $2.393$ & $4.755$ & $25.509$ & $2.362$          \\ \hline
DMD-Net 4                                   & $5.161$ & $25.98$ & $2.493$ & $4.812$ & $25.575$ & $2.246$          \\ \hline
\end{tabular}
}
\end{adjustbox}
\end{center}
\end{table}

\subsection{Dropout Rate Ablation}

In this study we include dropout layer in our network after each ReLU function. We experiment with four different dropout rate $D_r$: (a) $D_r=0.5$, (b) $D_r=0.35$, (c) $D_r=0.2$, and (d) $D_r=0$. In Table \ref{table:dropout}, we find that including dropout layer worsens the performance. Therefore, we do not include dropout layers in our network.

\subsection{Graph Attention Networks Ablation}

We test the effect of including graph attention network in the graph aggregation layer. We find in Table \ref{table:attention}, that, the inclusion of graph attention network degrades the performance. Hence, we exclude graph attention network from the proposed approach.

\subsection{Structure Ablation}

In this study we modify the structure of our network in various ways and show that our network in its current form performs the best. We devise the following variants:

\noindent \textbullet\ \textit{FGT-8}.
This is the architecture we propose in our work. In this method, the noisy input mesh is combined with the estimated local features before being transformed by the transformer.

\noindent \textbullet\ \textit{FGT-3}.
FGT-3 is similar in structure to FGT-8 except that, in this method, the noisy input mesh is directly passed to the transformer without being combined with the estimated local features.

\noindent \textbullet\ \textit{Without Transformer (W/O T)}.
Here we eliminate the transformer. The combination of the local features and the noisy mesh are directly passed to the primal stream of the denoiser. 

\noindent \textbullet\ \textit{Multiple Transformer (MT)}.
Here instead of computing a single transformation matrix, the transformer computes a cascade of transformations which is then applied to the combination of noisy mesh and the estimated features.

\noindent \textbullet\ \textit{Denoiser Only - Two Stream (DO-2S)}.
Here we eliminate both the feature extractor as well as the transformer. The noisy mesh is directly fed to both the streams of the denoiser. The purpose of this network is to validate the use of the feature extractor and transformer in the FGT paradigm.

\noindent \textbullet\ \textit{Denoiser Only - Single Stream (DO-1S)}.
This is same as the Denoiser Only (Two Stream) except that now the denoiser contains only a single stream. This also implies that the two-stream network does not contain the primal dual fusion block. The purpose of this network is to test how well a vanilla graph neural network works on the mesh denoising task.

\noindent \textbullet\ \textit{FGT - Single Stream (FGT-1S)}.
This is similar in structure to the proposed approach FGT-8, the only difference being that both the denoiser and the feature extractor contain a single stream. The purpose of this network is to validate the use of doing aggregation in both primal and dual graph.

The comparison is shown in Table \ref{table:structure_ablation}. We find that the proposed network FGT-8 performs the best. We use this architecture in the consequent ablation studies.

\subsection{Loss Ablation}

In this study we experiment with the weights of several loss functions. This study is conducted to find out the relevance of each loss function. Table \ref{table:loss_ablation} shows the results for various different configurations. The first row (With All) refers to the case where all the loss functions are used. In the second row (Without Vertex) when we drop only the vertex loss, we observe that all the three metrics degrade, thereby indicating the significance of vertex loss. When we drop only the normal loss (Without Normal), we observe significant improvement in two metrics but large degradation in the normal metric. When we drop curvature loss (Without Curvature) or chamfer loss (Without Chamfer), we do not see significant change in the three metrics. This indicates that chamfer loss and curvature loss are not of very high significance. When we drop both vertex loss and chamfer loss, we see a very huge jump in the vertex and chamfer metric, indicating that normal loss alone is not sufficient. Based on these results and some further experimentation, we arrive at the final weights of the loss functions: $\lambda_{V} = 1$, $\lambda_{N} = 0.2$, $\lambda_{\kappa} = 0.01$, $\lambda_{FE} = 1$, and $\lambda_{C} = 0.05$. The visual comparison of these loss ablation variants are made in Figure \ref{fig:LossAblationFigs}. In Figure \ref{fig:LossAblationFigs}(a)-(h1) the camera setting are kept same during rendering. Figure \ref{fig:LossAblationFigs}(h2) is the zoomed out image of \ref{fig:LossAblationFigs}(h1) to fit the entire object inside the field of view of the camera. As visible in Figure \ref{fig:LossAblationFigs}(h2), in the absence of both vertex loss and chamfer loss the output model achieves shape similar to the ground truth. But the object is a highly scaled up version hence the red patch in Figure \ref{fig:LossAblationFigs}(h1). This scaling up of the object is the reason why the vertex loss metric and the chamfer loss metric is very high in last row of Table \ref{table:loss_ablation}. As can be seen in the Figure \ref{fig:LossAblationFigs}(e), DMD-Net without normal loss performs very poorly, thereby signifying the importance of normal loss though this variant achieves the best vertex loss (Table \ref{table:loss_ablation})

\subsection{Depth Ablation}

Depth is defined as the number of two-stream networks in the denoiser. We perform this experiment with four different variants: 
\noindent (a) \textit{DMD-Net 1}.
contains only one two-stream networks
\noindent (b) \textit{DMD-Net 2}.
contains two two-stream networks
\noindent (c) \textit{DMD-Net 3}.
contains three two-stream networks
\noindent (d) \textit{DMD-Net 4}.
contains four two-stream networks
In Table \ref{table:depth_ablation} we find out that DMD-Net 2 performs the best. Thus, based on these results, we include only two two-stream networks in the denoiser in our proposed approach. Note that in depth ablation study all variant are trained for 200 epochs.


\section{Transformation Equivariance}

First we define what we mean by transformation equivariance. Let $\mathcal{T}$ be a transformation operator, let $\mathcal{D}$ be a mesh denoising algorithm and let $\mathcal{G}$ denote a mesh. We say that $\mathcal{D}$ has $\mathcal{T}$-equivariance, if for any given mesh, $\mathcal{G}$ we have  $\mathcal{T}(\mathcal{D}(\mathcal{G})) = \mathcal{D}(\mathcal{T}(\mathcal{G}))$. That is,  $\mathcal{D}$ is $\mathcal{T}$-equivariant, if  $\mathcal{D}$ and $\mathcal{T}$ commute. We now discuss whether DMD-Net is equivariant with respect to the following mentioned transformations.


\subsubsection{Scaling Equivariance}
Given a mesh, we first normalize it to fit inside a unit cube. We then denoise it using DMD-Net and unnormalize it back to it's original scale. Thus DMD-Net is scale equivariant as it first converts the mesh into a cannonical scale before denoising.
    
\subsection{Translation Equivariance}
Before denoising, we shift the mesh to the origin and after denoising, we shift it back to its original location. Thus, DMD-Net is translation equivariant as the object is always aligned to the origin.

\begin{table}[]
\begin{center}
\renewcommand{\arraystretch}{1.3}
\caption{Comparison of the proposed approach with and without Spatial Transformer Networks.}
\label{table:STN}
\footnotesize
\begin{adjustbox}{width=\linewidth,center}
{
\begin{tabular}{|l|c|c|c|c|c|c|}
\hline
\multicolumn{1}{|c|}{\multirow{2}{*}{\textbf{Model}}} & \multicolumn{3}{c|}{\textbf{test-intra}} & \multicolumn{3}{c|}{\textbf{test-inter}} \\ \cline{2-7} 
\multicolumn{1}{|c|}{} &
  \begin{tabular}[c]{@{}c@{}}Vertex\\ $(\times10^{-4})$\end{tabular} &
  \begin{tabular}[c]{@{}c@{}}Normal\\ (degrees)\end{tabular} &
  \begin{tabular}[c]{@{}c@{}}Chamfer\\ $(\times10^{-4})$\end{tabular} &
  \begin{tabular}[c]{@{}c@{}}Vertex\\ $(\times10^{-4})$\end{tabular} &
  \begin{tabular}[c]{@{}c@{}}Normal\\ (degrees)\end{tabular} &
  \begin{tabular}[c]{@{}c@{}}Chamfer\\ $(\times10^{-4})$\end{tabular} \\ \hline
W STN                     & $5.233$ & $26.238$ & $2.471$ & $4.692$ & $25.927$ & $2.189$          \\ \hline
W/O STN                  & \textbf{3.301} & \textbf{25.37} &\textbf{1.786} & \textbf{3.271} & \textbf{25.053} & \textbf{1.815}           \\ \hline
\end{tabular}
}
\end{adjustbox}
\vspace{-0.5cm}
\end{center}
\end{table}


\begin{figure}[t]
    \begin{center}
    \captionsetup[subfigure]{labelformat=empty}
    \includegraphics[width=0.24\linewidth ]{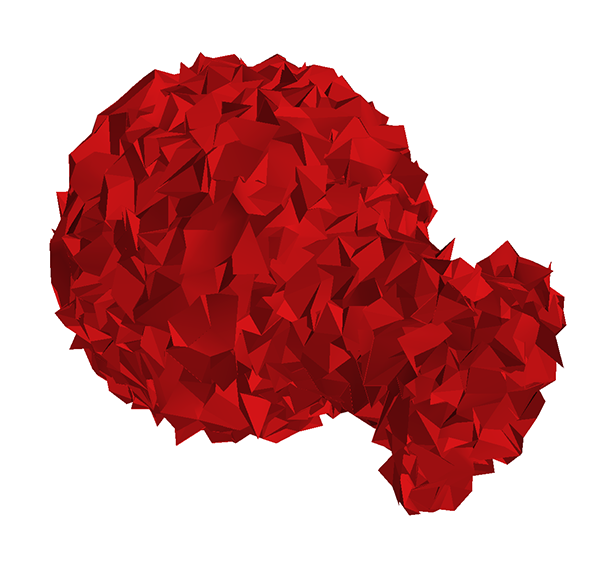}
    \includegraphics[width=0.24\linewidth ]{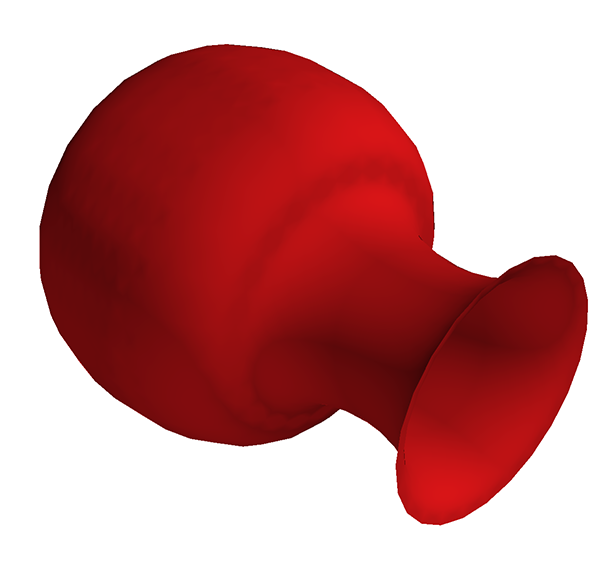}
    \includegraphics[width=0.24\linewidth ]{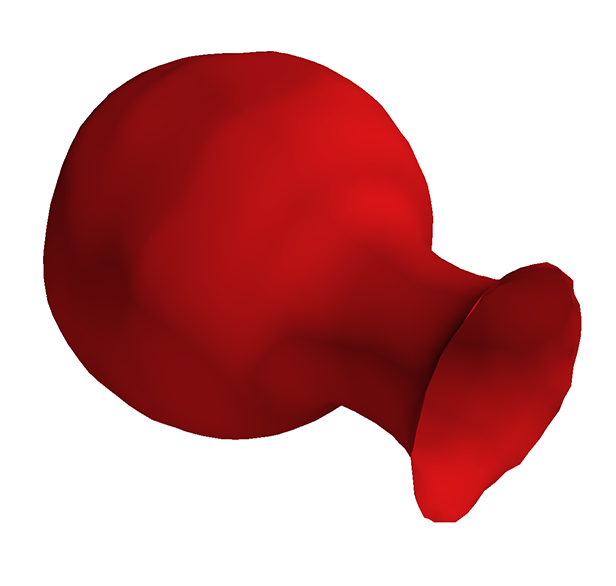}
    \includegraphics[width=0.24\linewidth ]{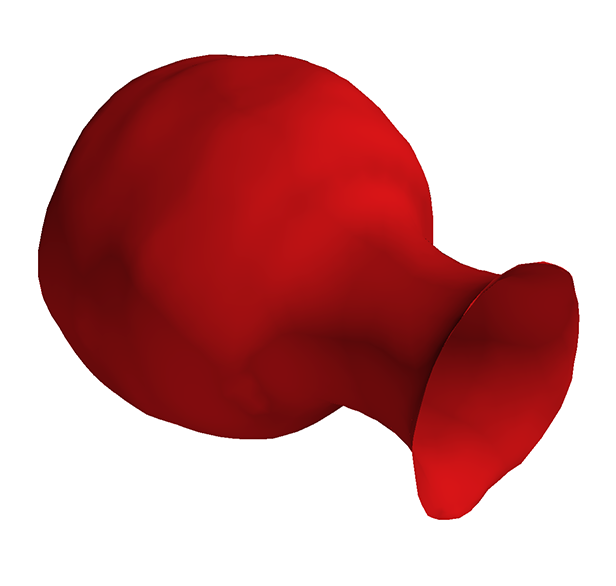}\\
    \includegraphics[width=0.24\linewidth ]{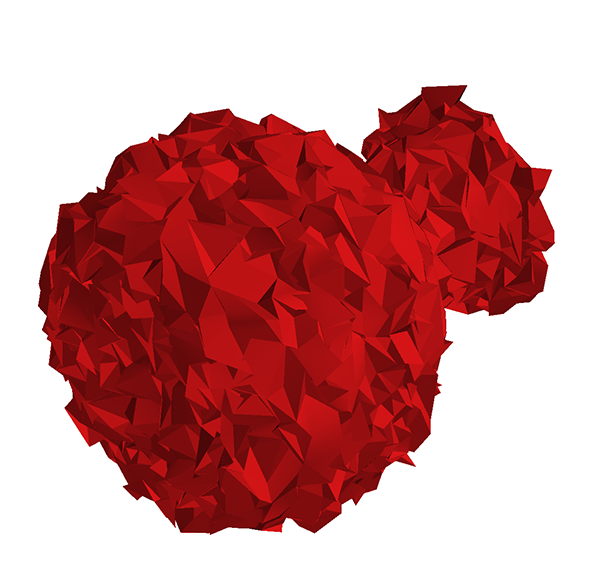}
    \includegraphics[width=0.24\linewidth ]{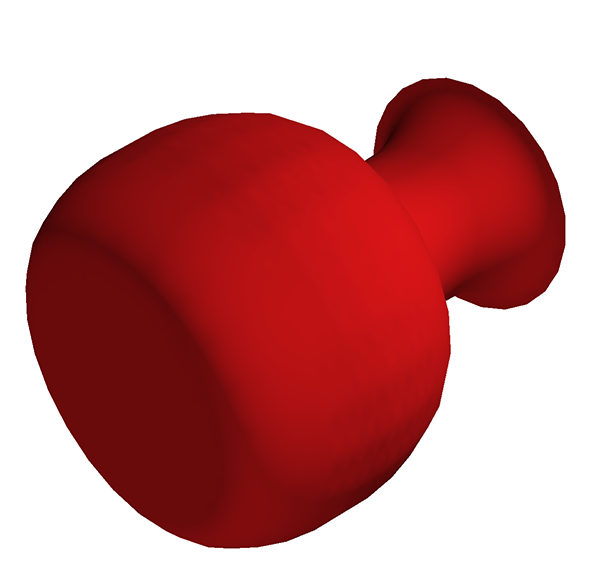}
    \includegraphics[width=0.24\linewidth ]{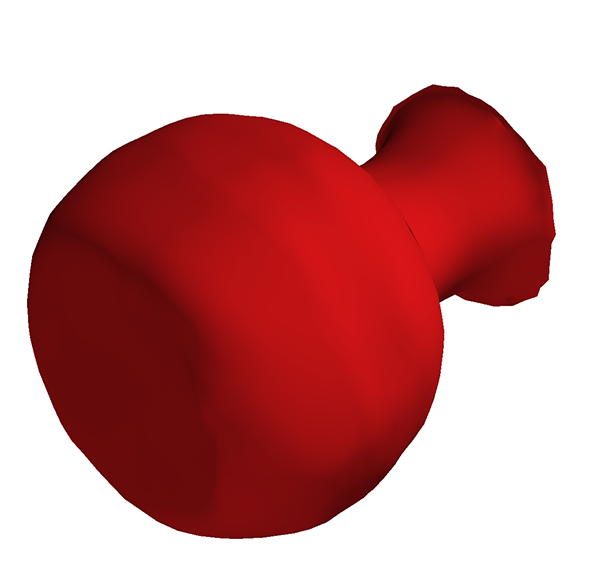}
    \includegraphics[width=0.24\linewidth ]{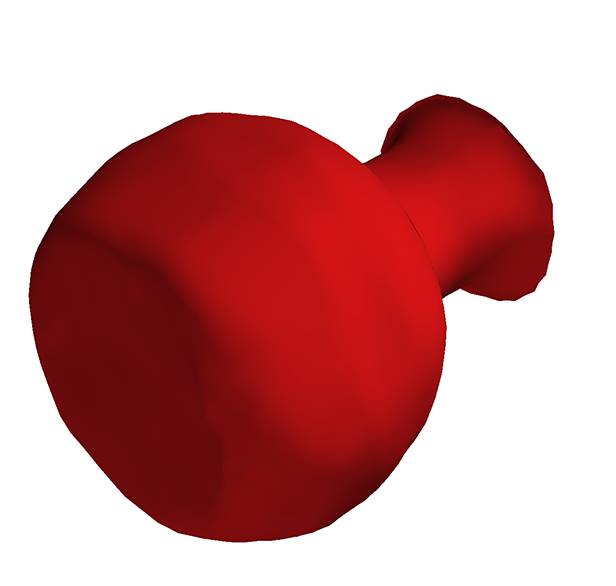}\\
    \includegraphics[width=0.24\linewidth ]{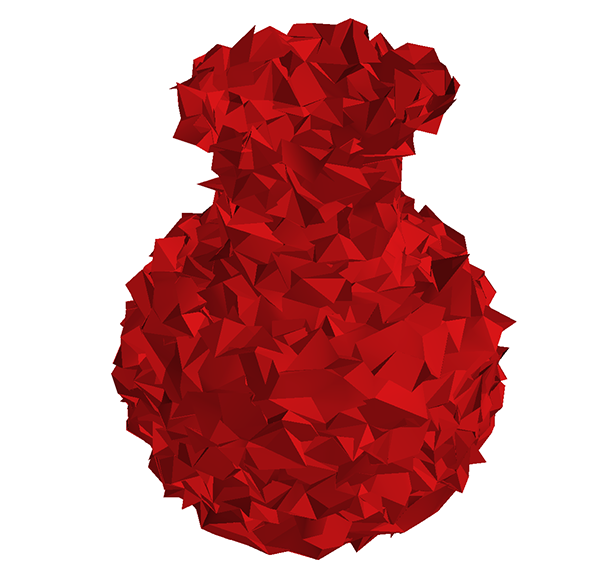}
    \includegraphics[width=0.24\linewidth ]{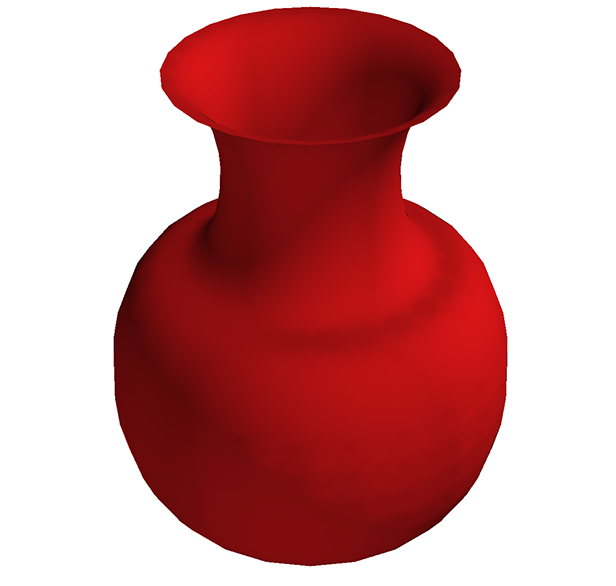}
    \includegraphics[width=0.24\linewidth ]{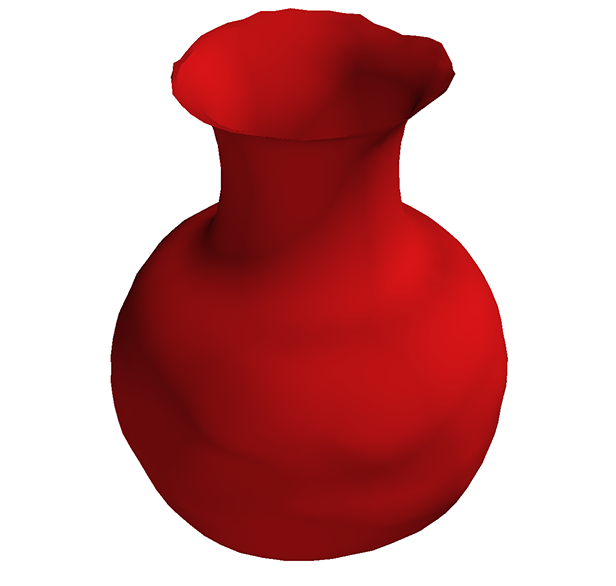}
    \includegraphics[width=0.24\linewidth ]{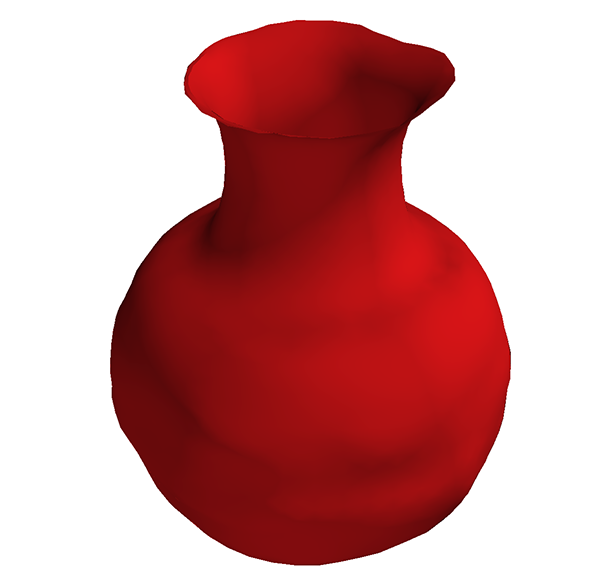}\\
    \includegraphics[width=0.24\linewidth ]{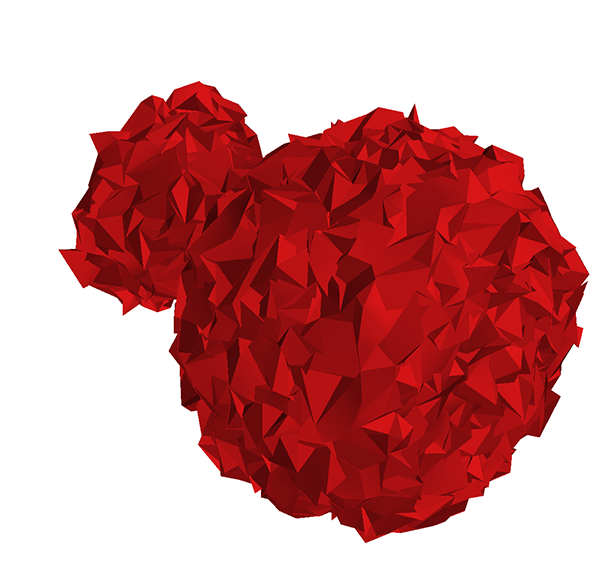}
    \includegraphics[width=0.24\linewidth ]{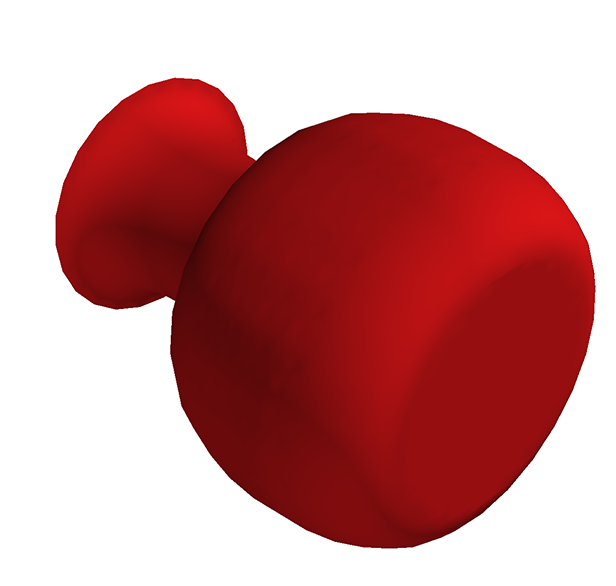}
    \includegraphics[width=0.24\linewidth ]{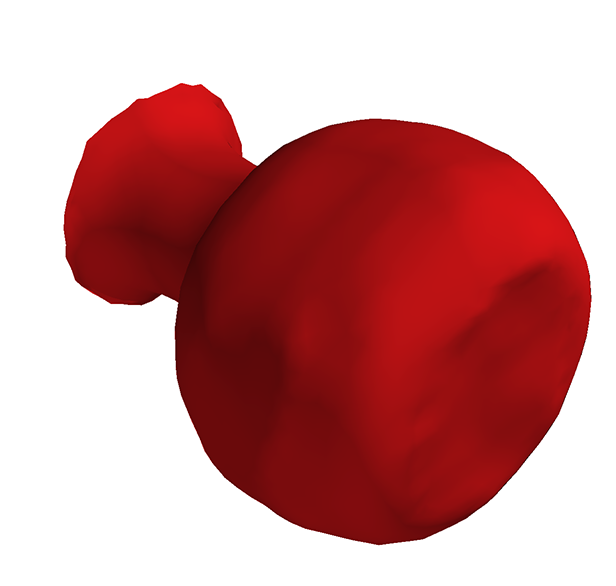}
    \includegraphics[width=0.24\linewidth ]{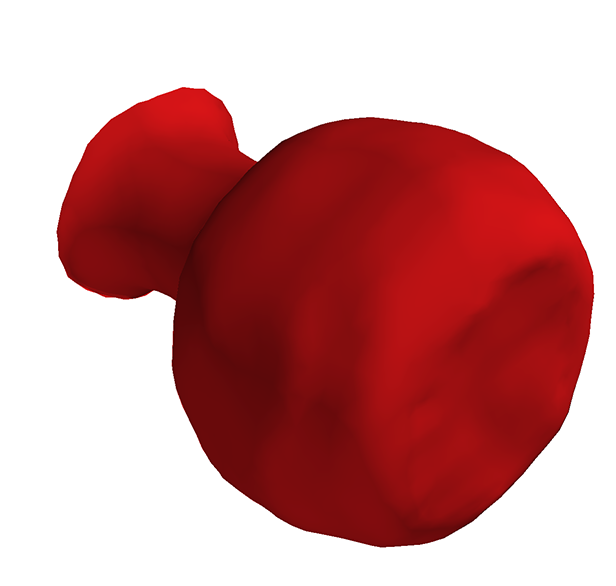}\\
    \vspace{-0.4cm}
    \subfloat[Noisy]{\includegraphics[width=0.25\linewidth ]{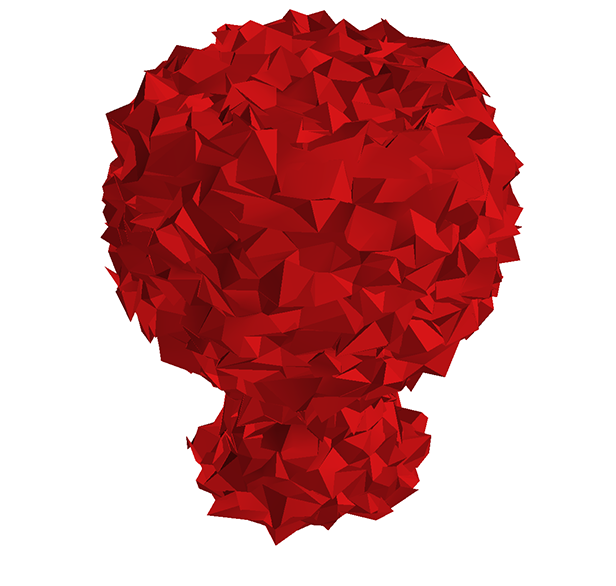}}
    \subfloat[Original]{\includegraphics[width=0.25\linewidth ]{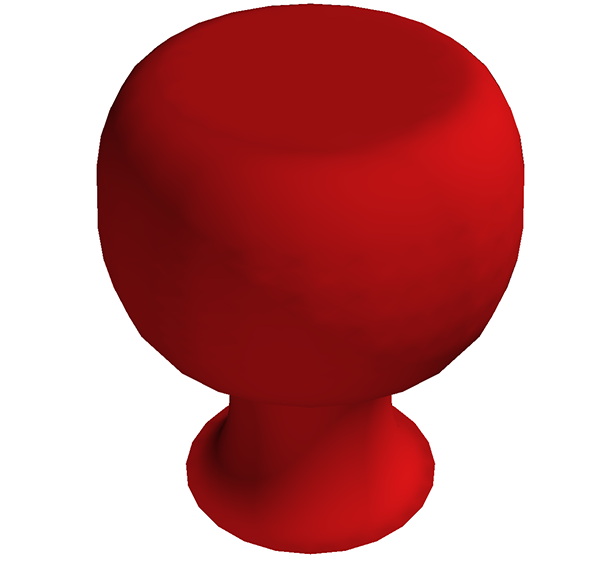}}
    \subfloat[DMD-Net]{\includegraphics[width=0.25\linewidth ]{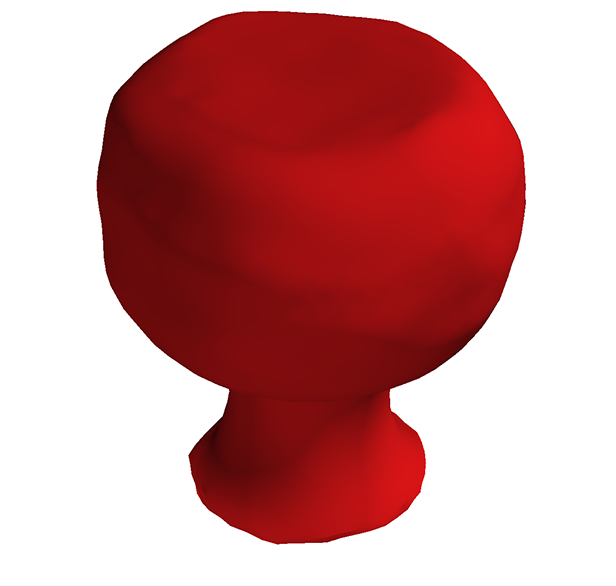}}
    \subfloat[DMD-Net+STN]{\includegraphics[width=0.25\linewidth ]{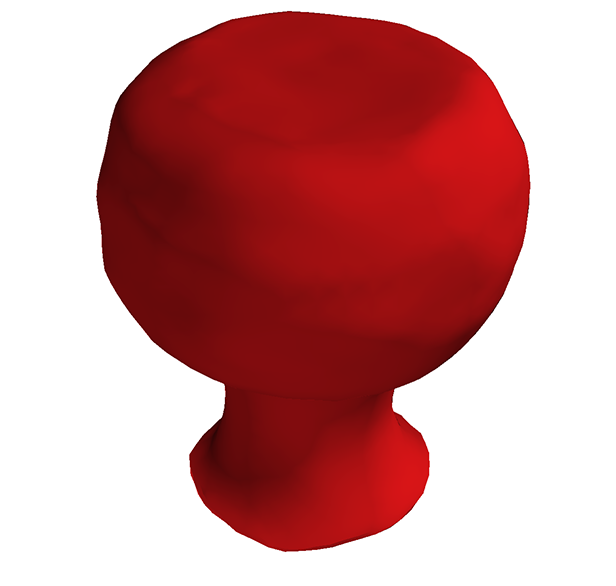}}
    \end{center}
    \caption{Visualization of the degree to which rotation equivariance is achieved by DMD-Net and DMD-Net+STN. Objects in the first column are rotated versions of each other. The second column refers to the corresponding ground truth objects. Third column and the fourth column are the outputs of the DMD-Net and DMD-Net+STN. }
    \label{fig:RotationEQ}
    \vspace{-0.5cm}
\end{figure}

DMD-Net is not rotation equivariant in the theoretical sense. However, we try to establish rotation equivariance by introducing concepts like rotation augmentation in the dataset and the use of spatial transformer networks (STN). In rotation augmentation, we augment the data during training by randomly rotating the ground truth mesh before adding noise. Spatial transformer networks (STN) was used by to canonicalize the orientation of the object to establish rotation invariance/equivariance in a deep learning framework that processes point clouds. We experiment with the idea of STN to check whether it establishes equivariance in our case. Through the use of a rotation equivariance test we evaluate the degree to which rotation equivariance is established in DMD-Net by these concepts. We first combine the test-intra and the test-inter sets to obtain a single test set $\mathcal{S}$. For the $i^{th}$ mesh $\mathcal{G}_i$ in this set we choose $\gamma=5$ random rotation operators, the $j^{th}$ rotation operator denoted as $\mathcal{R}_{ij}$. We then compute the following two meshes:  $\mathcal{G}_{ij}^\mathcal{RD} = \mathcal{R}_{ij}(\mathcal{D}(\mathcal{G}_i))$ and $\mathcal{G}_{ij}^\mathcal{DR} = \mathcal{D}(\mathcal{R}_{ij}(\mathcal{G}_i))$. We then find the distance between $\mathcal{G}_{ij}^\mathcal{RD}$ and $\mathcal{G}_{ij}^\mathcal{DR}$ by evaluating the following loss metrics: $\mathcal{L}_{vertex}(\mathcal{G}_{ij}^\mathcal{RD}, \mathcal{G}_{ij}^\mathcal{DR})$, $\mathcal{L}_{normal}(\mathcal{G}_{ij}^\mathcal{RD}, \mathcal{G}_{ij}^\mathcal{DR})$, and $\mathcal{L}_{chamfer}(\mathcal{G}_{ij}^\mathcal{RD}, \mathcal{G}_{ij}^\mathcal{DR})$. We then sum these loss metrics over the entire test set $\mathcal{S}$ as mentioned in Equations \ref{eq:RE_test_vertex}, \ref{eq:RE_test_normal}, and \ref{eq:RE_test_chamfer}.

    \begin{equation}
    \mathcal{L}_\mathcal{V}^\mathcal{RE} = \frac{1}{\gamma|S|}\sum_{i=1}^{|S|}\sum_{j=1}^{\gamma} \mathcal{L}_{vertex}(\mathcal{G}_{ij}^\mathcal{RD}, \mathcal{G}_{ij}^\mathcal{DR})
    \label{eq:RE_test_vertex}
    \end{equation}
    \begin{equation}
    \mathcal{L}_\mathcal{N}^\mathcal{RE} = \frac{1}{\gamma|S|}\sum_{i=1}^{|S|}\sum_{j=1}^{\gamma} \mathcal{L}_{normal}(\mathcal{G}_{ij}^\mathcal{RD}, \mathcal{G}_{ij}^\mathcal{DR})
    \label{eq:RE_test_normal}
    \end{equation}
    \begin{equation}
    \mathcal{L}_\mathcal{C}^\mathcal{RE} = \frac{1}{\gamma|S|}\sum_{i=1}^{|S|}\sum_{j=1}^{\gamma} \mathcal{L}_{chamfer}(\mathcal{G}_{ij}^\mathcal{RD}, \mathcal{G}_{ij}^\mathcal{DR})
    \label{eq:RE_test_chamfer}
    \end{equation}

\begin{table}[t]
\begin{center}
\renewcommand{\arraystretch}{1.3}
\caption{Rotation Equivariance Test for measuring the degree to which rotation equivariance is established.}
\label{table:REq}
\footnotesize
\begin{adjustbox}{width=\linewidth,center}
{
\begin{tabular}{|l|c|c|c|}
\hline
\multicolumn{1}{|c|}{\multirow{2}{*}{\textbf{Model}}} & \multicolumn{3}{c|}{\textbf{test intra + test inter}} \\ \cline{2-4} 
\multicolumn{1}{|c|}{} &
  \begin{tabular}[c]{@{}c@{}}Vertex\\ $(\times 10^{-4})$\end{tabular} &
  \begin{tabular}[c]{@{}c@{}}Normal\\ (degrees)\end{tabular} &
  \begin{tabular}[c]{@{}c@{}}Chamfer\\ $(\times 10^{-4})$\end{tabular} \\ \hline
Rotation Augmentation                   &   \textbf{0.704}         &   5.998        &     \textbf{0.501}      \\ \hline
Rotation Augmentation + STN  &   3.723         &   \textbf{5.711}        &    1.796         \\ \hline
\end{tabular}
}
\end{adjustbox}
\end{center}
\end{table}

We conduct rotation equivariance test in two cases: (a) Rotation Augmentation and (b) Rotation Augmentation with Spatial Transformer Network. The results are depicted in Table \ref{table:REq}. In terms of $ \mathcal{L}_\mathcal{V}^\mathcal{RE}$ and $ \mathcal{L}_\mathcal{C}^\mathcal{RE}$, case (a) performs far better but in terms of $ \mathcal{L}_\mathcal{N}^\mathcal{RE}$, case(b) performs slightly better. Thus, we discover that introducing STN does not provide significant improvement but rather degrades the overall equivariance metric. Moreover, in Table \ref{table:STN} we assess the effect of adding the STN by comparing the results on the mesh denoising task. This table has a similar setting as that of the ablation studies mentioned in Section \ref{sec:ablation}. We conclude that introducing STN does not establish rotation equivariance to a larger degree. Moreover, it also degrades the performance on mesh denoising. Thus, we do not include STN in the proposed approach and only use rotation augmentation for rotation equivariance. In Figure \ref{fig:RotationEQ}, we visually show that DMD-Net has rotation equivariance to a large degree. The five rows in the figure correspond to five different rotations. All the images are rendered using a fixed camera position and orientation.

\section{Supplementary Video}
A brief audio-visual explanation of DMD-Net is included in
the supplementary video, which has been uploaded online as an unlisted video on youtube. The video can be accessed through the following link: \underline{\textcolor{blue}{\url{https://youtu.be/wPJvUYqiL94}}}

\end{document}